\pdfoutput=1
\documentclass[journal,10pt]{IEEEtran}
\ifCLASSINFOpdf
\else
\fi

\usepackage[mathscr]{eucal} 
\usepackage{amsmath,amsthm,amscd,amssymb,amsbsy}
\usepackage{graphicx}
\usepackage[caption=false,font=footnotesize]{subfig}
\usepackage{epstopdf}
\usepackage{algorithm}
\usepackage{algorithmic}
\usepackage{verbatim}
\usepackage{array}
\usepackage{cite}
\usepackage{color}

\def\hx{{\hat{\mathbf x}}}
\def\hk{{\hat{\mathbf k}}}

\def\k{{\mathbf k}}
\def\x{{\mathbf x}}
\def\v{{\mathbf v}}
\def\b{{\mathbf b}}
\def\n{{\mathbf n}}

\def\h{{\mathbf h}}
\def\H{{\mathbf H}}
\def\A{{\mathbf A}}
\def\B{{\mathbf B}}
\def\K{{\mathbf K}}

\def\tu{\tilde{u}}
\def\tv{\tilde{v}}

\def\cF{{\mathcal F}}

\DeclareMathOperator*{\argmin}{\arg\min}

\newtheorem{claim}{Claim}

\newcolumntype{C}[1]{>{\centering}p{#1}}

\graphicspath{{figures/}}

\begin{document}
%
\title{Single Image Blind Deblurring Using Multi-Scale Latent Structure Prior}
%
%
%

\author{Yuanchao~Bai,~\IEEEmembership{Student Member,~IEEE,}
        Huizhu~Jia,
        Ming~Jiang,~\IEEEmembership{Senior Member,~IEEE,}\\
        Xianming~Liu,~\IEEEmembership{Member,~IEEE,}
        Xiaodong~Xie,
        Wen~Gao,~\IEEEmembership{Fellow,~IEEE}
\thanks{Yuanchao~Bai, Huizhu~Jia, Xiaodong~Xie and Wen~Gao are with the School of Electronics Engineering and Computer Science,
Peking University, Beijing, 100871, China. Huizhu~Jia is also with Cooperative Media-net Innovation Center and Beida~(Binhai) Information Research, Tianjin, 300450, China. Huizhu~Jia is the corresponding author (e-mail: \{yuanchao.bai, hzjia, donxie, wgao\}@pku.edu.cn).}
\thanks{Ming~Jiang is with the School of Mathematical Sciences, Peking University,
Beijing, 100871, China (e-mail: ming-jiang@pku.edu.cn).}
\thanks{Xianming~Liu is with the School of Computer Science and Technology, Harbin Institute of Technology,
Harbin, 150001, China (e-mail: csxm@hit.edu.cn).}
\thanks{Copyright \textcircled{c} 2019 IEEE. Personal use of this material is permitted. However, permission to use this material for any other purposes must be obtained from the IEEE by sending an email to pubs-permissions@ieee.org.}
}

\maketitle
\IEEEpeerreviewmaketitle
\begin{abstract}
    Blind image deblurring is a challenging problem in computer vision, which aims to restore both the blur kernel and the latent sharp image from only a blurry observation.
    Inspired by the prevalent self-example prior in image super-resolution, in this paper, we observe that a coarse enough image down-sampled from a blurry observation is approximately a low-resolution version of the latent sharp image.
    We prove this phenomenon theoretically and define the coarse enough image as a latent structure prior of the unknown sharp image. Starting from this prior, we propose to restore sharp images from the coarsest scale to the finest scale on a blurry image pyramid, and progressively update the prior image using the newly restored sharp image. These coarse-to-fine priors are referred to as \textit{Multi-Scale Latent Structures} (MSLS).
    Leveraging the MSLS prior, our algorithm comprises two phases: 1) we first preliminarily restore sharp images in the coarse scales; 2) we then apply a refinement process in the finest scale to obtain the final deblurred image.
    In each scale, to achieve lower computational complexity, we alternately perform a sharp image reconstruction with fast local self-example matching, an accelerated kernel estimation with error compensation, and a fast non-blind image deblurring, instead of computing any computationally expensive non-convex priors. We further extend the proposed algorithm to solve more challenging non-uniform blind image deblurring problem.
    Extensive experiments demonstrate that our algorithm achieves competitive results against the state-of-the-art methods with much faster running speed.
\end{abstract}


%
\IEEEpeerreviewmaketitle
\section{Introduction}
\label{sec:introduction}
\IEEEPARstart{B}{lur} is one of the most common artifacts of digital images. It happens mainly because the camera does not locate in focus, or the camera is held unsteadily, or the objects in the scene are moving quickly over the period of exposure time, resulting in that image sensor accumulates light not only from a single point but also from neighbouring regions. Consequently, a blurry image is captured with unclear edges and details, which greatly degrade the visual quality of the image. As a fundamental degradation model, the blur process is usually assumed to be shift-invariant, and can be represented as the convolution of a sharp image and a blur kernel,
\begin{equation}
    \b=\k\otimes\x+\n,
    \label{eq:blur_model_x}
\end{equation}
where $\b$ is the observed blurry image, $\x$ is the latent sharp image, $\k$ is the blur kernel, $\otimes$ is the convolution operation, and $\n$ is the additive Gaussian noise.

The blind image deblurring problem is to recover both the latent sharp image $\x$ and the blur kernel $\k$, given only the observed blurry image $\b$ \cite{almeida2010tip,Levin2011Pami}. It is a highly ill-posed problem because the solution is not only unstable but also non-unique. As a convolution in the spatial domain is equivalent to the point-wise product in the frequency domain, the observation $\b$ can be factorized into infinite feasible solution pairs of $\x$ and $\k$. It is difficult to achieve the unique correct pair without sufficient prior knowledge about $\x$ and $\k$.

Previous methods \cite{Fergus06,Levin2011EML,Krishnan2011,cai2012framelet,xuli2013Unnatural,Pan2016Pami,Lai2005color,Pan_2016_CVPR,Ren2016Low,Bai2019TIP} treated the blind image deblurring as a joint optimization problem and introduced sophisticated image priors, which can penalize image blurriness and promote image sharpness, such as the mixture of Gaussian functions that fits the heavy-tailed prior of natural images \cite{Fergus06,Levin2011EML}, normalized sparse prior \cite{Krishnan2011}, framelet based prior \cite{cai2012framelet}, $l_0$-norm based priors \cite{xuli2013Unnatural,Pan2016Pami}, color line prior \cite{Lai2005color}, dark channel prior \cite{Pan_2016_CVPR}, low-rank prior \cite{Ren2016Low} and graph based prior \cite{Bai2019TIP}, \emph{etc}. However, these image priors are usually computationally expensive, resulting in complicated optimization algorithms.

In this paper, we propose a novel multi-scale image prior to tackle the challenging blind image deblurring problem, inspired by the well-known self-example prior in image super-resolution (SR) \cite{ss_sr,Glasner2009,FreFat10}. In image SR, self-example prior is obtained through image down-sampling. An observed image is down-sampled to generate a coarse-scale image, and the self-example prior is modeled as mappings from patches in the coarse-scale image to the corresponding patches in the fine-scale observed image. Although the self-example prior has been successfully applied in image SR, when an observed image is blurred by an arbitrary blur kernel, such as a motion blur, there is no straightforward conclusion that the mappings can still model the image deblurring process. In \cite{Michaeli2014}, Michaeli and Irani first analyzed the effect of image down-sampling in the blind image deblurring problem. They found that image down-sampling increased the internal patch recurrence. With this observation, they proposed to promote the internal patch recurrence for blind image deblurring. However, iterative promotion of patch recurrence requires enormous complicated patch matchings \cite{TreeCANN2012ECCV}, which are extremely slow in practice. Different from \cite{Michaeli2014}, in this paper, we focus on the extreme case of image down-sampling. We keep down-sampling an observed blurry image to a very coarse level and observe that: \textit{the coarse enough image down-sampled from a blurry observation is approximately a low-resolution version of the latent sharp image}. We theoretically verify this phenomenon on 2D signals under a general down-sampling operator, and define the coarse enough image as a \textit{latent structure} prior of the latent sharp image.
We further introduce a multi-scale strategy to work in tandem with latent structure prior on an image pyramid.
Starting from the coarsest scale, we employ the prior image to blindly restore a finer scale sharp image. The newly restored sharp image is then employed as the latent structure prior of the next finer scale. The strategy works gradually from the coarsest scale to the finest scale.
Thus, we name our prior \textit{Multi-Scale Latent Structure} (MSLS) prior.

Tailored to the MSLS prior, we propose an efficient algorithm to solve the blind image deblurring problem, which is much faster than the aforementioned methods \cite{Fergus06,Levin2011EML,Krishnan2011,cai2012framelet,xuli2013Unnatural,Pan2016Pami,Lai2005color,Pan_2016_CVPR,Ren2016Low,Michaeli2014,Bai2019TIP}.
The proposed algorithm comprises two phases, \emph{i.e.}, we first preliminarily restore sharp images gradually from the coarsest scale to the finest scale, and then perform a refinement in the finest scale.
In each scale, we introduce a joint optimization to restore both the latent sharp image and the blur kernel.  Specifically, to achieve lower complexity, we alternately apply a sharp image reconstruction with fast local self-example matching inspired by image SR \cite{FreFat10}, an accelerated kernel estimation with error compensation, and a fast non-blind image deblurring.
Furthermore, beyond uniform image blur, we extend the proposed algorithm to solve a more challenging non-uniform blind image deblurring problem. Experimental results demonstrate that in both cases our algorithm achieves the state-of-the-art deblurring performance.

The contributions of this paper can be summarized as follows:

\setlength{\hangindent}{2.3em}
1) We introduce a \textit{Multi-Scale Latent Structure} (MSLS) prior of latent sharp image for deblurring. We theoretically verify this prior under a general down-sampling operation.

\setlength{\hangindent}{2.3em}
2) We propose a powerful and efficient blind image deblurring algorithm based on the MSLS prior, which achieves the state-of-the-art deblurring performance and is much faster than the previous methods with computationally expensive image priors.

\setlength{\hangindent}{2.3em}
3) We further demonstrate that the proposed algorithm can be extended to tackle more complex non-uniform motion blur, apart from simply uniform blur.

\vspace{5pt}

The outline of the paper is as follows. In Section~\ref{sec:related_works}, we review related work in image deblurring. In Section~\ref{sec:convergence}, we provide a theoretical analysis of our observation and introduce the MSLS prior. In Section~\ref{sec:formulation_algorithm}, we present the proposed blind deblurring algorithm in detail. Experiments and final conclusion are provided in Section~\ref{sec:experiments} and Section~\ref{sec:conclusion}, respectively.

\section{Related Work}
\label{sec:related_works}
Image deblurring is an ill-posed inverse problem, which is to recover a latent sharp image $\x$ from a blurry observation $\b$. Based on whether the blur kernel $\k$ is known or not, the problem has been divided into two categories, \emph{i.e.}, non-blind image deblurring and blind image deblurring.

\subsection{Non-blind Image Deblurring}
For non-blind image deblurring, the blur kernel is given and the problem is to recover the latent sharp image from the blurry observation with the blur kernel. Non-blind image deblurring is a very unstable process and is easily disturbed by noise, even a small amount of noise will lead to severe distortions in the estimation. This phenomenon is called ill-posedness. Many researches have been done to deal with the ill-posed problem including classical filtering methods \cite{Wiener1950filter,Richardson1972,Lucy1974} and regularization based methods \cite{TV1992,Yuan2008nonblind,Levin2007coded,nips2009sparse,Zoran2011patch_restoration,Xu2014FastDeconvolution}. In the regularization based methods, regularization terms are usually proposed based on the prior of the latent sharp image, such as Total Variation (TV) prior \cite{TV1992} or sparse priors \cite{Levin2007coded,nips2009sparse}. By combining an $l_2$-norm or an $l_1$-norm \cite{Bar2006IJCV} constrained data fidelity term with regularization terms, image deblurring can be modeled as an optimization problem. A latent sharp image can be recovered by solving the optimization problem. Besides Gaussian noise, non-blind deblurring is also disturbed by outliers, \emph{e.g.}, pixel saturation. Interested readers can refer to \cite{Harmeling2010ICIP,cho2011outlier_nonblind,Whyte2014nonblind} for more details.

\subsection{Blind Image Deblurring}
For blind image deblurring, the blur kernel is unknown and the problem is to recover both the latent sharp image and the blur kernel from only a blurry observation. The blind image deblurring is much more challenging than the non-blind problem because the solution for the problem is not only unstable but also non-unique.

In order to deal with this highly ill-posed blind image deblurring problem, there have been many pioneering works which attempted to approximate the problem in different perspectives. In early works, blind image deblurring has been processed like image enhancing. Image diffusion and shock filtering methods \cite{Osher1990shock,Alvarez1994shock,Gilboa2002shock} were proposed to enhance the edges of blurry images blurred by out-of-focus or Gaussian kernel. Another kind of approximation is to introduce parameterized forms of blur kernels or to impose specific constraints, such as centrosymmetry, on the blur kernels \cite{Pavlovic1992Maximum,Wei1996parameter,Yitzhaky1998Parameter,chan1998TV_deblur,YuLi1999Blind}. These methods can handle specific blur kernels, but cannot handle general motion blur which is usually irregular. There are also numerous works focusing on the multi-image deblurring \cite{Bascle1996images,Nayar2004images,Sroubek2005images,Yuan2007deblur_noise,cho2007images,Jia2008images}. Multiple images can provide more data constraints for kernel estimation and help to solve the blind image deblurring problem. Nevertheless, multiple relevant images in their assumptions are sometimes unavailable, which limits their practicability.

Recently, with the fast progress of regularization and optimization theory, many sophisticated image priors \cite{Fergus06,Levin2011EML,Krishnan2011,cai2012framelet,xuli2013Unnatural,Pan2016Pami,Lai2005color,Ren2016Low,Pan_2016_CVPR,Bai2019TIP} were proposed to handle the single image blind image deblurring problem with general blur kernels, such as the mixture of Gaussians prior that fits the heavy-tailed prior of natural images \cite{Fergus06,Levin2011EML}, normalized sparse prior \cite{Krishnan2011}, framelet based prior \cite{cai2012framelet}, $l_0$-norm based priors \cite{Pan2016Pami,xuli2013Unnatural}, color line prior \cite{Lai2005color}, dark channel prior \cite{Pan_2016_CVPR}, low rank prior \cite{Ren2016Low} and graph based prior \cite{Bai2019TIP}, \emph{etc}. The priors promote image sharpness and penalize image blurriness, which work as a regularizer in the optimization model guiding the solver to converge to the latent sharp image. Nevertheless, these image priors are usually non-convex, which result in very computationally expensive optimization algorithms.

\section{Observation and Multi-scale Latent Structure Prior}
\label{sec:convergence}
In this section, we introduce in detail the proposed multi-scale latent structure prior.

\subsection{Observation and Theoretical Analysis}
Michaeli and Irani \cite{Michaeli2014} first analyzed the effect of image down-sampling in the blind image deblurring problem and found that image down-sampling increased the internal patch recurrence.
Differently, in this work, we focus on the extreme case of image down-sampling and down-sample a blurry image to a very coarse level.
We observe that \textit{a coarse enough image down-sampled from a blurry observation is approximately a low-resolution version of the latent sharp image}.
We provide a theoretical proof on 2D signals to verify this observation under a general down-sampling operation. In the proof, we assume a noise-free blurred image. We empirically demonstrate that the observation is also robust with noise, by deblurring noisy and blurry images in the experiments in Section~\ref{sec:experiments}.

\begin{claim}
    {A coarse enough image down-sampled from a blurry observation is approximately a low-resolution version of the latent sharp image.}
\end{claim}
{\itshape Proof:} According to the degradation model of (\ref{eq:blur_model_x}), a noise-free blurry image $\b$ can be represented as
\begin{equation}
    \b[i,j]=\k\otimes\x=\iint \k(u,v)\x(i-u,j-v)dudv,
    \label{eq:blur_continuous}
\end{equation}
where $\x$ is the latent sharp image and $\k$ is the blur kernel.

When $\b$ is $\alpha$-times down-sampled, its down-sampled version is given by
\begin{align}
    \b_\alpha[i, j]&=\left(\b\otimes\h\right)\downarrow=\left(\k\otimes\x\otimes\h\right)\downarrow \notag \\
    &= \left(\k\otimes(\x\otimes\h)\right)\downarrow= \left(\k\otimes\x_h\right)\downarrow\notag\\
    &=\iint \k(u,v)\x_h(\alpha i-u,\alpha j-v)dudv,
    \label{eq:blur_continuous_downsample}
\end{align}
where $\b_\alpha[i,j]$ is a down-sampled blurry image, $\h$ is a general low-pass filter to avoid aliasing, $\x_h$ is a filtered band-limited image, $\downarrow$ represents the pixel extraction process and $\alpha$ is the down-sampling factor. By replacing $u$, $v$ with $\alpha t$, $\alpha s$, (\ref{eq:blur_continuous_downsample}) can be further written as
\begin{align}
    \b_\alpha[i,j]&=\iint \alpha^2\k(\alpha t, \alpha s)\cdot \x_h(\alpha i-\alpha t,\alpha j-\alpha s) dtds \notag \\
    &= \iint \k_\alpha(t,s)\cdot \x_\alpha(i-t,j-s)dtds.
    \label{eq:blur_continuous_downsample2}
\end{align}
where $\x_\alpha$ and $\k_\alpha$ are the down-sampled sharp image and the down-sampled blur kernel, respectively. From (\ref{eq:blur_continuous_downsample2}), the down-sampled blur kernel is:
\begin{equation}
    \k_\alpha(u,v)=\alpha^2 \k(\alpha u, \alpha v).
    \label{eq:blur_continuous_downsample3}
\end{equation}
The normalization condition $\iint \k_\alpha = 1$ is satisfied if the assumption $\iint \k=1$ is imposed.
It follows that the support of down-sampled blur kernel $\k_\alpha$ is $\alpha$ times smaller than that of blur kernel $\k$ in both horizontal and vertical directions.

\begin{figure}[!t]
\centering
\subfloat[]{
\label{fig:spatial_domain}
\includegraphics[width=0.48\linewidth]{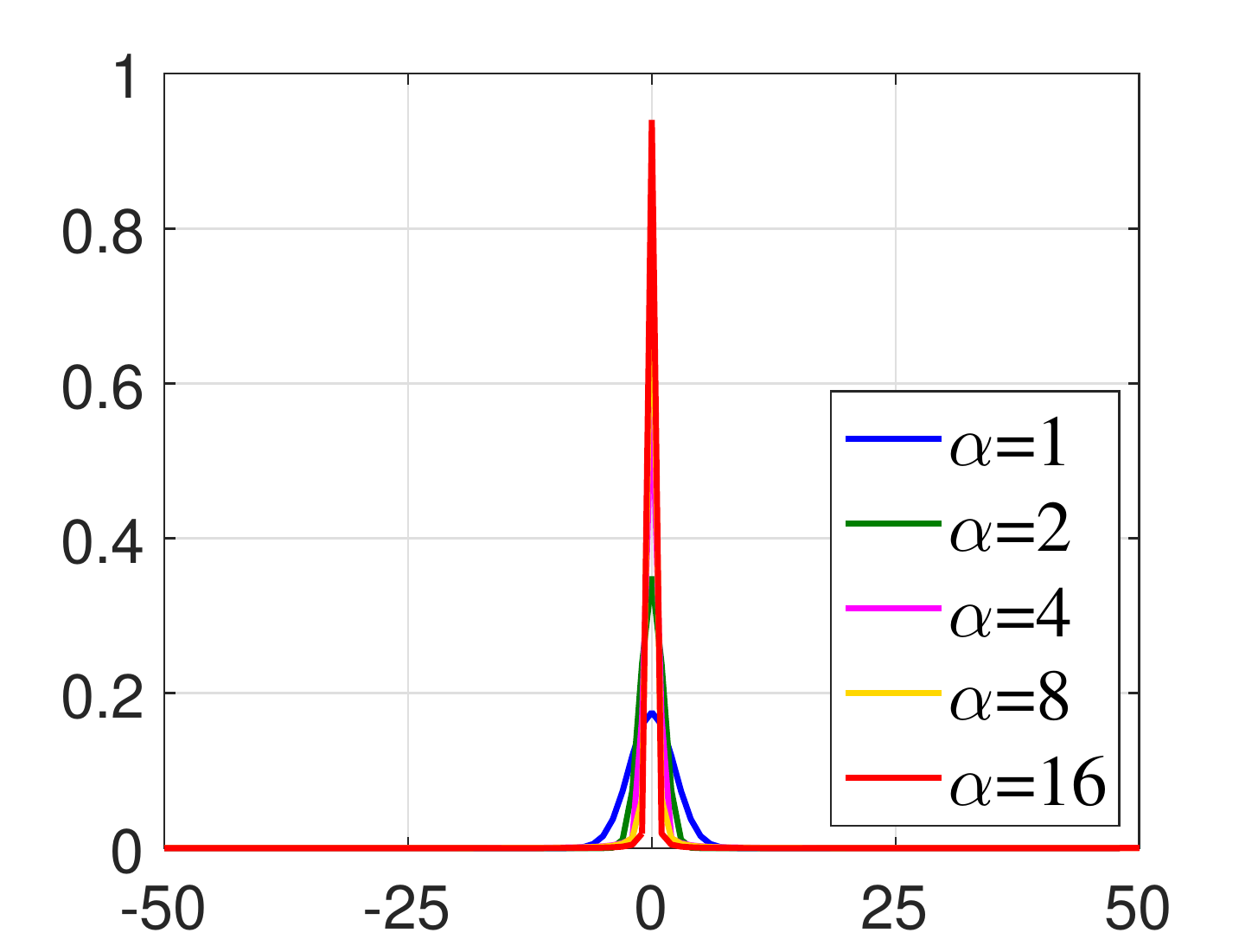}}
\subfloat[]{
\label{fig:frequency_domain}
\includegraphics[width=0.48\linewidth]{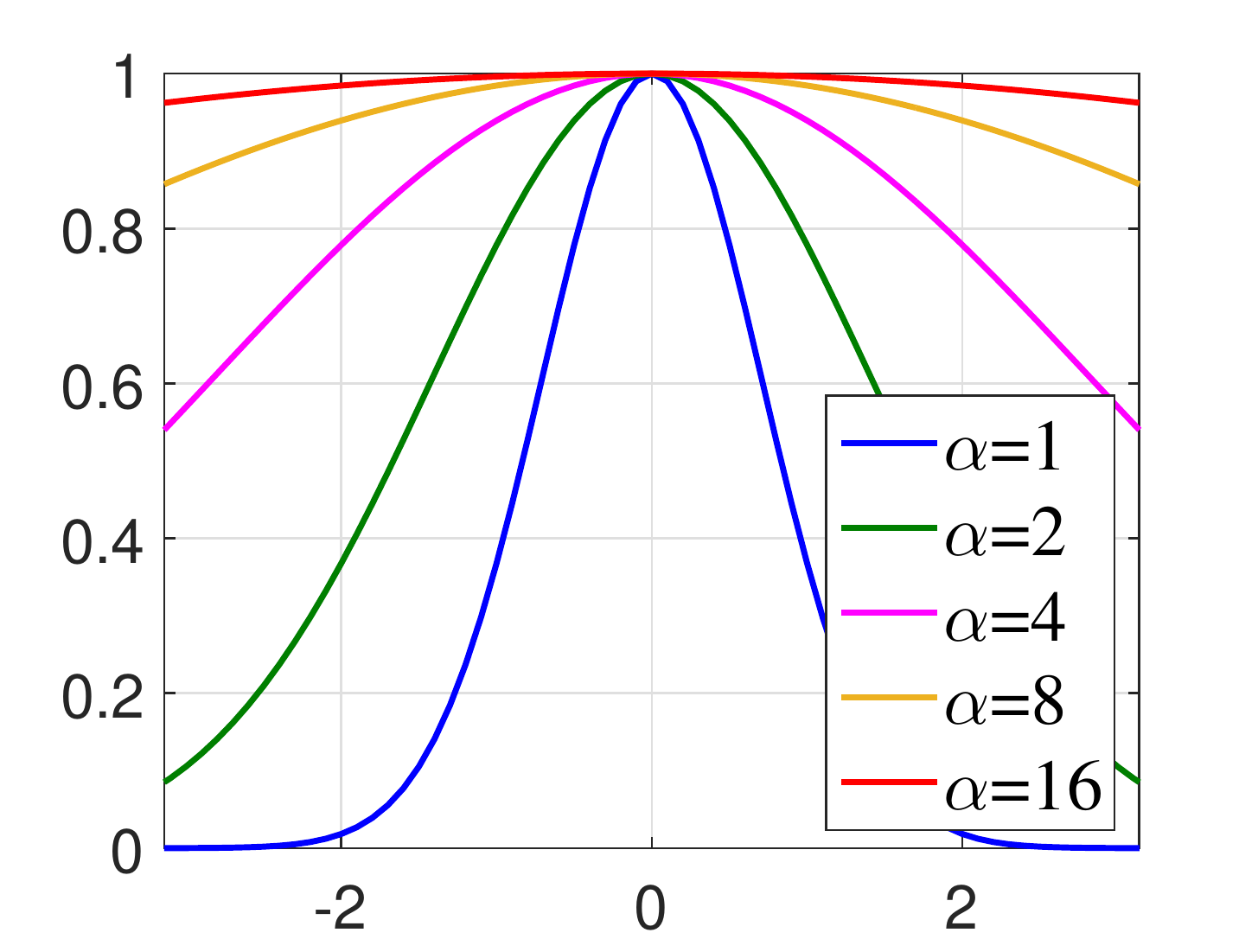}}
\caption{1D illustration of down-sampling a discrete gaussian kernel in the spatial and frequency domain. (a) spatial domain. (b) frequency domain in $[-\pi, \pi]$.}
\label{fig:frequency_spatial}
\end{figure}

We further transform $\k_\alpha$ into the frequency domain:
\begin{align}
    \K_\alpha(\tu,\tv)&=\cF\left(\k_\alpha(u,v)\right)=\alpha^2\cF\left(\k(\alpha u, \alpha v)\right) \notag \\
    &=\alpha^2 \left( \frac{1}{\alpha^2}\K(\tu/\alpha, \tv/\alpha)\right) \notag \\
    &=\K(\tu/\alpha, \tv/\alpha)
    \label{eq:blur_k_frequency}
\end{align}
where $\cF(\cdot)$ denotes the Fourier transform and $\K(\tu,\tv)=\cF(\k(u,v))$. $\K_\alpha$ is expanded by $\alpha$ in the frequency domain. Though $\K_\alpha$ is expanded, $\x_h$ and $\x_\alpha$ are low-pass filtered and band-limited. Thus, aliasing is avoided while down-sampling but only the information within the band-limit is preserved.
As the down-sampling becomes deeper, \emph{i.e.}, $\alpha$ is larger, $\K_\alpha$ in the band-limit tends to constant $1$ and its corresponding spatial response will converges to Dirac delta function $\delta$ \cite{stein_weiss_1971}.

The proof so far works in the continuous domain, \emph{i.e.}, $\k$ and $\x$ are two continuous signals. For discrete signals in practise, we sample the continuous signals and do normalization on $\k$ afterwards to ensure $\sum_{uv} \k_\alpha(u,v) =1$. Approximately, $\k_\alpha(u,v)$ will converge to a delta kernel in the discrete domain:
\begin{equation}
    \delta(i,j)=\begin{cases}
    1, & i=0\ and\ j=0,\\
    0, & i\ne0\ or\ j\ne0.
    \end{cases}
\end{equation}
which satisfies $\sum_i\sum_j \delta(i,j)=1$. An illustrative experiment in the discrete domain is shown in Fig.~1.
Finally, we have
\begin{align}
    \b_\alpha[i,j] \to \x_\alpha \otimes \delta =\x_\alpha,
    \label{eq:blur_continuous_delta}
\end{align}
\emph{i.e.}, the down-sampled blurry image converges to the low-resolution version of the latent sharp image.

\subsection{Multi-Scale Latent Structure Prior}
\label{sec:mlsp}
\textit{Claim~1} is very useful for blind image deblurring, since it gets rid of the effect of the unknown blur kernel and provides a low-resolution prior of the latent sharp image, referred as the \textit{latent structure} prior in our algorithm.

Based on \textit{Claim~1}, we further introduce a multi-scale strategy to work in tandem with latent structure prior on an image pyramid.
Starting from the coarsest scale, the prior image is employed to blindly restore a finer scale sharp image.
The newly restored sharp image is then employed as the latent structure prior of the next finer scale.
The strategy works gradually from the coarsest scale to the finest scale, thus we name the coarse-to-fine prior---\textit{Multi-Scale Latent Structure} (MSLS) prior.
In the next section, we will elaborate the blind image deblurring algorithm tailored to the proposed MSLS prior.

\section{Blind Image Deblurring Algorithm}
\label{sec:formulation_algorithm}

In this section, we introduce a joint optimization scheme to estimate both the latent sharp image and the blur kernel based on the proposed MSLS prior. The algorithm comprises two phases, \emph{i.e.}, preliminary restoration in coarse scales in Sec.\;\ref{sec:kernel_estimation} and refined restoration in the finest scale in Sec.\;\ref{sec:k_refine}. We further extend the proposed algorithm to solve non-uniform blind image deblurring problem in Sec.\;\ref{sebsec:non-uniform}.

\subsection{Preliminary Restoration in Coarse Scales}
\label{sec:kernel_estimation}
Blind image deblurring is modeled to solve the following optimization problem,
\begin{equation}
    \argmin_{\k,\x} \Phi(\k\otimes\x-\b)
    +\lambda_1\Psi_1(\k)+\lambda_2\Psi_2(\x)
    \label{eq:bid_regularization}
\end{equation}
where $\Phi(\k\otimes\x-\b)$ is the data fidelity term, $\Psi_1(\k)$ and $\Psi_2(\x)$ are the regularizers of blur kernel $\k$ and latent sharp image $\x$, respectively.
Previous methods, such as \cite{Fergus06,Levin2011EML,Krishnan2011,cai2012framelet,xuli2013Unnatural,Pan2016Pami,Lai2005color,Pan_2016_CVPR,Ren2016Low,Michaeli2014,Bai2019TIP}, introduced sophisticated image priors, which are usually either non-convex or computationally expensive.

To tackle the challenging problem mentioned above, we take advantage of the proposed MSLS prior and introduce an efficient local self-example matching strategy to substitute for complex non-convex regularization of $\x$ in sharp image reconstruction.
In each coarse scale, our algorithm comprises three steps, as sketched in Algorithm~1.

\begin{algorithm}[htb]
\label{al:1}
\caption{Preliminary Restoration in Each Coarse Scale}
\begin{algorithmic}[1] %
\small
\REQUIRE  
Blurry image $\b$ in current scale, prior image $\x_{pr}$.\\
~~~  Kernel size $h\times h$.
\ENSURE  
Estimated kernel $\hk$ and the latent sharp image $\x_{l}$.\\
\STATE Initialize latent image $\x_{l} = \b$ and $\hk = \delta$. \\
\STATE
\textbf{for} \textit{iter} $=$ $1 \rightarrow max\_iteration$ \textbf{do}\\
\ \ \ \ \ \ (a)\ Estimate $\hx$ using $\x_{l}$ and $\x_{pr}$.\\
\ \ \ \ \ \ (b)\ Minimize (\ref{eq:op_k_xc}) to estimate $\hk$, given $\hx$.\\
\ \ \ \ \ \ (c)\ Minimize (\ref{eq:op_tv}) to update $\x_{l}$, given $\hk$.\\
\textbf{endfor}\\

\RETURN $\hk$ and $\x_{l}$; %
\end{algorithmic}
\end{algorithm}

Specifically, in step (a), we blindly reconstruct a sharp image with fast local self-example matching. Step (b) and (c) correspond to a novel kernel estimation with error compensation and a non-blind image deblurring, which are both convex optimization and can be efficiently solved with \textit{Fast Fourier Transform} (FFT) acceleration. The prior image is first initialized as the coarsest scale image $\x_\alpha$. After solving one scale, we update the prior image with the newly deblurred result and continue to restore the next finer scale. The diagram of preliminary restoration is illustrated in Fig.\;\ref{fig:priliminary}.
In the following, we introduce in detail these three steps:
\medskip

\begin{figure}[!t]
\centering
\includegraphics[width=0.9\linewidth]{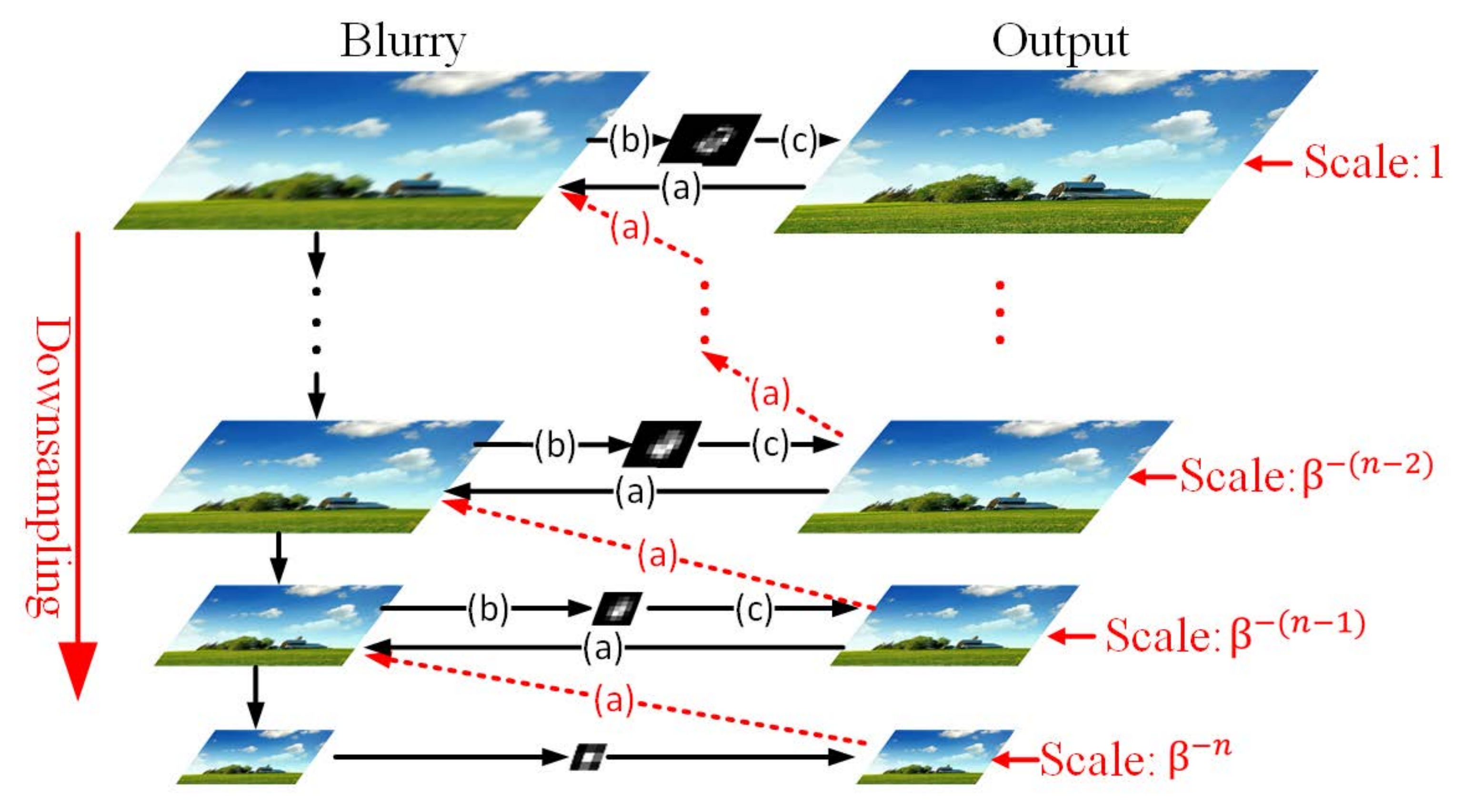}
\caption{The diagram of preliminary restoration in coarse scales. (a) sharp image reconstruction. (b) kernel estimation. (c) non-blind deblurring. We use scale factor $\beta=\log_23$ for multi-scale image pyramid construction. In the coarsest scale, $\lfloor\beta^n\rfloor= \alpha$ leads to a low-resolution sharp image.}
\label{fig:priliminary}
\end{figure}

\begin{figure}[!t]
\centering
\includegraphics[width=0.9\linewidth]{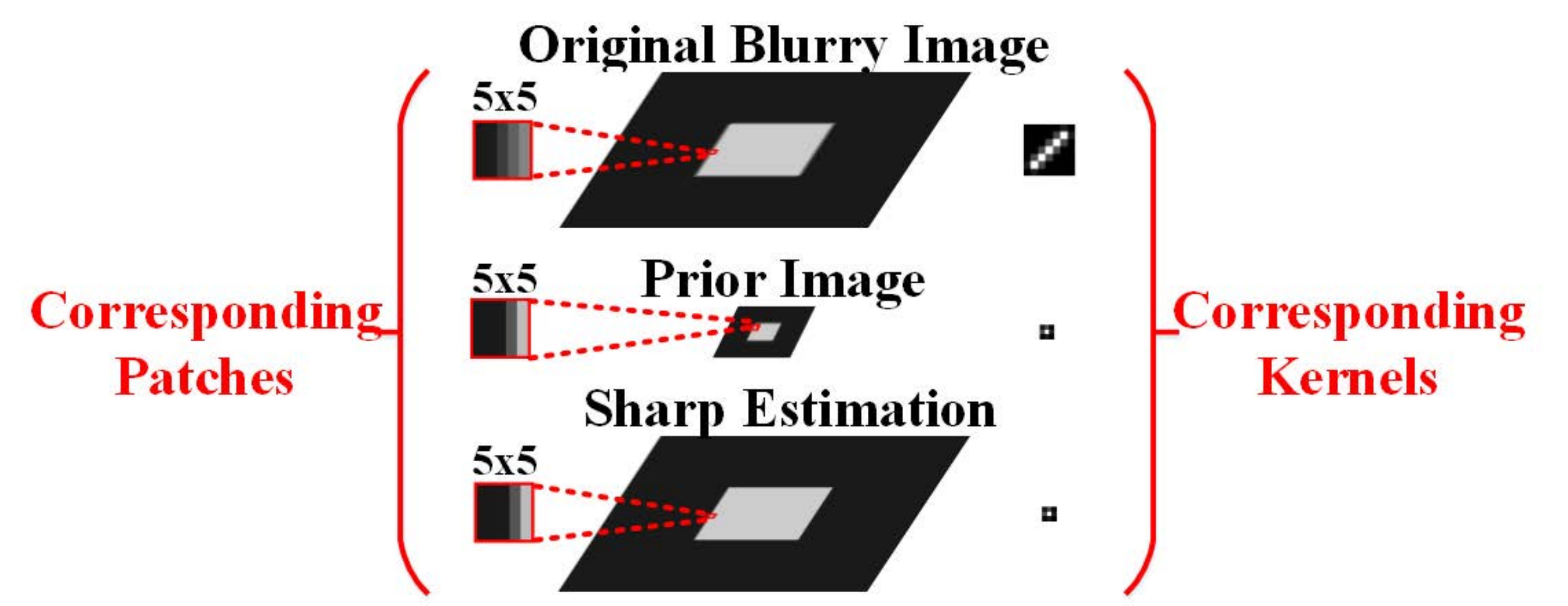}
\caption{A single-scale illustrative experiment of sharp image reconstruction using local self-example matching strategy. The latent sharp image is blurred by a 7$\times$7 motion blur kernel. The blurry image is down-sampled until the blur kernel is approximately a delta function. According to Claim 1, the coarse enough image is a prior of the latent sharp image. The corresponding kernels are shown on the right. Patches in the blurry image find their NN patches in corresponding local areas of the prior image. An example of matched patches is shown on the left. All sharp NN patches are fused together to reconstruct the sharp image.}
\label{fig:prior_ex}
\end{figure}

{\noindent \bfseries Step (a): Sharp Image Reconstruction with Local Self-Example Matching:}
The finer scale latent sharp image is reconstructed with a fast local self-example matching strategy, inspired by \cite{FreFat10}. The latent sharp image $\x_l$ is first initialized as the blurry image $\b$.
We divide the latent image $\x_l$ into overlapped small patches (usually of size $5\times5$ pixels with $50\%$ overlap). Then, each patch $i$ finds its nearest neighbor (NN) in the corresponding local area of the coarser-scale sharp prior image $\x_{pr}$, measured by (\ref{eq:distance}),
\begin{align}
    d(i,j)=\|\mathbf{P}_i\x_l-\mathbf{P}_j\x_{pr}\|_2,\ \ j\in \mathcal{N}(i{'})
    \label{eq:distance}
\end{align}
where $\mathbf{P}_i$ is the patch extraction operation at position $i$. $i{'}$ is a position on the sharp prior image $\x_{pr}$. If $\x_l$ is down-sampled to the size of $\x_{pr}$, the position $i$ on $\x_l$ will be projected on the position $i'$ on $\x_{pr}$. $\mathcal{N}(i{'})$ represents the local neighborhood of the position $i{'}$. $d(i,j)$ denotes the distance between the patch $i$ and the patch $j$.
The prior image $\x_{pr}$ is $\beta$-times smaller than $\x_l$, as shown in Fig.~\ref{fig:priliminary}.

Finally, we fuse the searched NN patches together by computing weighted average in the overlapped areas, in order to avoid blocking artifacts. The weights are defined by Hamming window\footnote{https://en.wikipedia.org/wiki/Window\_function}. The 2D Hamming window function is defined as follows:
\begin{align}
    w(i)=\theta&-\gamma\cos\left(\frac{2\pi i}{N-1}\right) \\
    W(i,j)&=w(i)\cdot w(j)
    \label{eq:hamming}
\end{align}
where $\theta=0.54$ and $\gamma=0.46$ are fixed parameters, $N$ is the size of window, $w$ and $W$ are 1D and 2D weights of Hamming window, $i$ and $j$ are the indices. An illustrative example of a single-scale sharp image reconstruction is shown in Fig.~\ref{fig:prior_ex}.

In our algorithm, \textit{local patch matching} works not as an approximation of the global patch matching but a better choice for sharp image reconstruction, which avoids the patch over-fitting and considerably reduces the computational complexity of the NN search:
\subsubsection{Avoid Patch Over-fitting}
\begin{figure}[!t]
\centering
\subfloat[]{
\label{fig:local_nonlocal_ex1}
\includegraphics[width=0.75\linewidth]{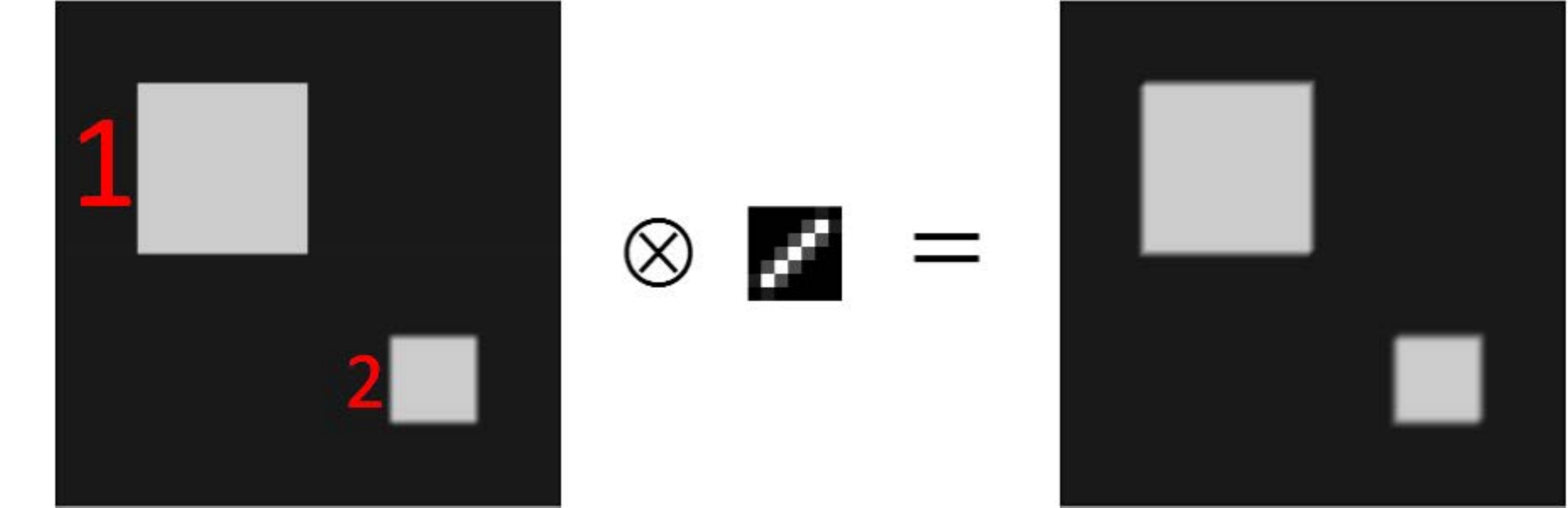}}\\
\subfloat[]{
\label{fig:local_nonlocal_ex2}
\includegraphics[width=0.75\linewidth]{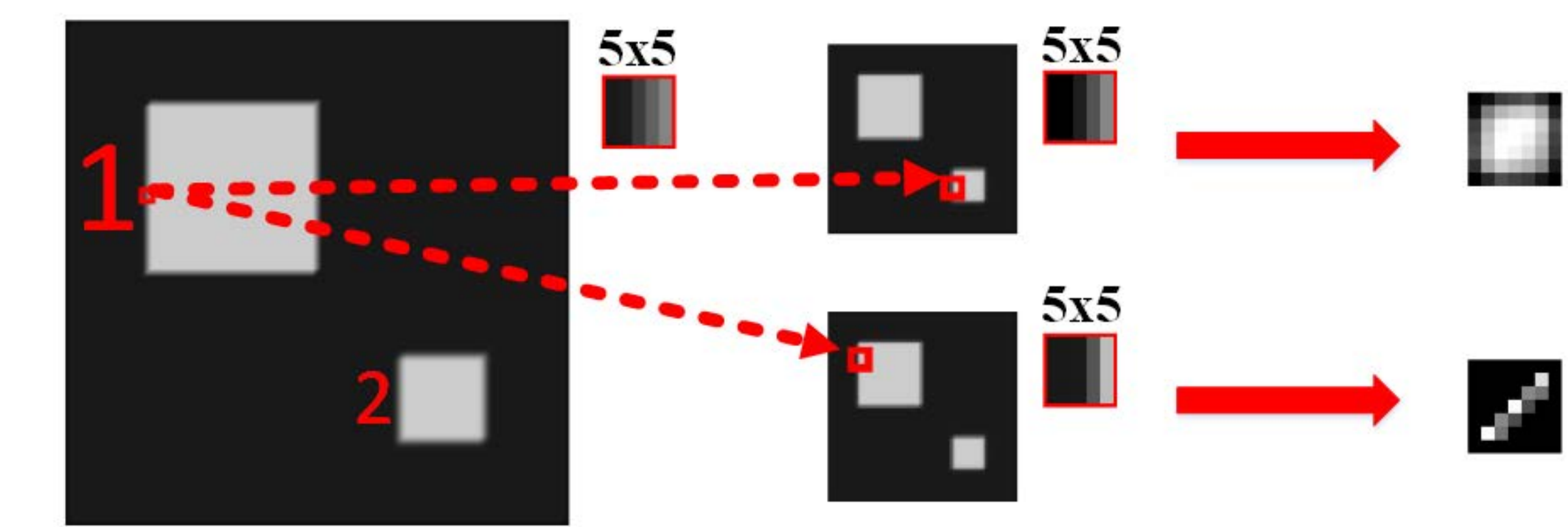}}
\caption{An single-scale illustrative experiment to demonstrate the efficacy of {\itshape locality}. (a) Square 1 is in the sharp foreground and square 2 is in the blurry background that is first blurred by a Gaussian kernel with $\sigma=2$. Then, the image is blurred by a 7$\times$7 motion blur kernel. (b) The patch in the blurry image finds its NN patch in the prior image. For the global patch matching, the NN patch is in the background square 2, which is an over-fitting blurry patch. For the local patch matching, the NN patch is in the foreground square 1, which is a sharp patch. The local searching method leads to a more appropriate kernel estimation. }
\label{fig:local_nonlocal_ex}
\end{figure}

In motion blurred images with different depths of field, the blurry patch in the foreground tends to find an over-fitting blurry NN in the background area of the prior image, if the searching distance is not constrained. This misleads the sharp image reconstruction and causes the incorrect kernel estimation in the followed step. We report a single-scale illustrative experiment in Fig.~\ref{fig:local_nonlocal_ex}. When the search range is constrained by the locality property, more wanted sharp patches are found and the kernel estimation result is also improved.

\subsubsection{Accelerate Searching}
The computational complexity of local patch matching is $O(C)$, where $C$ is a constant that accounts for the search range. The search range is set to $10\times10$ for $5\times5$ sized patches in our algorithm.
There are $(10-2\times\lfloor\frac52\rfloor)\times(10-2\times\lfloor\frac52\rfloor)=36$ times \textit{mean square error} (MSE) computations at each location.

{\noindent \bfseries Step (b):  Kernel Estimation with Error Compensation:}
Given the sharp image reconstructed with MSLS prior, we estimate the blur kernel afterwards.
Since the finer-scale sharp image is reconstructed from the coarser-scale prior image, there is an inevitable error between the reconstructed image and the ground-truth.

In this step, we propose a new image blur model (\ref{eq:blur_model_hx_xc}) for our kernel estimation, in order to compensate for the error:
\begin{equation}
    \b=\k\otimes(\hx+\x_c)+\n.
    \label{eq:blur_model_hx_xc}
\end{equation}
where $\hx$ is the reconstructed image and $\x_c$ is the compensatory layer. With the adjustable $\x_c$ layer, $\hx+\x_c$ can be equivalent to the latent sharp image $\x$ and the estimated kernel $\hk$ can be more accurate.

Given $\hx$ and model (\ref{eq:blur_model_hx_xc}), computing $\hk$ and $\x_c$ results in solving an optimization problem (\ref{eq:op_k_xc}). The model (\ref{eq:blur_model_hx_xc}) is used as a fidelity term to constrain that the $\hk$ and $\hx_c$ follow the image convolution process. Instead of directly using (\ref{eq:blur_model_hx_xc}), we transform it into gradient domain, in order to reduce ringing artifacts \cite{cho_lee}. The regularizer of $\k$ is penalized by an $l_2$-norm and the regularizer of $\x_c$ is penalized by an $l_1$-norm to ensure the spatial sparsity of $\x_c$:
\begin{align}
    \argmin_{\k,\x_c} \frac12\|\k\otimes\nabla(\hx+\x_c)-\nabla\b\|_2^2
    +\lambda_1\|\k\|_2^2+\lambda_2\|\x_c\|_1
    \label{eq:op_k_xc}
\end{align}
where $\lambda_1$ and $\lambda_2$ are two trade-off parameters to control the regularization strength.

At first glance, (\ref{eq:op_k_xc}) is a non-convex optimization problem, since there exists a $\k\otimes\nabla\x_c$. Here we apply a variable substitution to make it convex, \emph{i.e.}, replacing $\k\otimes\nabla\x_c$ with a new variable $\v$. Note that $\x_c$ is assumed sparse and the size of $\k$ is much smaller than the size of $\x_c$, so $\k\otimes\nabla\x_c$ can also be considered sparse. We thus substitute $\|\v\|_1$ for $\|\x_c\|_1$. Now the objective function (\ref{eq:op_k_xc}) can be reformulated as a convex optimization problem (\ref{eq:op_k_v}):
\begin{align}
    \argmin_{\k,\v} \frac12\|\k\otimes\nabla\hx+\v-\nabla\b\|_2^2+\lambda_1\|\k\|_2^2+\lambda_2\|\v\|_1
    \label{eq:op_k_v}
\end{align}

We alternately compute $\k$ and $\v$ by fixing one and addressing another. The $\nabla = \{\nabla_x, \nabla_y\}$ are derivative operators in horizontal and vertical directions respectively. The $\v = \{\v_x, \v_y\}$ are the corresponding layers of horizontal and vertical directions and are initialized to zeros.

In (\ref{eq:op_k_v}), terms related to $\k$ are all quadratic, so a closed-form solution can be derived for $\k$. Moreover, to fast solve $\k$, we compute the solution in the frequency domain,
\begin{equation}
    \k=\mathcal{F}^{-1}\left(\frac{\overline{\mathcal{F}(\nabla\hx)}\mathcal{F}(\nabla\b)-\overline{\mathcal{F}(\nabla\hx)}\mathcal{F}(\v)}
    {\overline{\mathcal{F}(\nabla\hx)}\mathcal{F}(\nabla\hx)+2\lambda_1}\right)
    \label{eq:op_k3}
\end{equation}
where $\mathcal{F}(\cdot)$ and $\mathcal{F}^{-1}(\cdot)$ are Fourier and inverse Fourier transform implemented by FFT. The $\overline{\mathcal{F}(\cdot)}$ means the complex conjugate of fourier transformed values.

As each element in $\v$ is independent, we use proximal mapping \cite{Parikh:2014:proximal} to solve the $l_1$-norm optimization problem. The solver works as a fast soft-thresholding:
\begin{equation}
    v_i=sgn(z_i)\cdot\max\left\{0,|z_i|-\lambda_2\right\}
    \label{eq:op_v2}
\end{equation}
where $i$ means the $i$-th pixel in $\v_x$ or $\v_y$, $z_i=\nabla b_i-(\k\otimes\nabla\hx)_i$.

\begin{figure}[!t]
\centering
\subfloat[]{
\label{fig:xc_blur}
\includegraphics[width=0.31\linewidth]{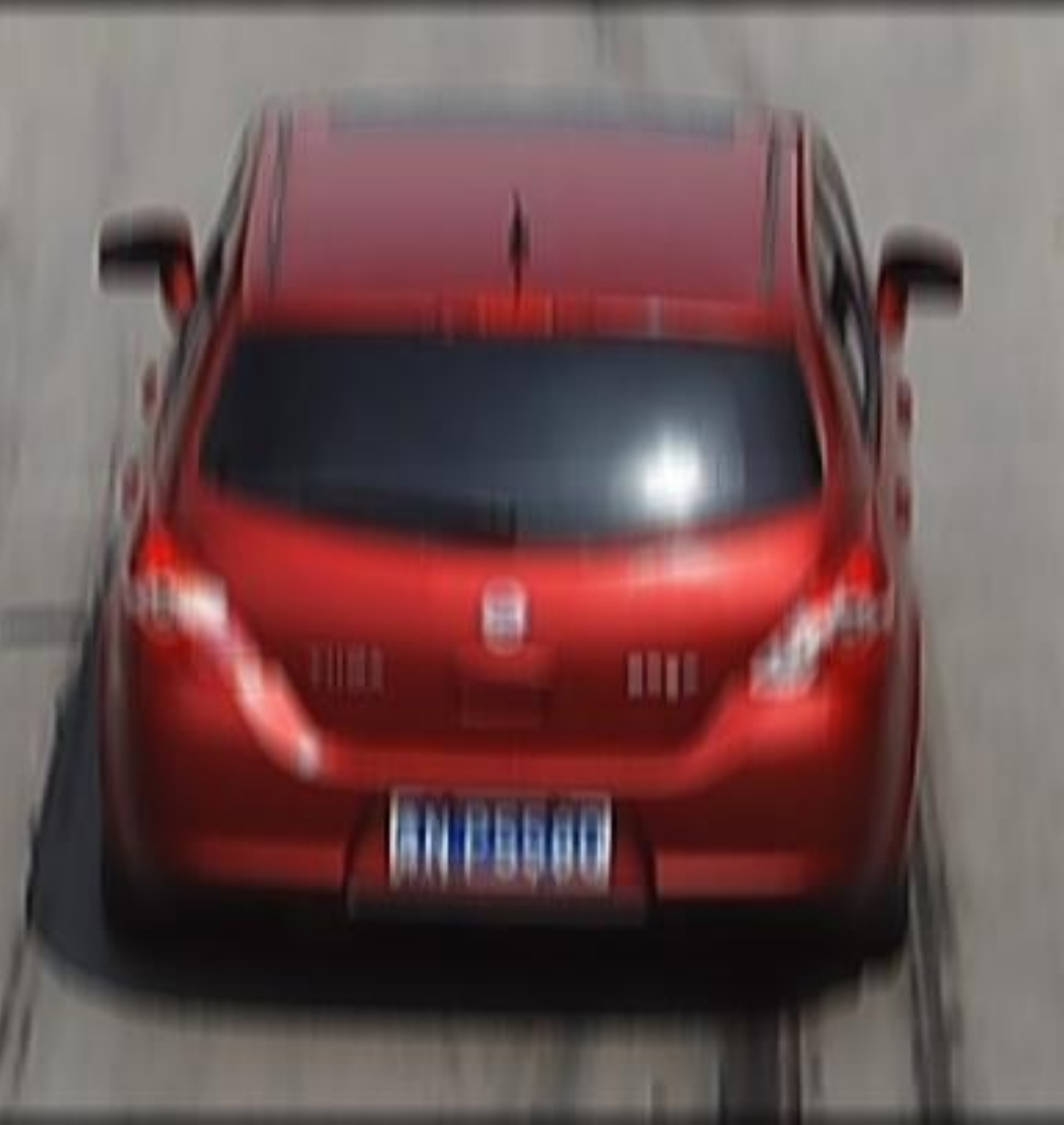}}
\subfloat[]{
\label{fig:xc_de_no}
\includegraphics[width=0.31\linewidth]{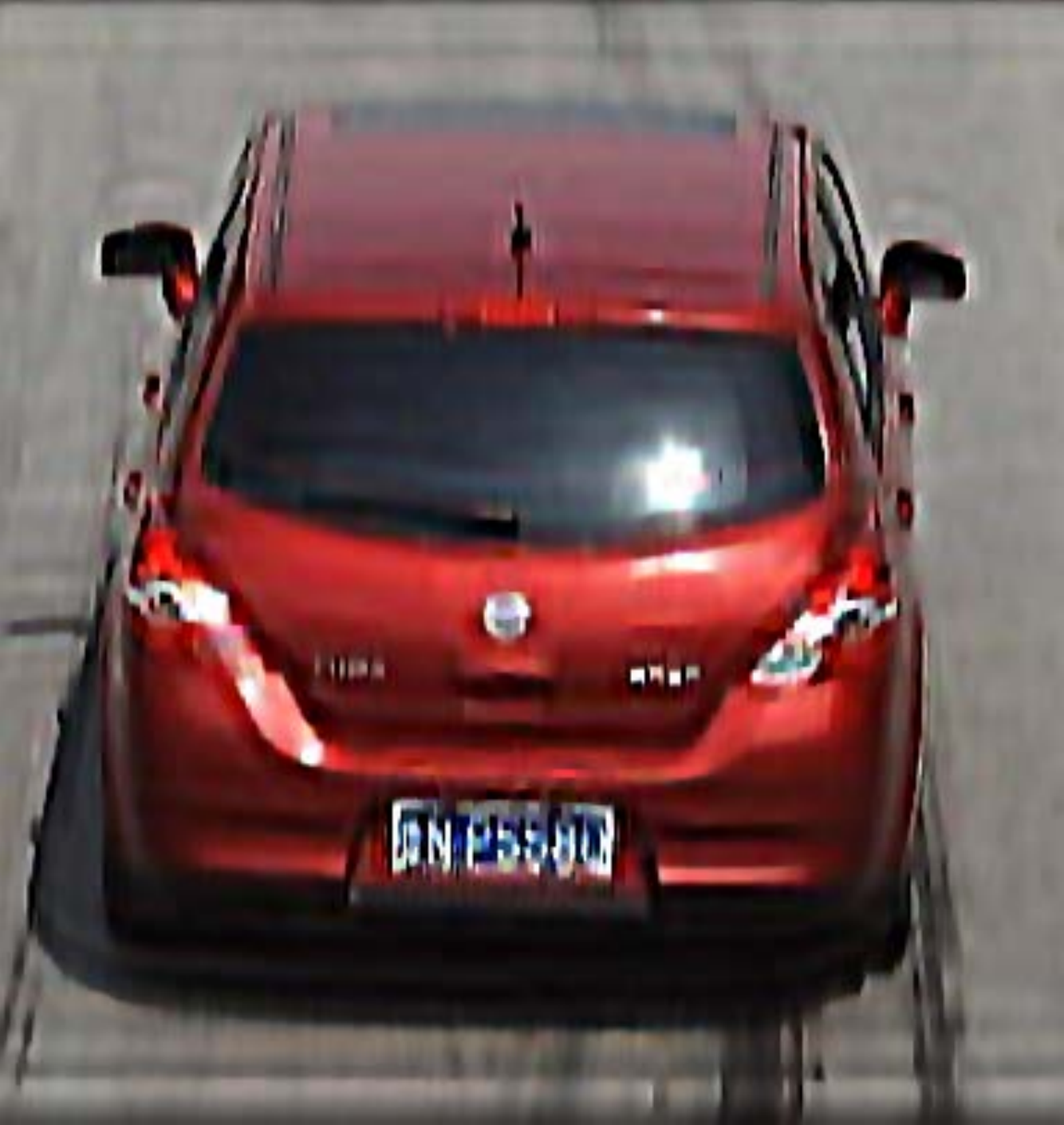}}
\subfloat[]{
\label{fig:xc_de}
\includegraphics[width=0.31\linewidth]{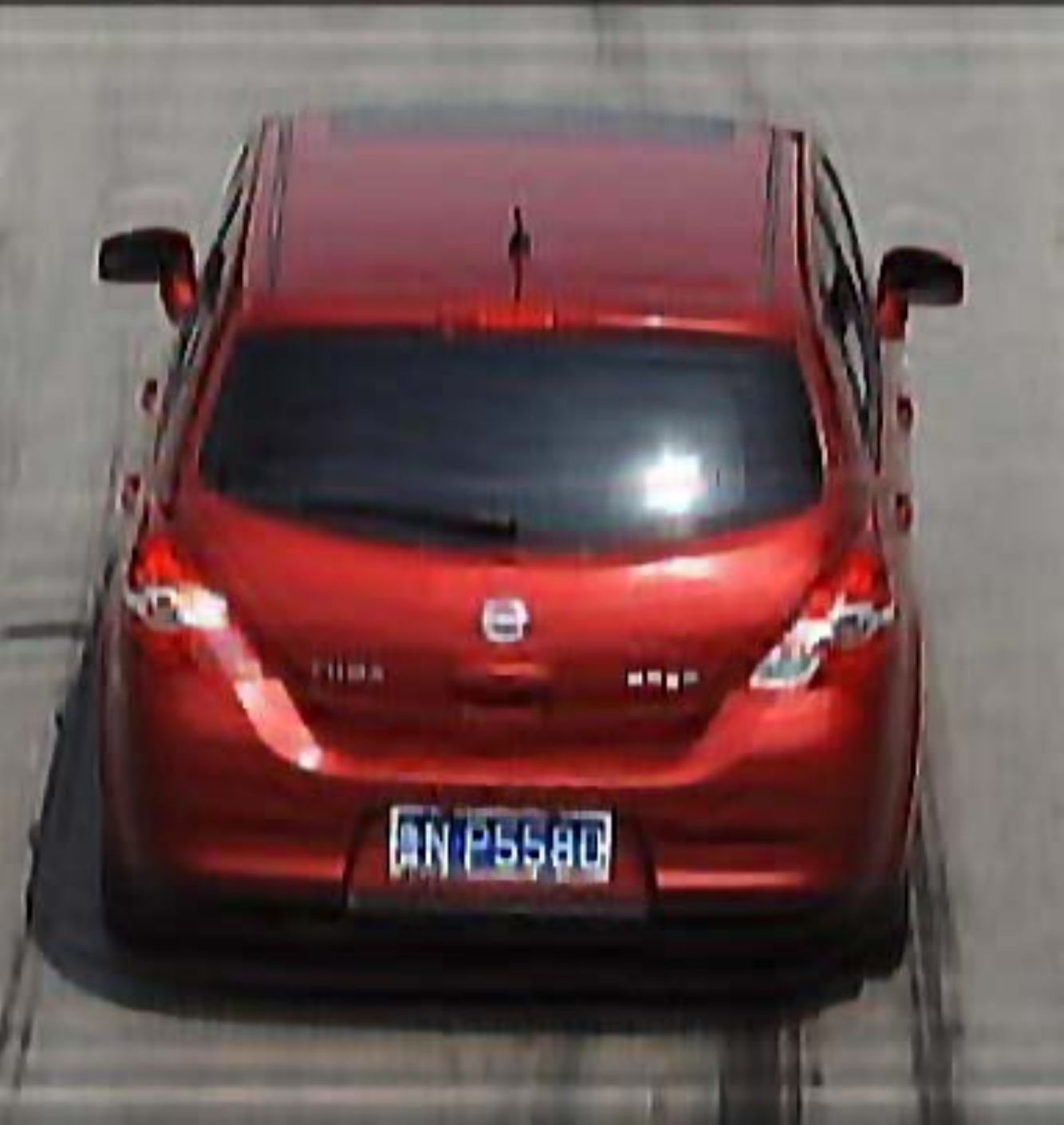}}\\
\subfloat[]{
\label{fig:xc_blur_cut}
\includegraphics[width=0.31\linewidth]{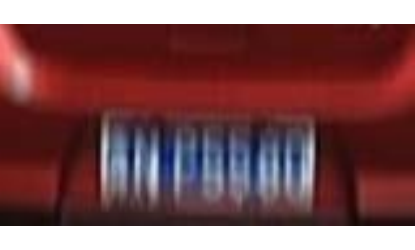}}
\subfloat[]{
\label{fig:xc_de_no_cut}
\includegraphics[width=0.31\linewidth]{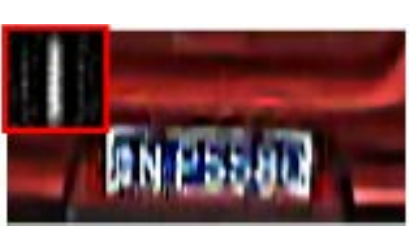}}
\subfloat[]{
\label{fig:xc_de_cut}
\includegraphics[width=0.31\linewidth]{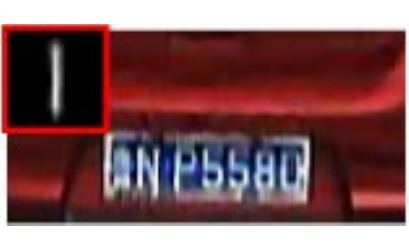}}\\
\subfloat[]{
\label{fig:xc_sharpWindow}
\includegraphics[width=0.31\linewidth]{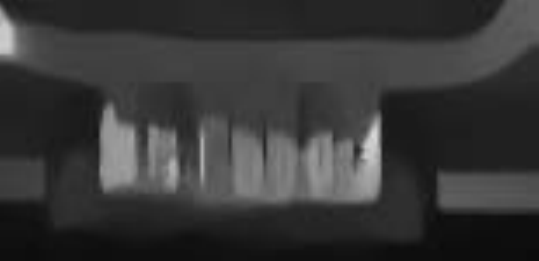}}
\subfloat[]{
\label{fig:xc_x}
\includegraphics[width=0.31\linewidth]{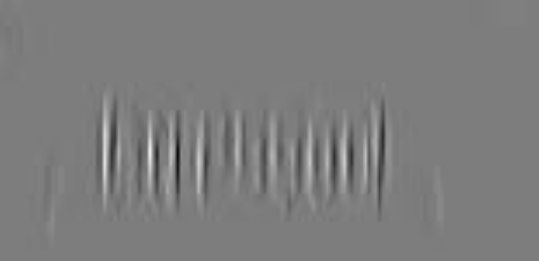}}
\subfloat[]{
\label{fig:xc_y}
\includegraphics[width=0.31\linewidth]{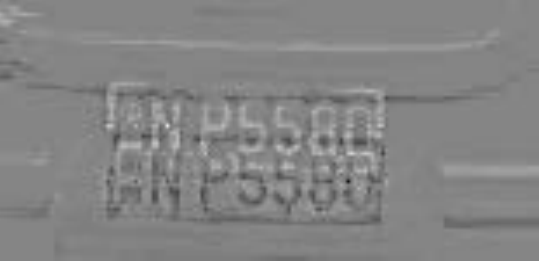}}
\caption{(a) A blurry image with size 384$\times$406 and kernel size 27$\times$27. (b) Deblurred result with only the MSLS prior. (c) Deblurred result with both the MSLS prior and the compensatory layer. (d) Close-up of the license plate in (a). (e) Close-up of the license plate in (b) and corresponding kernel estimation. (f) Close-up of the license plate in (c) and corresponding kernel estimation. (g) Close-up of the license plate in $\hx$. (h) Close-up of the license plate in $\v_x$. (i) Close-up of the license plate in $\v_y$.}
\label{fig:xc_cp}
\end{figure}

An illustrative deblurring example is illustrated in Fig.~\ref{fig:xc_cp}. Fig.~\ref{fig:xc_blur} is a blurry image. Fig.~\ref{fig:xc_de_no} and Fig.~\ref{fig:xc_de} show two blind deblurred results without and with the compensatory layer $\x_c$. Fig.~\ref{fig:xc_blur_cut}, Fig.~\ref{fig:xc_de_no_cut} and Fig.~\ref{fig:xc_de_cut} are their close-ups of the license plate, respectively. Solving with $\x_c$ achieves clearer deblurred result. The corresponding estimated kernel is also much better, in which less outliers appear beside the true trajectory.

We further analyze the role of the auxiliary variable $\v$ in optimization of (\ref{eq:op_k_v}). Fig.~\ref{fig:xc_sharpWindow} is the close-up of the license plate in $\hx$, which is reconstructed from the coarser-scale prior image. It can be found that $\hx$ contains sharp edges but loses some details, \emph{e.g.}, the license number. To overcome this issue, we introduce the $\x_c$ to compensate for the error. In this case, ideally, $\x_c$ should be the license number. To solve both $\k$ and $\x_c$ in (\ref{eq:op_k_xc}), we substitute a new variable $\v$ for $\k\otimes\nabla\x_c$ and solve (\ref{eq:op_k_v}). Fig.~\ref{fig:xc_x} and Fig.~\ref{fig:xc_y} are the $\v_x$ and $\v_y$ obtained by solving (\ref{eq:op_k_v}). It can be found that the restored $\v_x$ and $\v_y$ are the derivatives of the license number $\nabla\x_c$ convolving with the kernel $\k$, which demonstrates that the reconstructed result matches the theoretical analysis.

{\noindent \bfseries Step (c): Non-blind Image Deblurring:} After the kernel $\hk$ is computed, we conduct the non-blind deblurring to update the latent image $\x_l$. We apply the Total Variation (TV) regularized non-blind deblurring algorithm, which is formulated as follows:
\begin{align}
    \argmin_\x \frac12\|\hk\otimes\x-\b\|_2^2+\mu\|\nabla\x\|_1
    \label{eq:op_tv}
\end{align}
where $\mu$ is a regularization parameter. We apply the \textit{alternating direction method of multipliers} (ADMM) \cite{Boyd:2011:ADMM, Goldstein:2009TV_solver} to lead to a fast solver in the frequency domain.

\subsection{Refined Restoration in the Finest Scale}
\label{sec:k_refine}

\begin{figure}[!t]
\centering
\includegraphics[width=0.85\linewidth]{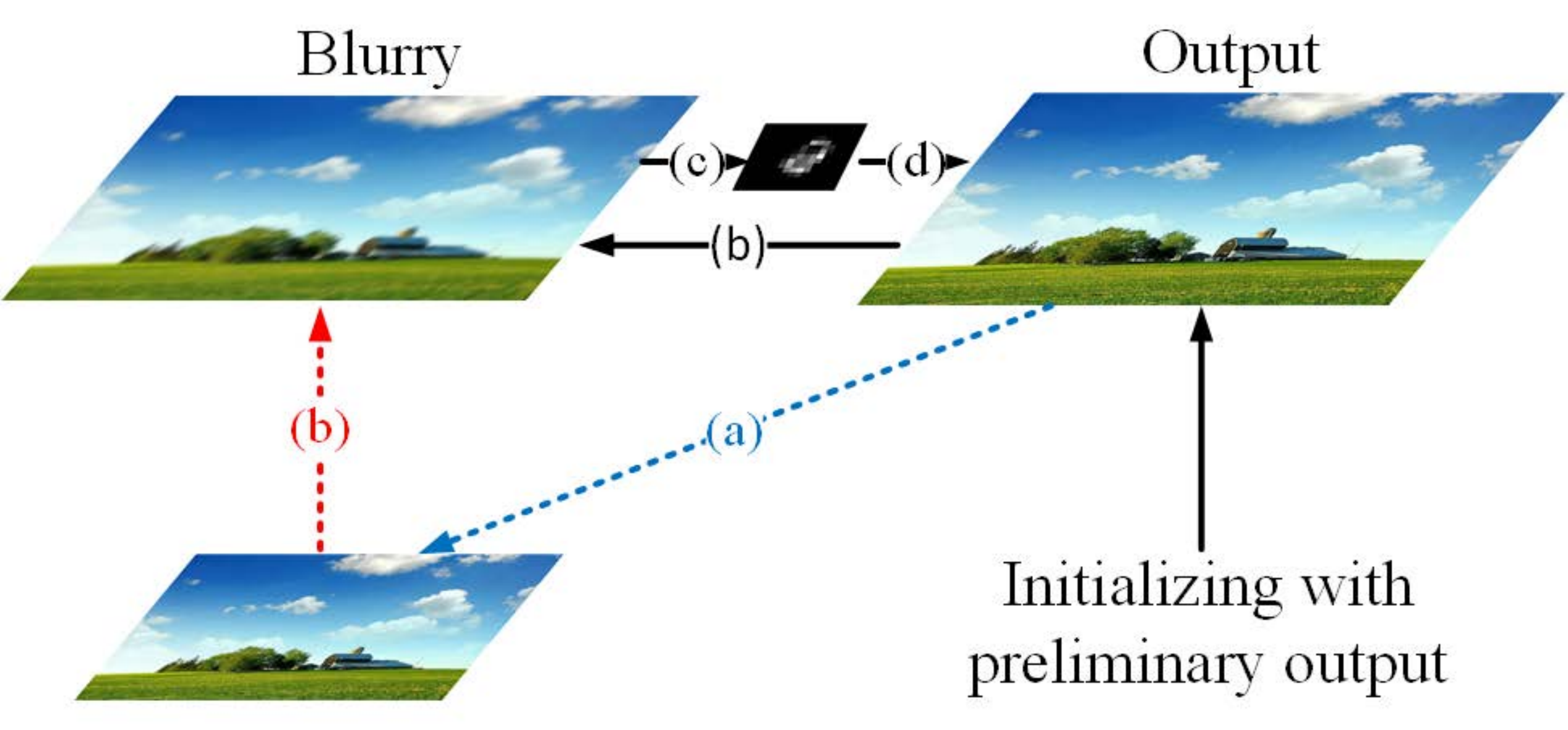}
\caption{The diagram of refined restoration in the finest scale. (a) prior update. (b) sharp image reconstruction. (c) kernel estimation. (d) non-blind deblurring.}
\label{fig:refine}
\end{figure}

Preliminary restoration can fast estimate sharp images in the coarse scales but possibly results in artifacts.
In order to achieve more accurate restoration for complex cases, we propose to perform a refined restoration in the finest scale, inspired by \cite{XuJia2010eccv}.
Given the preliminary restored image as initialization, the diagram of refined restoration in the finest scale is shown in Fig.\;\ref{fig:refine}.

Since the refined restoration is employed in the finest scale, there is no prior image updated from coarse scales. Thus, we update the prior image by filtering and down-sampling the restored image in each iteration. We use the edge-preserving guided filter \cite{Guided_filter} to reduce the possible artifacts without introducing any extra blur. Besides, we also employ higher-order derivatives in the data fidelity term to add more constraints for accurate kernel estimation \cite{hqdeblurring_siggraph2008}:
\begin{align}
    \argmin_{\k,\v} \frac12\|\k\otimes\nabla_*\hx+\v_*-&\nabla_*\b\|_2^2 \notag\\
    +&\lambda_3\|\k\|_2^2+\lambda_4\|\v_*\|_1.
    \label{eq:op_k_v_refine}
\end{align}
where $\nabla_*$$=$\{$\nabla_x$, $\nabla_y$, $\nabla_{xx}$, $\nabla_{yy}$, $\nabla_{xy}$\} and $\v_*$$=$\{$\v_x$, $\v_y$, $\v_{xx}$, $\v_{yy}$, $\v_{xy}$\}. The algorithm is outlined in Algorithm~2.

\begin{algorithm}[htb]
\label{al:2}
\caption{Refined Restoration in the Finest Scale}
\begin{algorithmic}[1] %
\small
\REQUIRE   
Blurry image $\b$ and preliminary estimation $\tilde{\x}$. \\
\ \ \ \ Kernel size $h\times h$.
\ENSURE    
Refined kernel $\hk$, the latent sharp image $\x_l$.\\
\STATE Initialize latent image $\x_l = \tilde{\x}$. \\
\STATE
\textbf{for} \textit{iter} $=$ $1 \rightarrow max\_iteration$ \textbf{do}\\
\ \ \ \ (a)\ Update prior $\x_{pr}$ by filtering and down-sampling $\x_l$.\\
\ \ \ \ (b)\ Estimate $\hx$ from $\x_l$ and $\x_{pr}$.\\
\ \ \ \ (c)\ Minimize (\ref{eq:op_k_v_refine}) to update $\hk$, given $\hx$.\\
\ \ \ \ (d)\ Minimize (\ref{eq:op_tv}) to update $\x_l$, given $\hk$.\\

\textbf{endfor}\\

\RETURN $\hk$, $\x_l$; %
\end{algorithmic}
\end{algorithm}

\begin{figure}[!t]
\centering
\subfloat[]{
\label{fig:x_r_blur}
\includegraphics[width=0.23\linewidth]{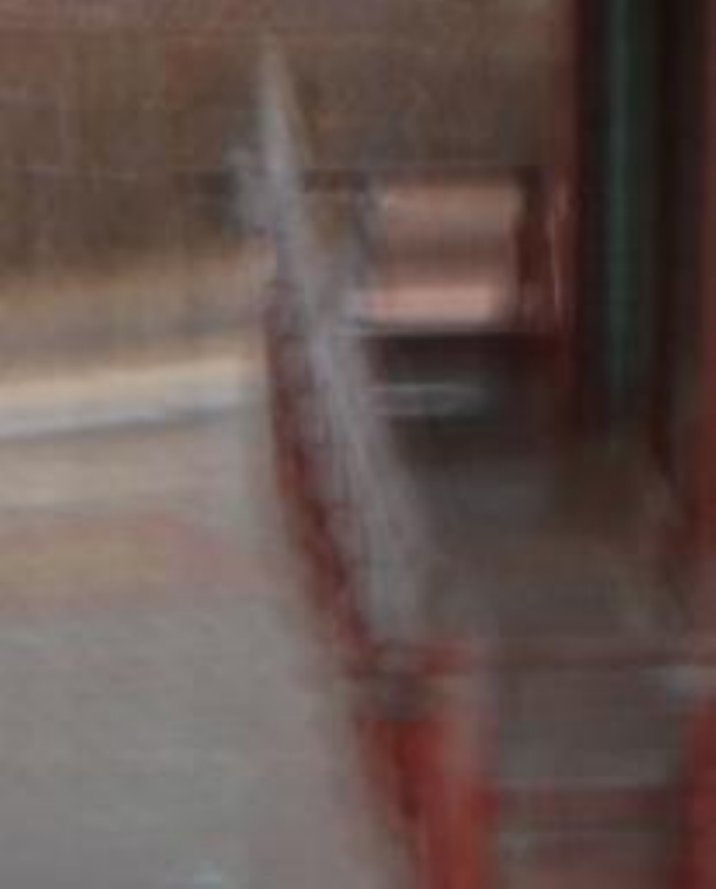}}
\subfloat[]{
\label{fig:x_r_1}
\includegraphics[width=0.23\linewidth]{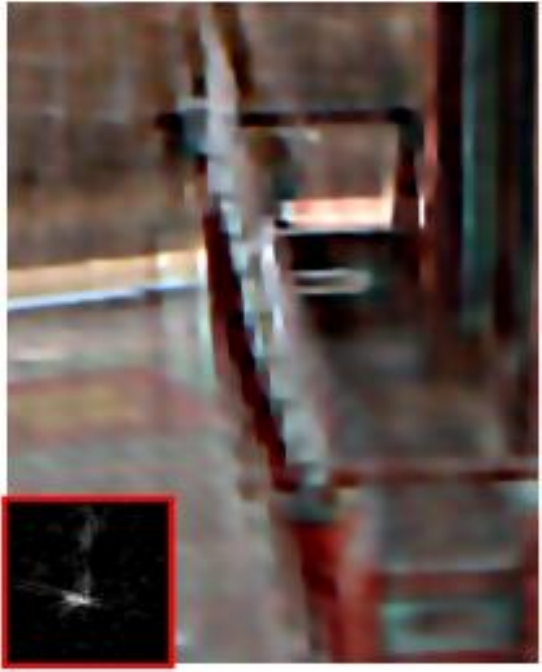}}
\subfloat[]{
\label{fig:x_r_2}
\includegraphics[width=0.23\linewidth]{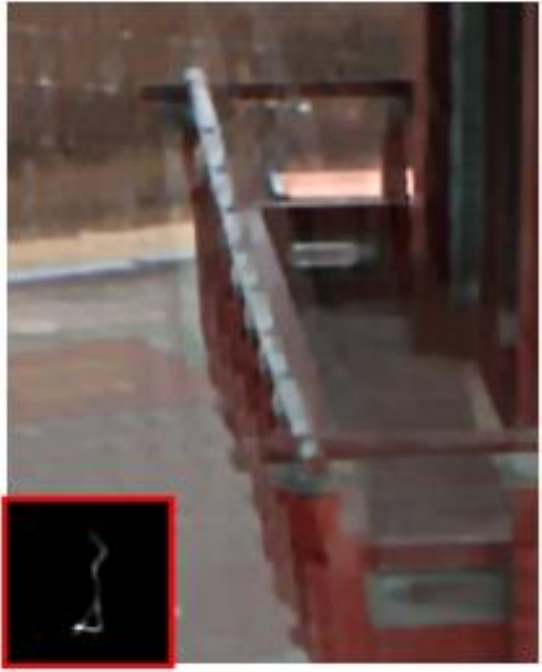}}
\subfloat[]{
\label{fig:x_r_mi}
\includegraphics[width=0.23\linewidth]{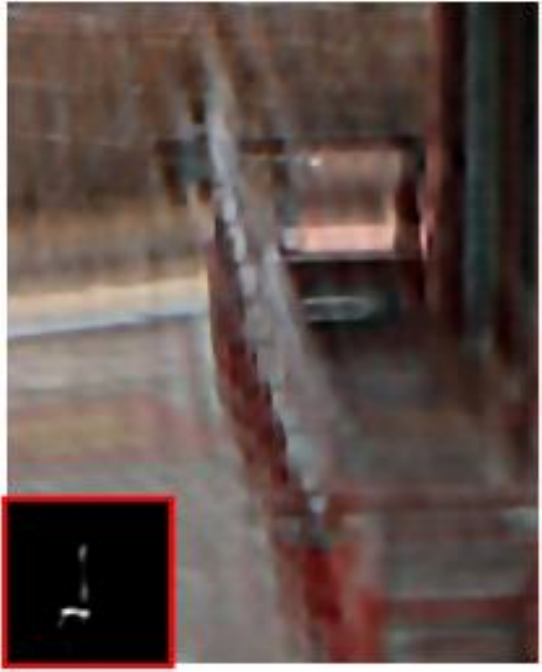}}
\caption{(a) A close-up of a blurry image with image size 972$\times$966 and kernel size 69$\times$69. (b) Preliminary restored result before refinement. (c) Refined result. (d) Restoration result from Michaeli \& Irani \cite{Michaeli2014}.}
\label{fig:x_refine}
\end{figure}

Fig.~\ref{fig:x_refine} illustrates an example of our refined restoration for a challenging blur kernel with fine details. Through preliminary estimation and refinement, the kernel becomes more accurate and the artifacts are reduced in the deblurred image. As a comparison, Fig.~\ref{fig:x_r_mi} shows the result of \cite{Michaeli2014}, which converges to a sub-optimal solution.

\subsection{Extension to Non-Uniform Deblurring}
\label{sebsec:non-uniform}
The non-uniform image deblurring is a more challenging problem than the uniform one, which includes complicated camera motions, such as rotation and translation. Specifically, by representing the camera motion as a more general projective transformation \cite{non_uniform_Hirsh,non_uniform_Whyte}, the blur process can be modeled as:
\begin{align}
    \b=\sum_i k_i\H_i\x+\n
    \label{eq:nu_blur_model}
\end{align}
where $\x$ and $\b$ are vectors of the latent sharp image and the blurry observation, respectively. $\H_i$ is the $i$-th projective transformation matrix of the camera, $k_i$ is the weight of the $i$-th transformation in the blur kernel $\k$, $\n$ is the additive noise.
(\ref{eq:nu_blur_model}) is a linear transformation, which can be further represented as:
\begin{align}
    \b=\A_\k\x+\n=\B_\x\k+\n
    \label{eq:nu_blur_model2}
\end{align}
where matrix $\A_\k=\sum_i k_i\H_i$ and $col(\B_\x)_i=\H_i\x$. $col(\cdot)_i$ denotes the $i$-th column.

Our algorithm can be extended to non-uniform deblurring by assuming that non-uniform image blur is locally uniform.
Based on this assumption, the MSLS prior still works and we apply the same sharp image reconstruction with local self-example matching.
For the kernel estimation in each scale, we replace the convolution model (\ref{eq:blur_model_x}) with (\ref{eq:nu_blur_model2}) and modify (\ref{eq:op_k_xc}) to the following formulation:
\begin{align}
    \argmin_{\k,\x_c} \frac12\|\B_{\nabla(\hx+\x_c)}\k-&\nabla\b\|_2^2 \notag \\
    +&\lambda_1\|\k\|_2^2+\lambda_2\|\x_c\|_1.
    \label{eq:op_k_xc_nu}
\end{align}
where $col(\B_{\nabla(\hx+\x_c)})_i=\H_i(\nabla(\hx+\x_c))$.

We replace $\B_{\nabla\x_c}\k$ with $\v$ and replace $\|\x_c\|_1$ with $\|\v\|_1$ like (\ref{eq:op_k_v}). Then, to solve $\k$ results in a linear equation:
\begin{equation}
    (\B_{\nabla\hx}^T\B_{\nabla\hx}+2\lambda_1\mathbf{I})\k=\B_{\nabla\hx}^T\nabla\b-\B_{\nabla\hx}^T\v.
    \label{eq:k_non_uniform}
\end{equation}
where $col(\B_{\nabla\x_c})_i=\H_i(\nabla\x_c)$, $col(\B_{\nabla\hx})_i=\H_i(\nabla\hx)$, and $\mathbf{I}$ is an identity matrix.
Unlike the fast solver in the frequency domain (\ref{eq:op_k3}), we use \textit{conjugate gradient} method to solve (\ref{eq:k_non_uniform}), because non-uniform blur no longer follows the convolution theorem in the frequency domain. We apply the same soft-thresholding to solve $\v$ by replacing $z_i=\nabla b_i-(\k\otimes\nabla\hx)_i$ with $z_i=\nabla b_i-(\B_{\nabla\hx}\k)_i$ in (\ref{eq:op_v2}).

For TV regularized non-blind deblurring, (\ref{eq:op_tv}) becomes:
\begin{align}
    \argmin_\x \frac12\|\A_\hk\x-\b\|_2^2+\mu\|\nabla\x\|_1,
    \label{eq:op_tv_nu}
\end{align}
which can also be solved with ADMM. Similar modification can be done in the refined restoration of the finest scale.

\section{Experiments and Discussions}
\label{sec:experiments}
In this section, we have done extensive experiments to demonstrate the performance of our algorithm on the artificial dataset of Sun \emph{et~al.} \cite{Sun_iccp2013}, real uniformly blurred images and non-uniformly blurred images. Our experimental platform is a Windows 7 desktop computer with Intel i5 CPU. For uniform deblurring, 8G memory is more than enough. For non-uniform deblurring, we increase the memory to 24G. We use Matlab R2014b to run and test all the matlab codes. We tune the parameters on the dataset of Sun \emph{et al.} and find that they generate satisfactory results for all other images. The parameters are set as follows. That is,     $\lambda_1=\lambda_3=5$, $\lambda_2=\lambda_4=0.05$, $\mu=0.01$.
Besides, the down-sampling factor $\beta$ is set to $\log_23$ for the image pyramid construction. We downsample a blurry image scale by scale until its corresponding kernel is a delta function of size $1\times 1$. The $max\_iteration = 3$ in the preliminary restoration and the refined restoration as a trade-off between accuracy and speed.

\subsection{Artificial Blurry Dataset}
\label{subsec:comparison}
In this experiment, we compare our algorithm with previous blind image deblurring algorithms on a large dataset introduced by Sun \emph{et al.} \cite{Sun_iccp2013}. This dataset includes 640 blurry images (typically $1024\times768$), which are made by performing convolutions between 80 sharp nature images and 8 blur kernels offered by \cite{Levin2011EML}. Each image is then added $1\%$ white Gaussian noise.
The blur kernels are assumed unknown and the kernel sizes are set to $51\times51$ for all test cases.

\begin{figure*}[!t]
\centering
\subfloat[]{
\label{fig:dataset_cp2_blur}
\includegraphics[width=0.15\linewidth]{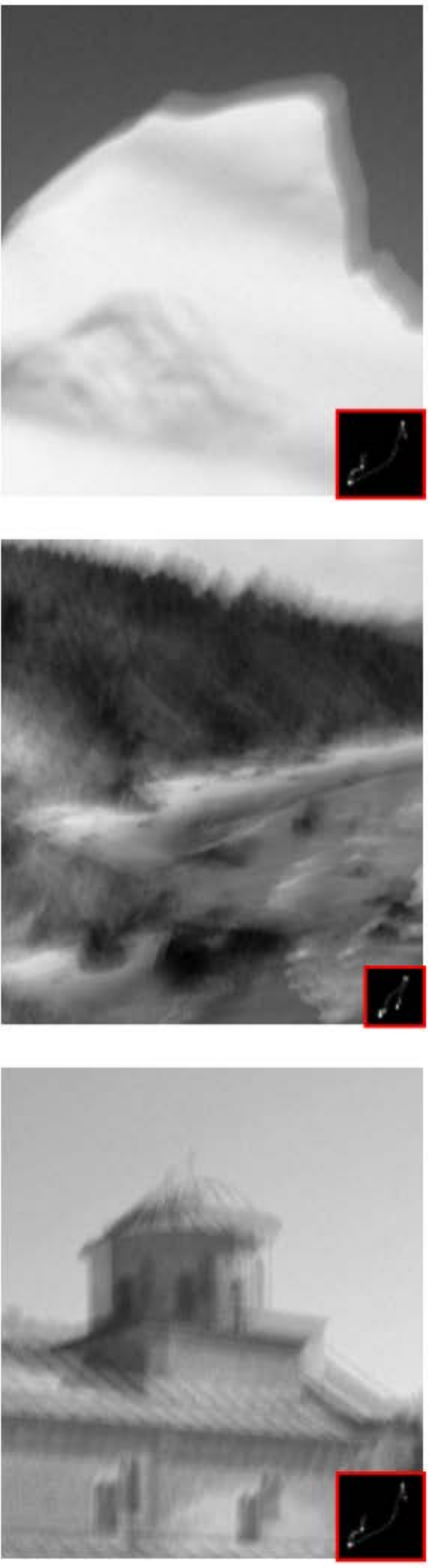}}
\subfloat[]{
\label{fig:dataset_cp2_cho}
\includegraphics[width=0.15\linewidth]{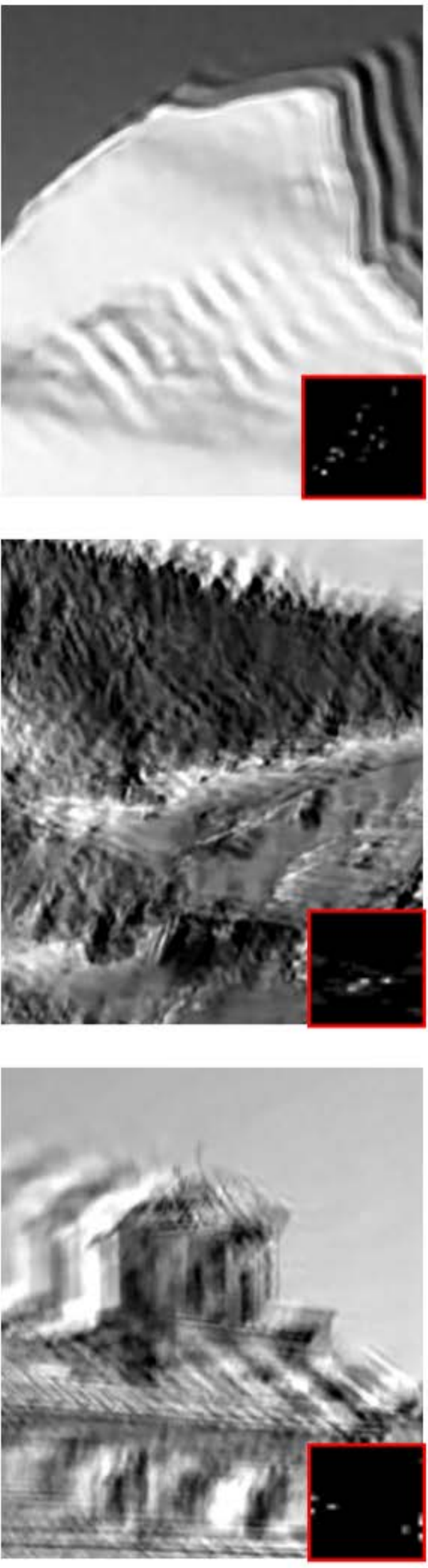}}
\subfloat[]{
\label{fig:dataset_cp2_cho_lee}
\includegraphics[width=0.15\linewidth]{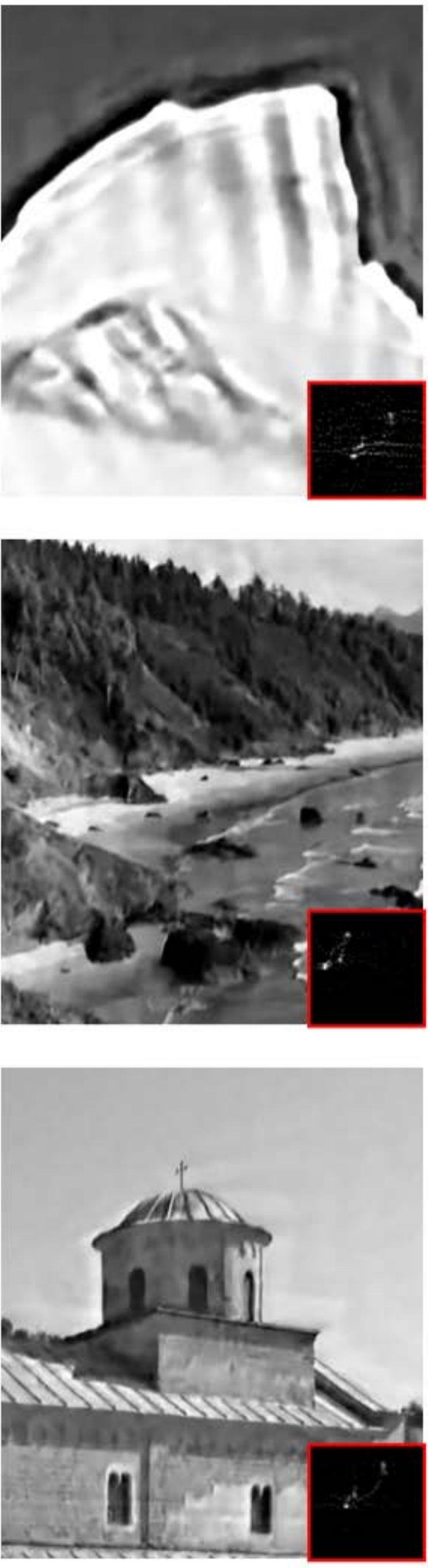}}
\subfloat[]{
\label{fig:dataset_cp2_krishnan}
\includegraphics[width=0.15\linewidth]{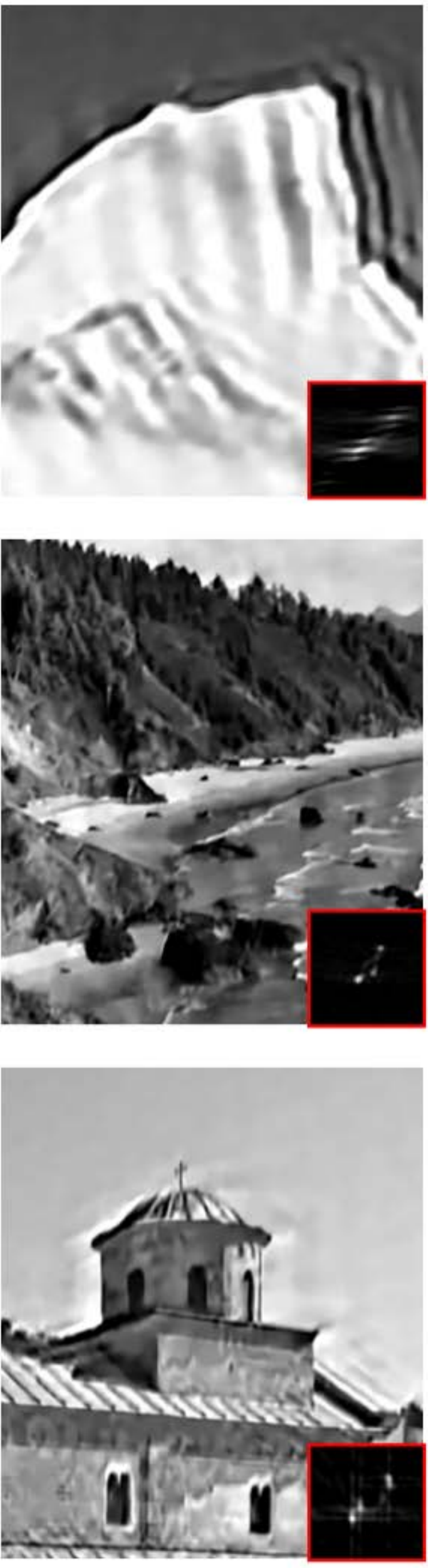}}
\subfloat[]{
\label{fig:dataset_cp2_MI}
\includegraphics[width=0.15\linewidth]{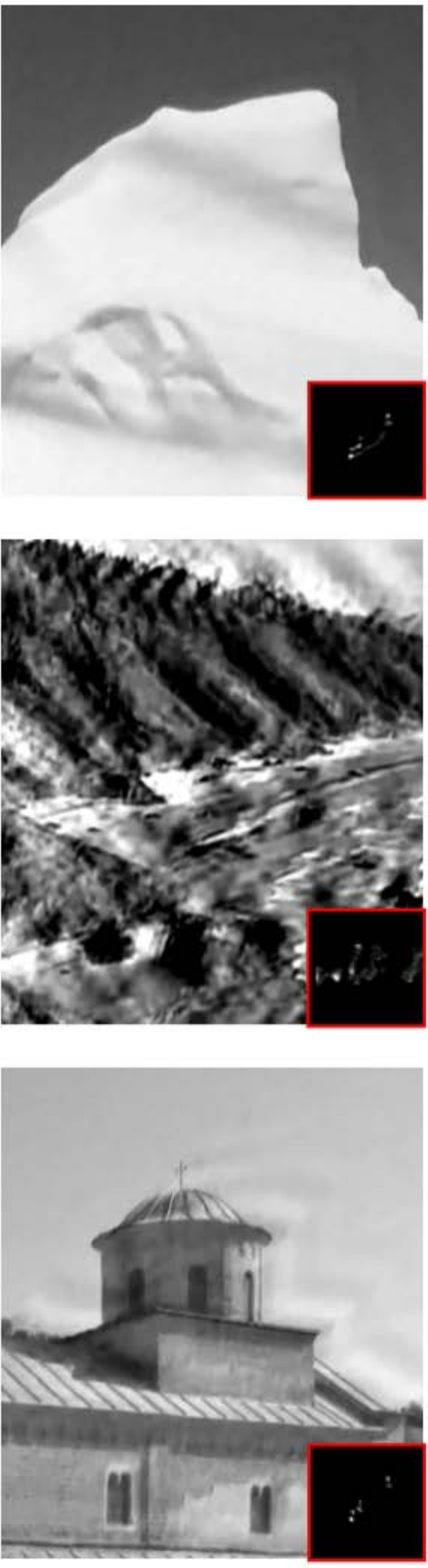}}
\subfloat[]{
\label{fig:dataset_cp2_ours}
\includegraphics[width=0.15\linewidth]{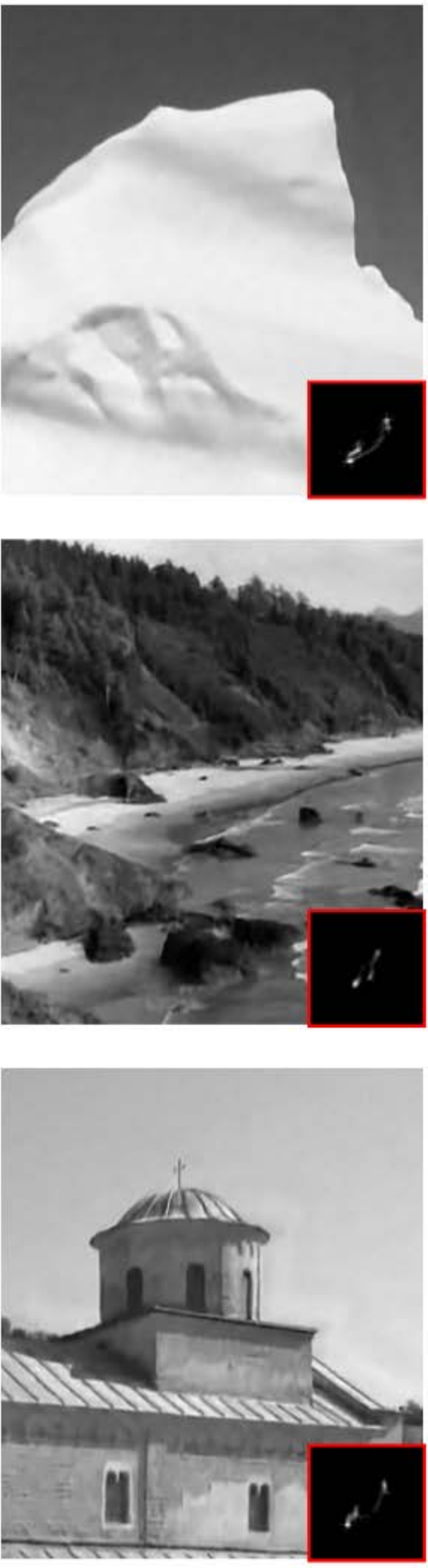}}
\caption{Blurry images and deblurred results with zoomed regions. (a) Blurry images and ground-truth kernels. (b) Cho \emph{et al.} \cite{cho}. (c) Cho \& Lee \cite{cho_lee}. (d) Krishnan \emph{et al.} \cite{Krishnan2011}. (e) Lai \emph{et al.} \cite{Lai2005color}. (f) Ours.}
\label{fig:dataset_cp2}
\end{figure*}

\begin{figure*}[!t]
\centering
\subfloat[]{
\label{fig:dataset_cp_blur}
\includegraphics[width=0.15\linewidth]{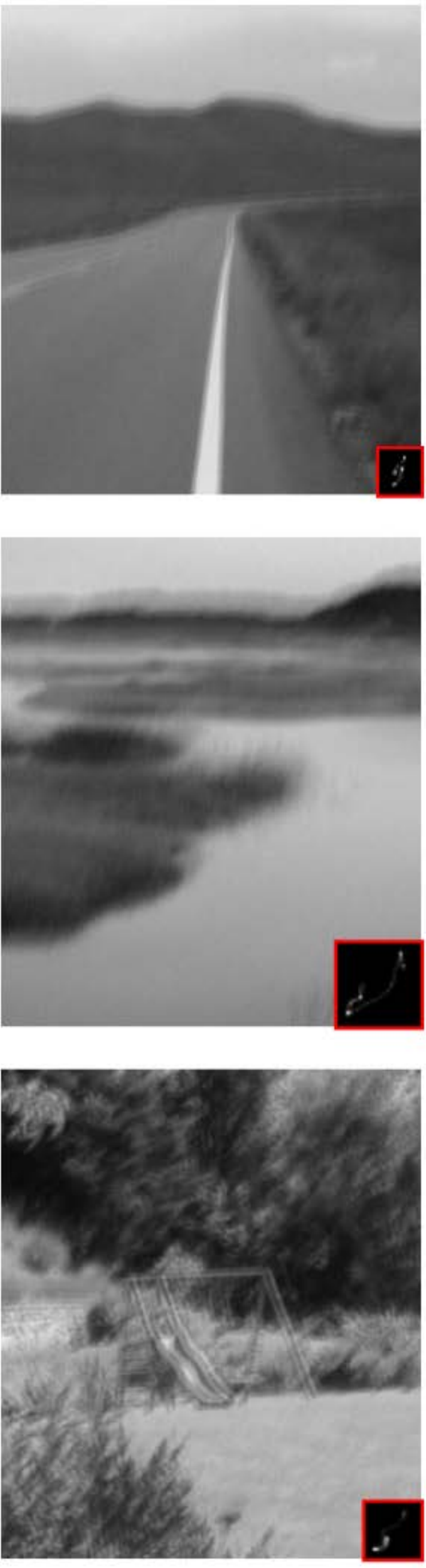}}
\subfloat[]{
\label{fig:dataset_cp_xj}
\includegraphics[width=0.15\linewidth]{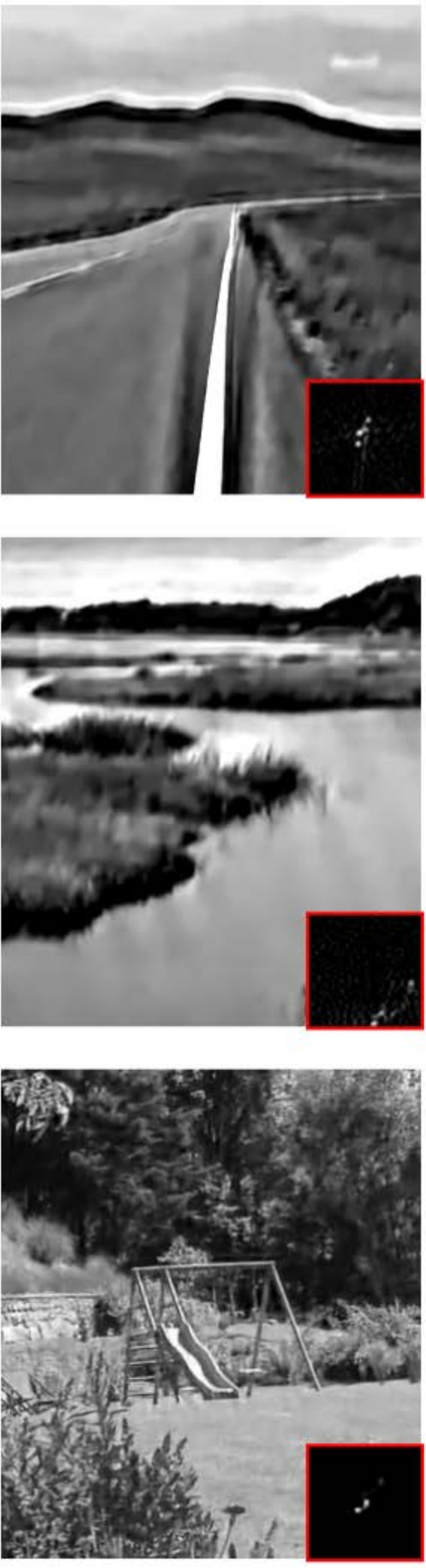}}
\subfloat[]{
\label{fig:dataset_cp_sun}
\includegraphics[width=0.15\linewidth]{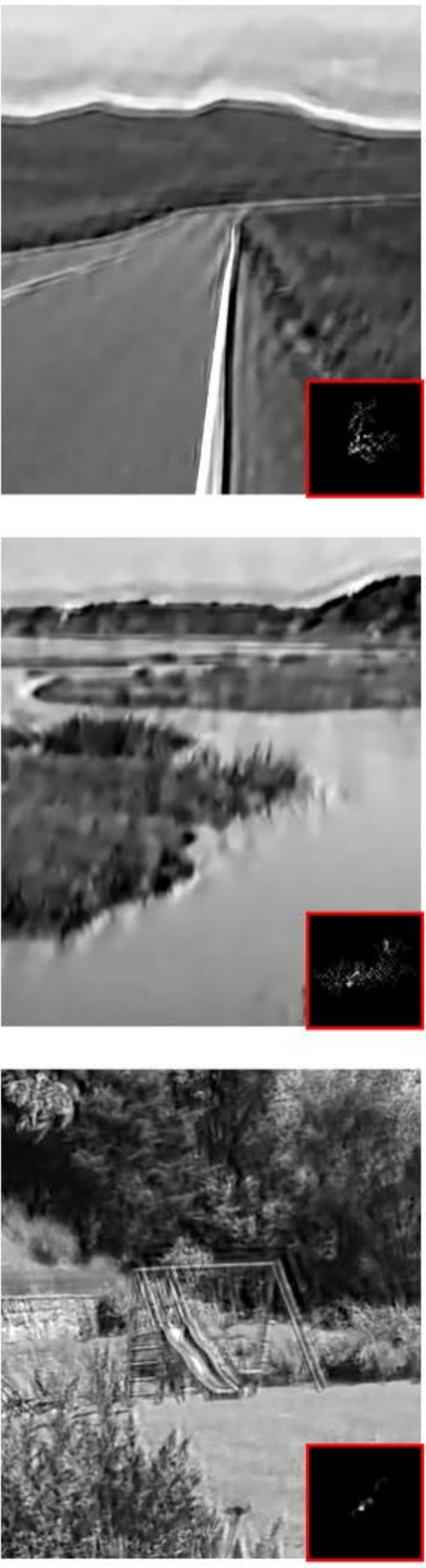}}
\subfloat[]{
\label{fig:dataset_cp_MI}
\includegraphics[width=0.15\linewidth]{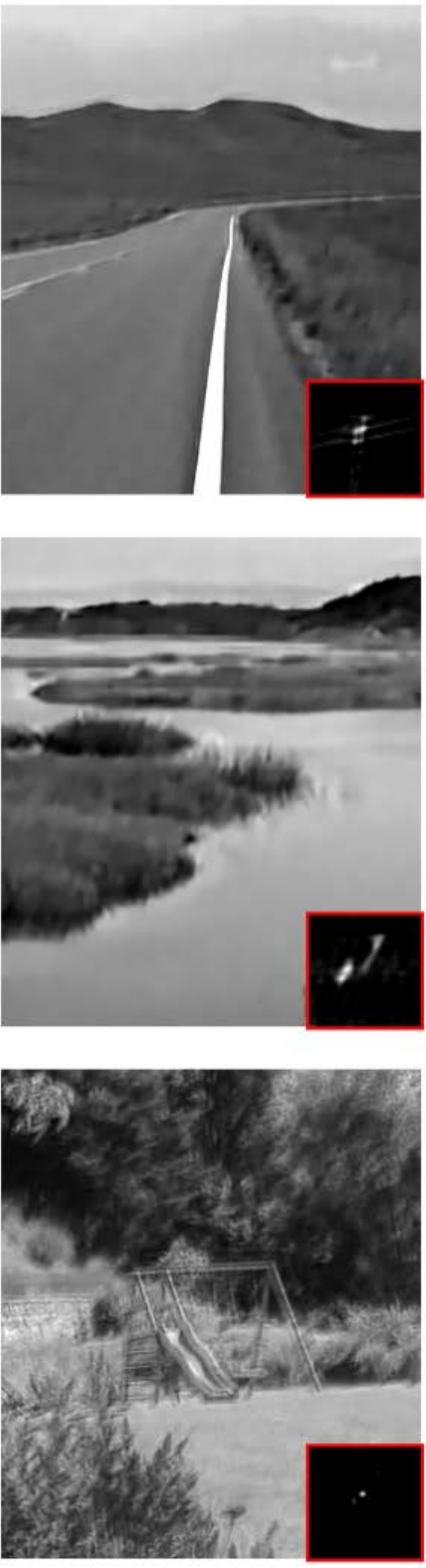}}
\subfloat[]{
\label{fig:dataset_cp_jinshan}
\includegraphics[width=0.15\linewidth]{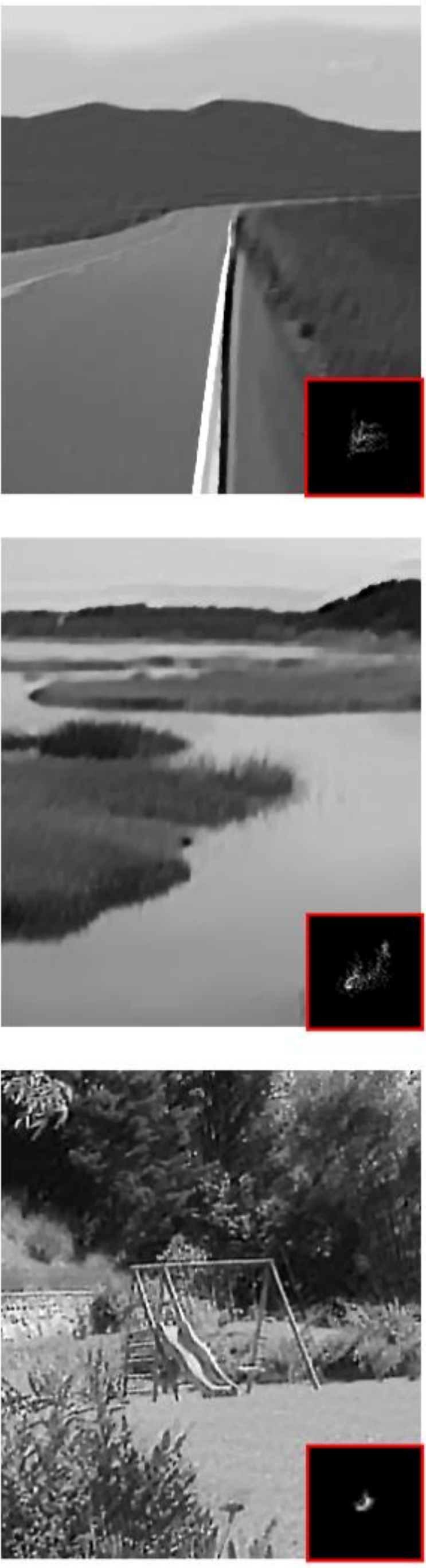}}
\subfloat[]{
\label{fig:dataset_cp_ours}
\includegraphics[width=0.15\linewidth]{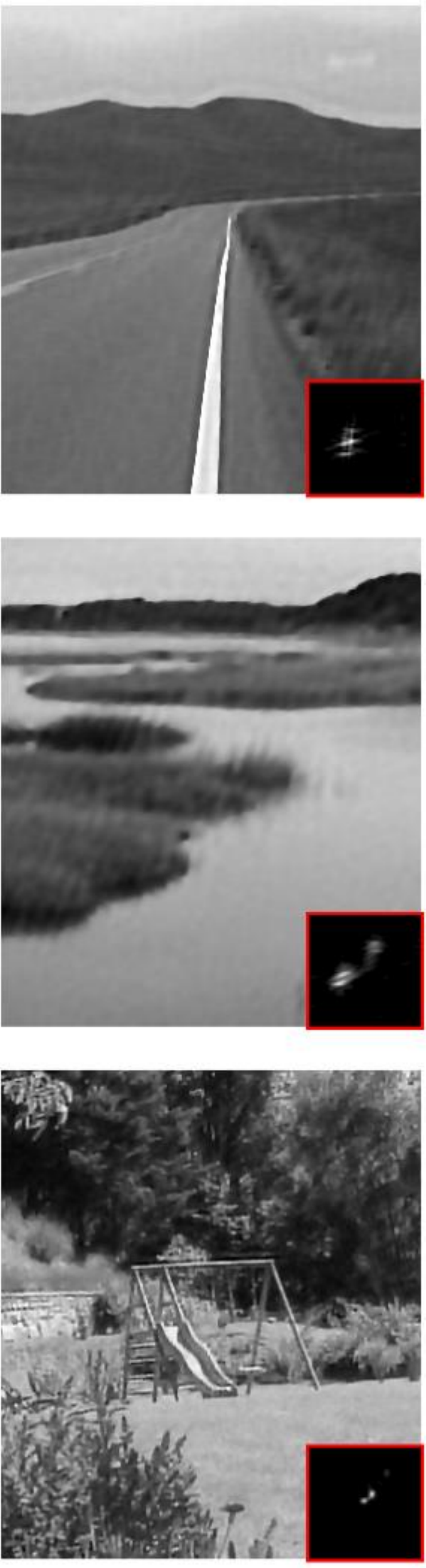}}
\caption{Blurry images and deblurred results with zoomed regions. (a) Blurry images and ground-truth kernels. (b) Xu \& Jia\cite{XuJia2010eccv}. (c) Sun \emph{et al.} \cite{Sun_iccp2013}. (d) Michaeli \& Irani\cite{Michaeli2014}. (e) Pan \emph{et al.} \cite{Pan_2016_CVPR}. (f) Ours. }
\label{fig:dataset_cp}
\end{figure*}

In Fig.~\ref{fig:dataset_cp2} and \ref{fig:dataset_cp}, we show the visual results deblurred by our algorithm and eight competing algorithms, \emph{i.e.}, Krishnan \emph{et al.} \cite{Krishnan2011}, Lai \emph{et al.} \cite{Lai2005color}, Pan \emph{et al.} \cite{Pan_2016_CVPR}, Michaeli \& Irani \cite{Michaeli2014}, Cho \& Lee \cite{cho_lee}, Xu \& Jia \cite{XuJia2010eccv}, Sun \emph{et al.} \cite{Sun_iccp2013}, Cho \emph{et al.} \cite{cho}.
For Pan et al. \cite{Pan_2016_CVPR}, the results are generated with their code. For other algorithms, the results are provided by their authors or by Sun \emph{et al.} in the dataset.
As shown in Fig.~\ref{fig:dataset_cp2} and \ref{fig:dataset_cp}, our algorithm performs visually better than other algorithms. The edges and details are well restored and the artifacts in our results are much less than those in others.

\begin{figure}[!t]
\centering
\includegraphics[width=0.95\linewidth]{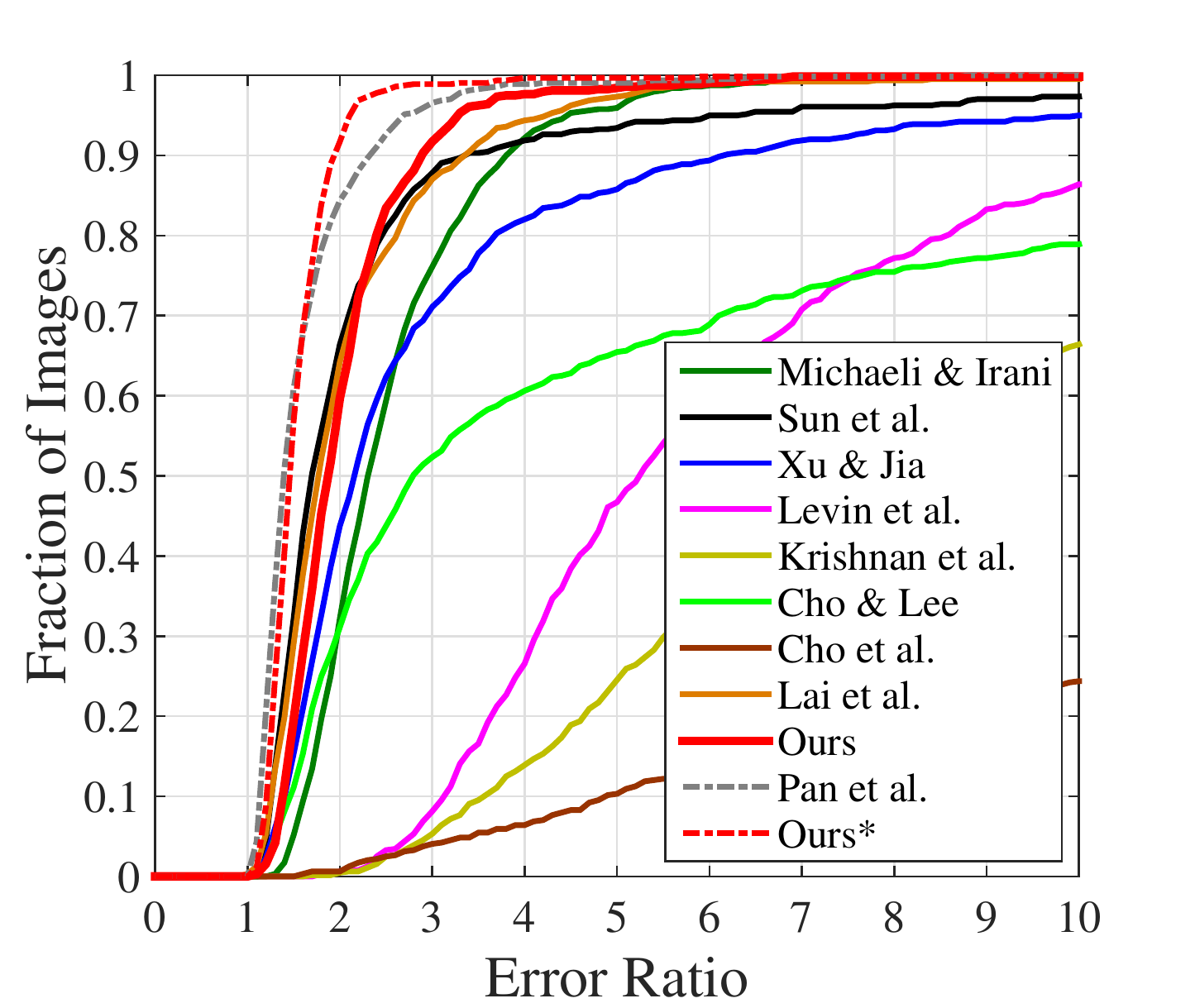}
\caption{Cumulative distribution of error ratios. Each curve represents the fractions of 640 images that can be deblurred under different error ratios.}
\label{fig:error_ratio}
\end{figure}

Besides the visual quality assessment, we also measure the quality of the deblurred results using error ratio, which was introduced by \cite{Levin2011Pami}. All the algorithms estimate blur kernels and use the same non-blind deblurring algorithm to restore the latent sharp images. The error ratios are then computed by
\begin{align}
    r=\frac{\|\x-\x_{\hk}\|^2}{\|\x-\x_{\k}\|^2}
    \label{eq:error_ratio}
\end{align}
where $\x_{\hk}$ is the deblurred result computed using the estimated kernel $\hk$, $\x_{\k}$ is the deblurred result computed using the ground-truth kernel $\k$, $\x$ is the ground-truth sharp image. If $r=1$, it means that the blindly deblurred result is as good as the result restored using the ground-truth kernel. The smaller $r$ is, the better the blindly deblurred result is. We also agree with \cite{Michaeli2014} that the visual quality is satisfactory when $r\le5$ and consider it as the threshold to decide the success. For all the competing methods except \cite{Pan_2016_CVPR}, of which the results are provided by their authors or by Sun \emph{et al.}, they estimated the blur kernels and used non-blind deblurring algorithm \cite{Zoran2011patch_restoration} to restore sharp images.
For Pan \emph{et al.} \cite{Pan_2016_CVPR}, we generate deblurred images with their code.
Their code estimates the blur kernels and uses the non-blind method proposed in \cite{Pan2016Pami} to restore sharp images. In order to fairly compare with all the algorithms, we run our algorithms on both settings (Using \cite{Zoran2011patch_restoration} and \cite{Pan2016Pami} for the last-step non-blind deblurring, respectively).
Fig. \ref{fig:error_ratio} reports the cumulative performance of error ratio.
Each curve represents the fractions of 640 images that can be deblurred under different error ratios. For error ratios of results restored using \cite{Zoran2011patch_restoration} (solid lines), the curves of Sun \emph{et al.} \cite{Sun_iccp2013} and Lai \emph{et al.} \cite{Lai2005color} are slightly higher than ours when $error\ ratio<2.2$. 
Since all results in this range are very well recovered, the visual differences between their results and ours are hardly noticed. 
When $error\ ratio\ge2.2$, our curve achieves the highest and the improvement is obvious. For error ratios of results restored using \cite{Pan2016Pami} (dash lines), the two curves are very close at the beginning and our curve soon becomes higher than that of Pan \emph{et al.} \cite{Pan_2016_CVPR}.

In Table \ref{tb:error_ratio}, we further report three statistical measures, \emph{i.e.}, mean error ratio, worst error ratio and success rate, to further compare all the blind deblurring algorithms. According to the table, the mean errors of our algorithm are the smallest and the success rates of our algorithm are the highest among all the algorithms on both settings. Although our worst error ratios are larger than Michaeli \& Irani \cite{Michaeli2014} and Pan \emph{et al.} \cite{Pan_2016_CVPR}, our second worst results are better on both settings.
\begin{table}[!t]
\caption{Quantitative comparison of all methods over the entire dataset (640 blurry images).}
\label{tb:error_ratio}
\centering
\begin{tabular}{|C{9em}|C{5em}C{5em}C{5em}|}
\hline
  Blind Deblurring Method & Mean error ratio& Worst error ratio&Success rate (r $<$ 5) \tabularnewline
\hline
\hline
Cho \emph{et al.} \cite{cho}       & 28.1 & 165.0 & 11.7\%      \tabularnewline
Krishnan \emph{et al.} \cite{Krishnan2011}  & 11.7 & 133.2 & 24.8\% \tabularnewline
Levin \emph{et al.} \cite{Levin2011EML}     & 6.6 & 40.9 & 46.7\%    \tabularnewline
Cho \& Lee   \cite{cho_lee}            & 8.7 & 111.1 & 65.5\%    \tabularnewline
Xu \& Jia    \cite{XuJia2010eccv}            & 3.6 & 65.3 & 85.8\%    \tabularnewline
Sun \emph{et al.}   \cite{Sun_iccp2013}            & 2.5 & 30.5 & 93.4\%    \tabularnewline
Michaeli \& Irani \cite{Michaeli2014}  & 2.6 & \textbf{9.3} & 95.9\%    \tabularnewline
Lai \emph{et al.} \cite{Lai2005color}                & \textbf{2.1} & 17.9         & 97.3\% \tabularnewline
Ours                     & \textbf{2.1} & 14.0 & \textbf{98.4\%}    \tabularnewline
\hline
\hline
Pan et al. \cite{Pan_2016_CVPR} & \textbf{1.6} & \textbf{8.8}&  99.1\%  \tabularnewline
Ours*                           & \textbf{1.6} & 10.4&  \textbf{99.7\%}  \tabularnewline
\hline
\end{tabular}
\end{table}

\begin{figure}[!t]
\centering
\subfloat[]{
\label{fig:bad_case_blurred}
\includegraphics[width=0.48\linewidth]{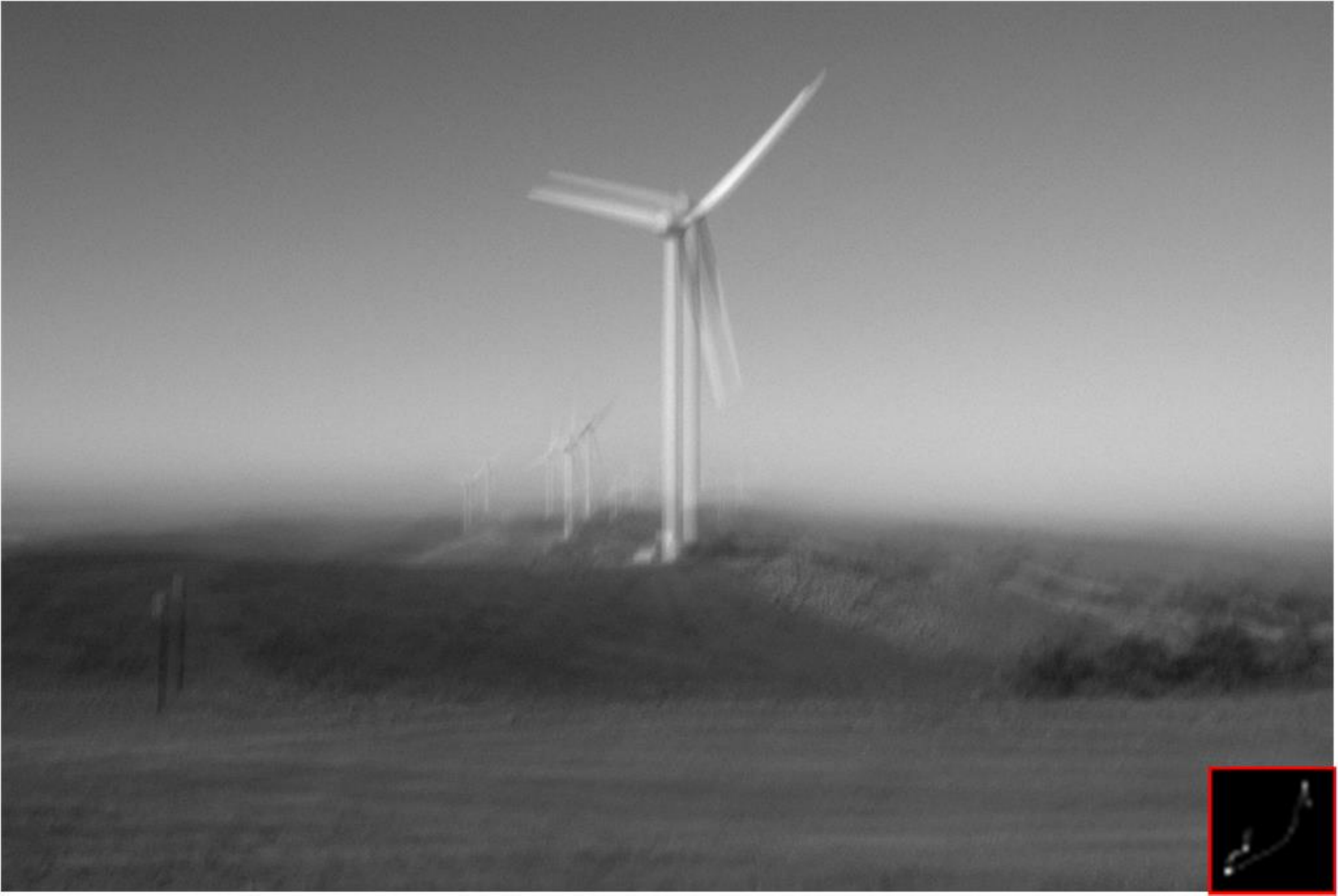}}
\subfloat[]{
\label{fig:bad_case_deblurred}
\includegraphics[width=0.48\linewidth]{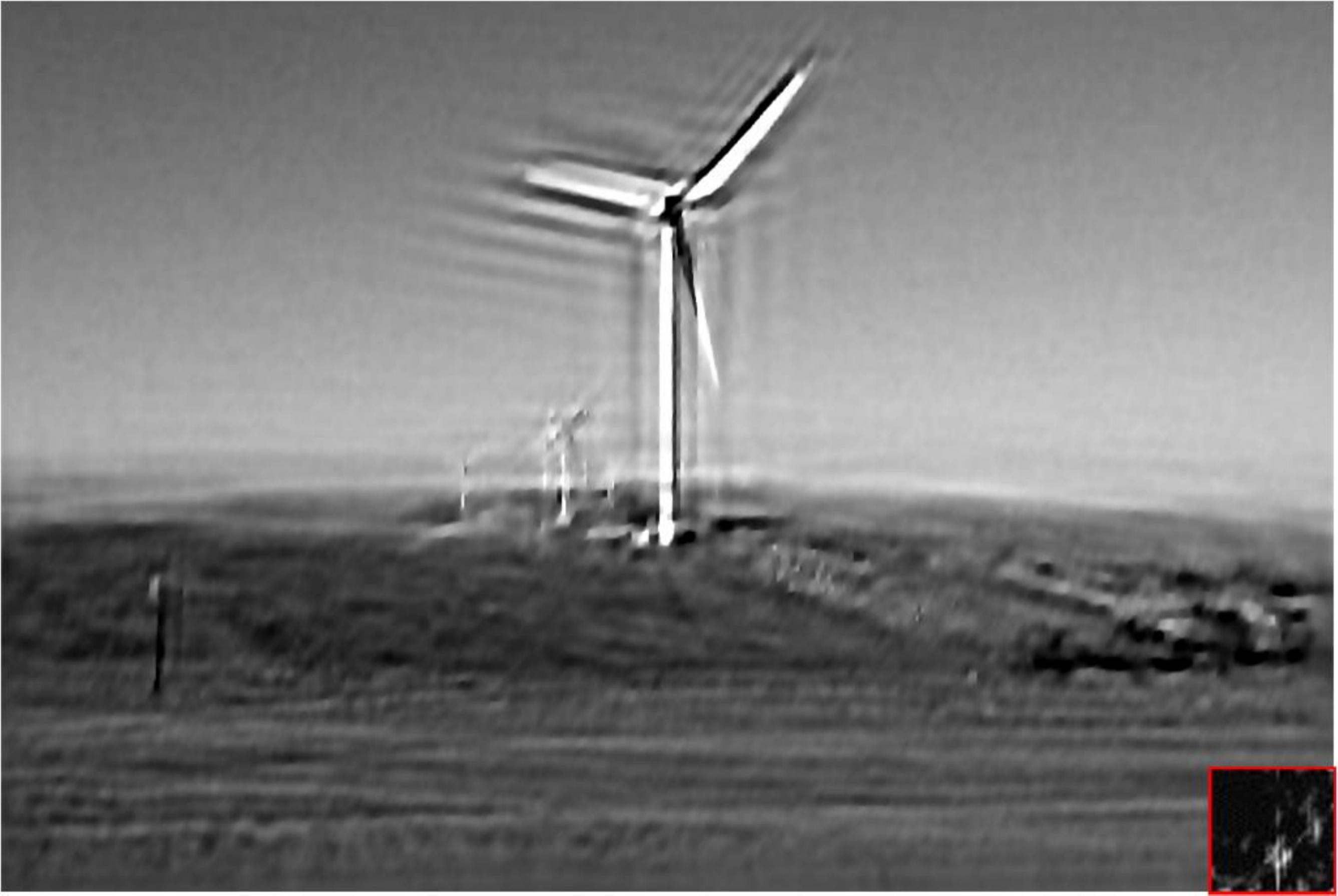}}
\caption{An example of an unsatisfactory blind image deblurring result. (a) A blurred image and its corresponding true blur kernel. (b) The deblurred result from our algorithm.}
\label{fig:bad_case}
\end{figure}

In Fig.~\ref{fig:bad_case}, we show an example of an unsatisfactory blind image deblurring result. In our approach, we require that there are objects with profiles, of which the sizes are much larger than that of the blur kernel. However, in this case, smooth sky accounts for most of the image. Only a windmill and foothills in the foreground provide profiles. The small windmills in the background will be removed after down-sampling. Thus, there aren't enough constraints in the data fidelity term for the complex blur kernel restoration, leading to the unsatisfactory blind deblurring result.

\subsection{Real Blurry Images}
\begin{figure}[!t]
\centering
\subfloat[]{
\label{fig:flower1_blur}
\includegraphics[width=0.48\linewidth]{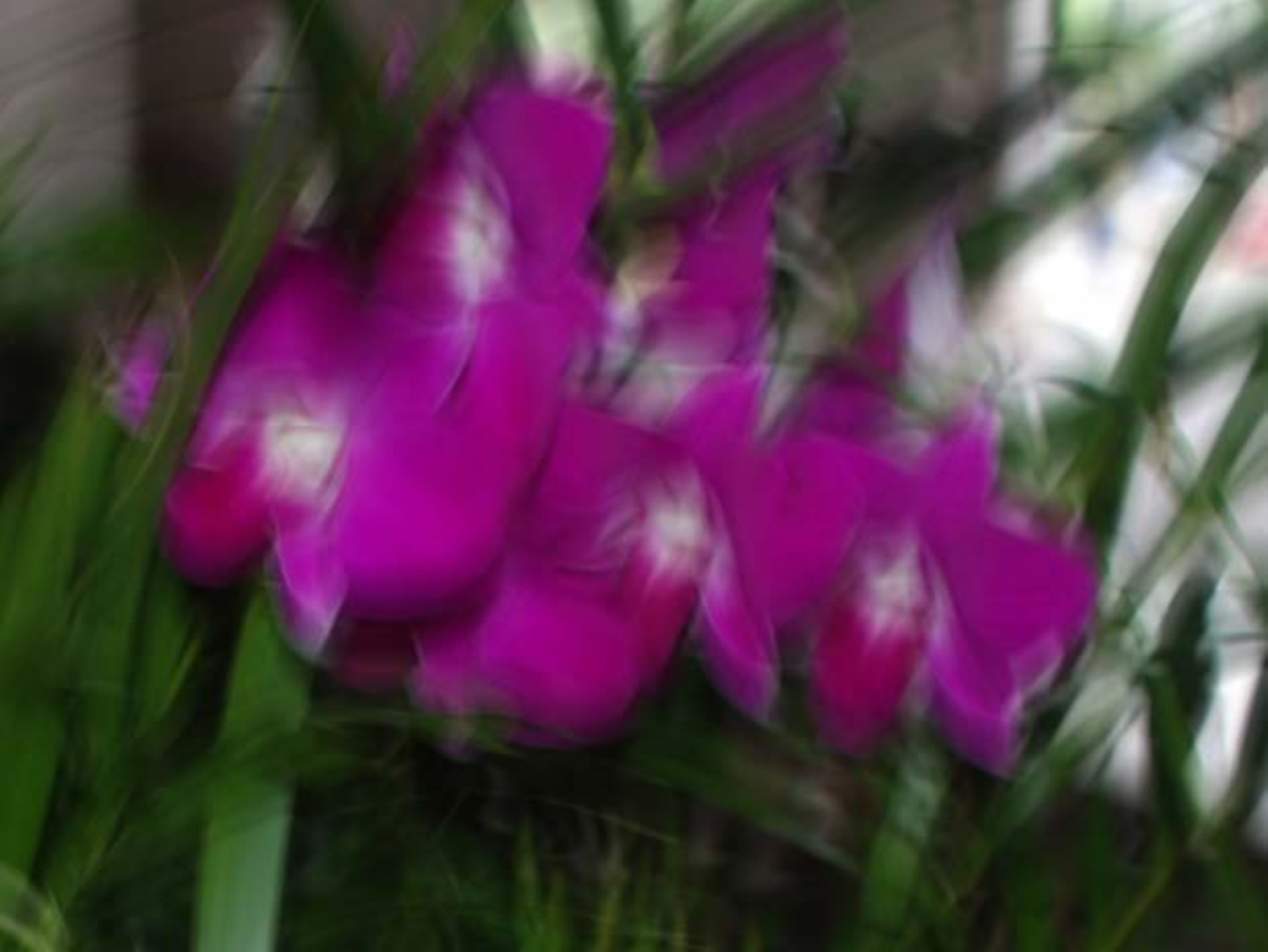}}
\subfloat[]{
\label{fig:flower1_krishnan}
\includegraphics[width=0.48\linewidth]{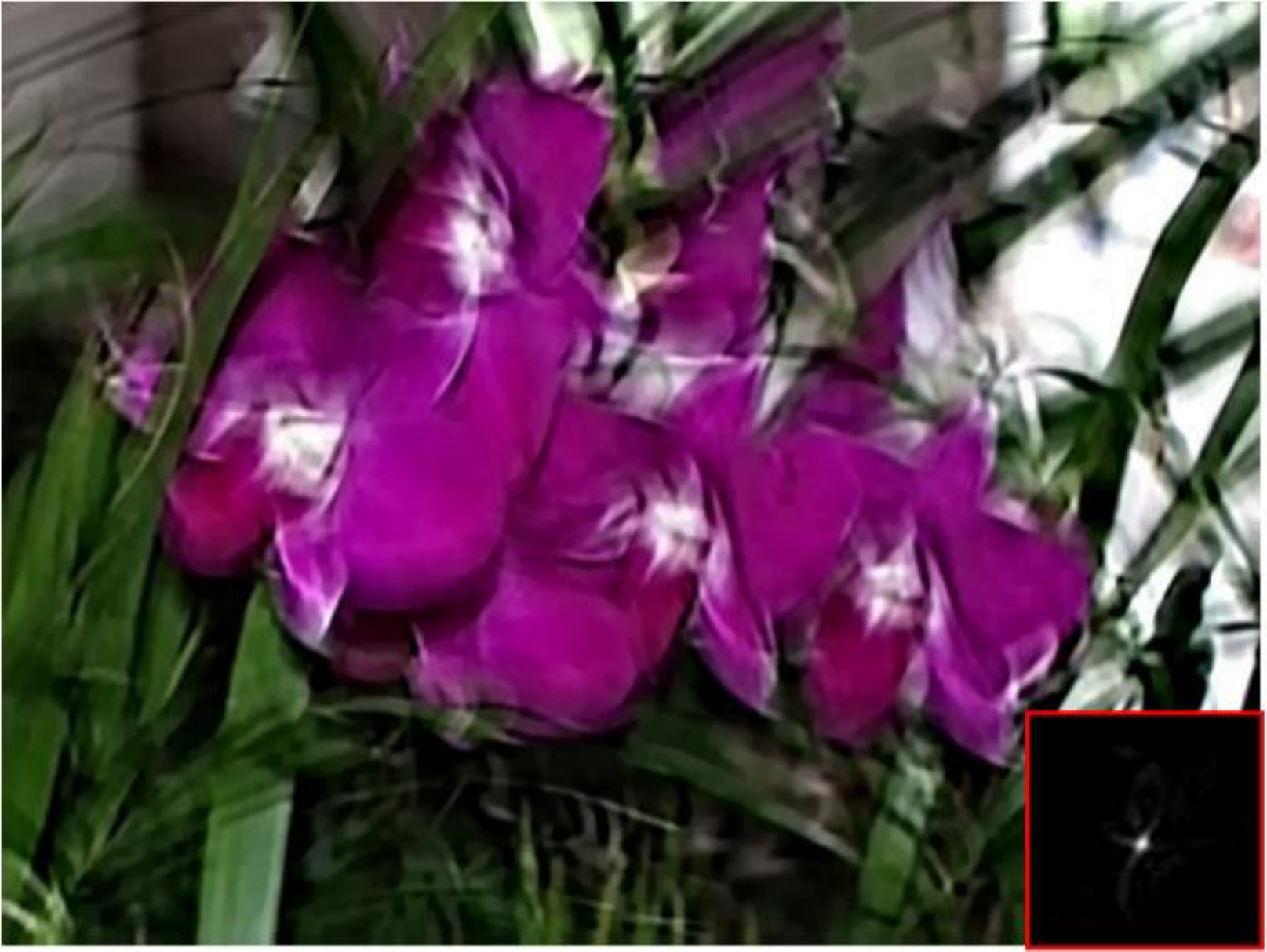}}\\
\subfloat[]{
\label{fig:flower1_levin}
\includegraphics[width=0.48\linewidth]{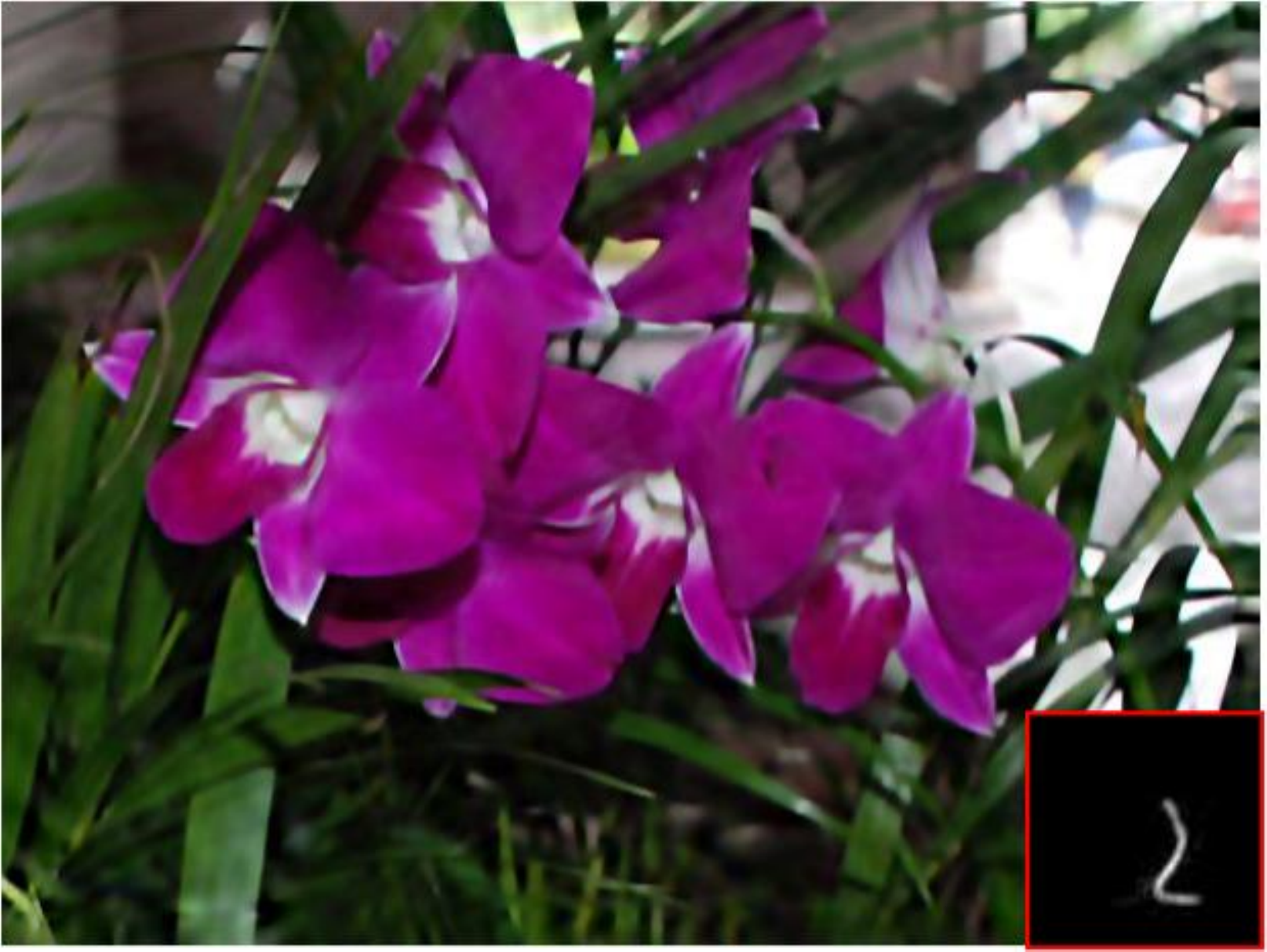}}
\subfloat[]{
\label{fig:flower1_michaeli}
\includegraphics[width=0.48\linewidth]{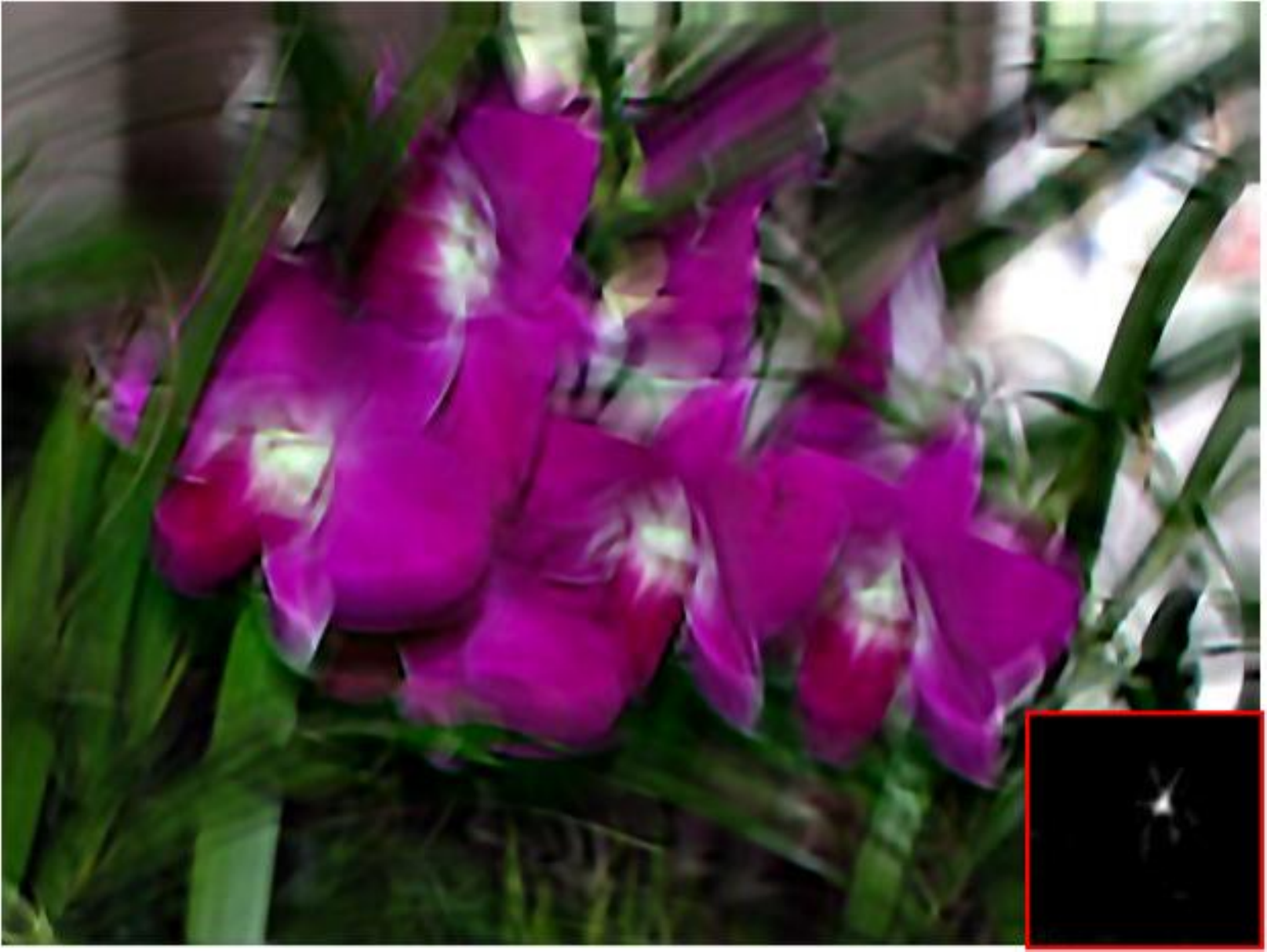}}\\
\subfloat[]{
\label{fig:flower1_pan}
\includegraphics[width=0.48\linewidth]{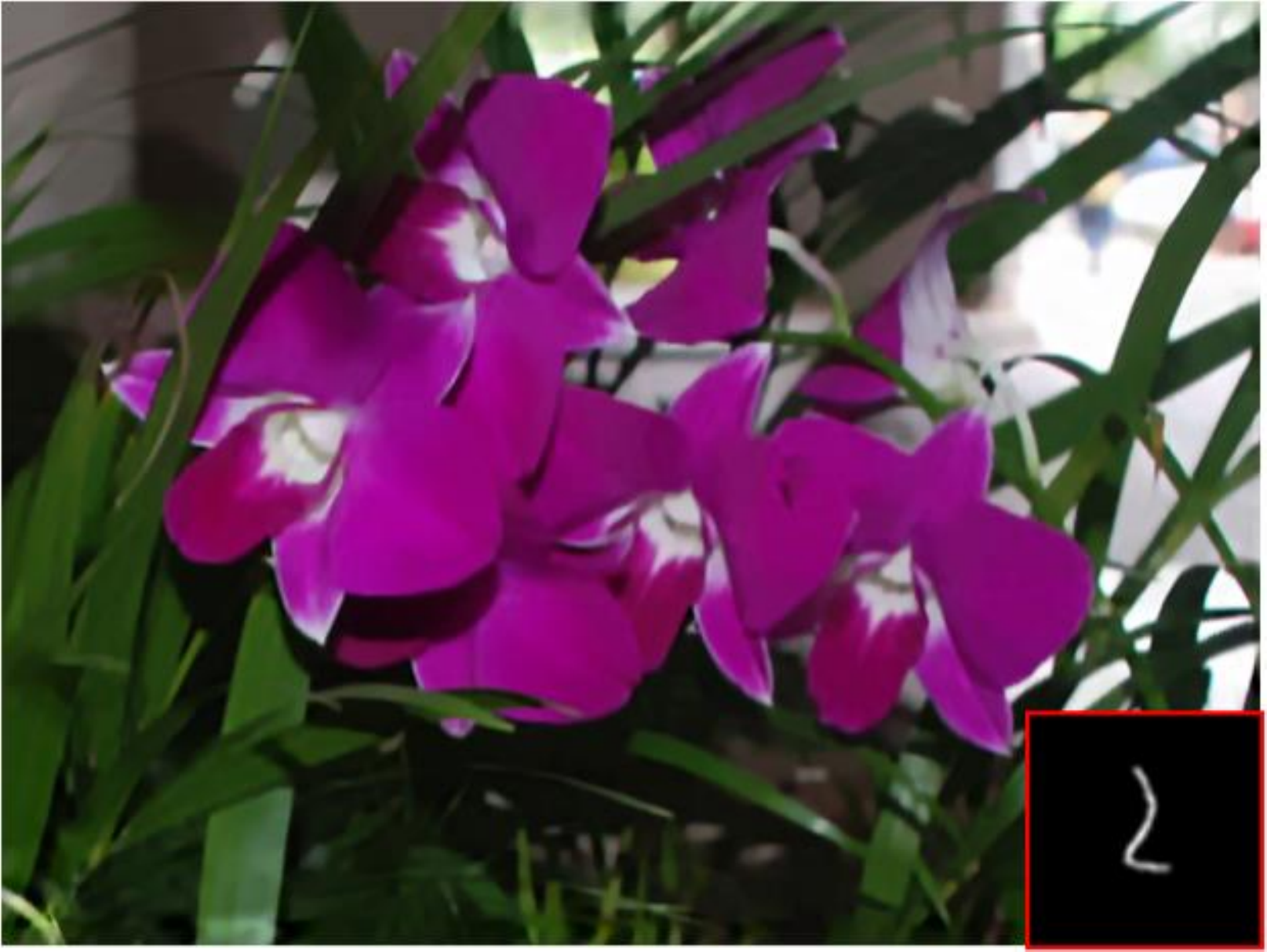}}
\subfloat[]{
\label{fig:flower1_ours}
\includegraphics[width=0.48\linewidth]{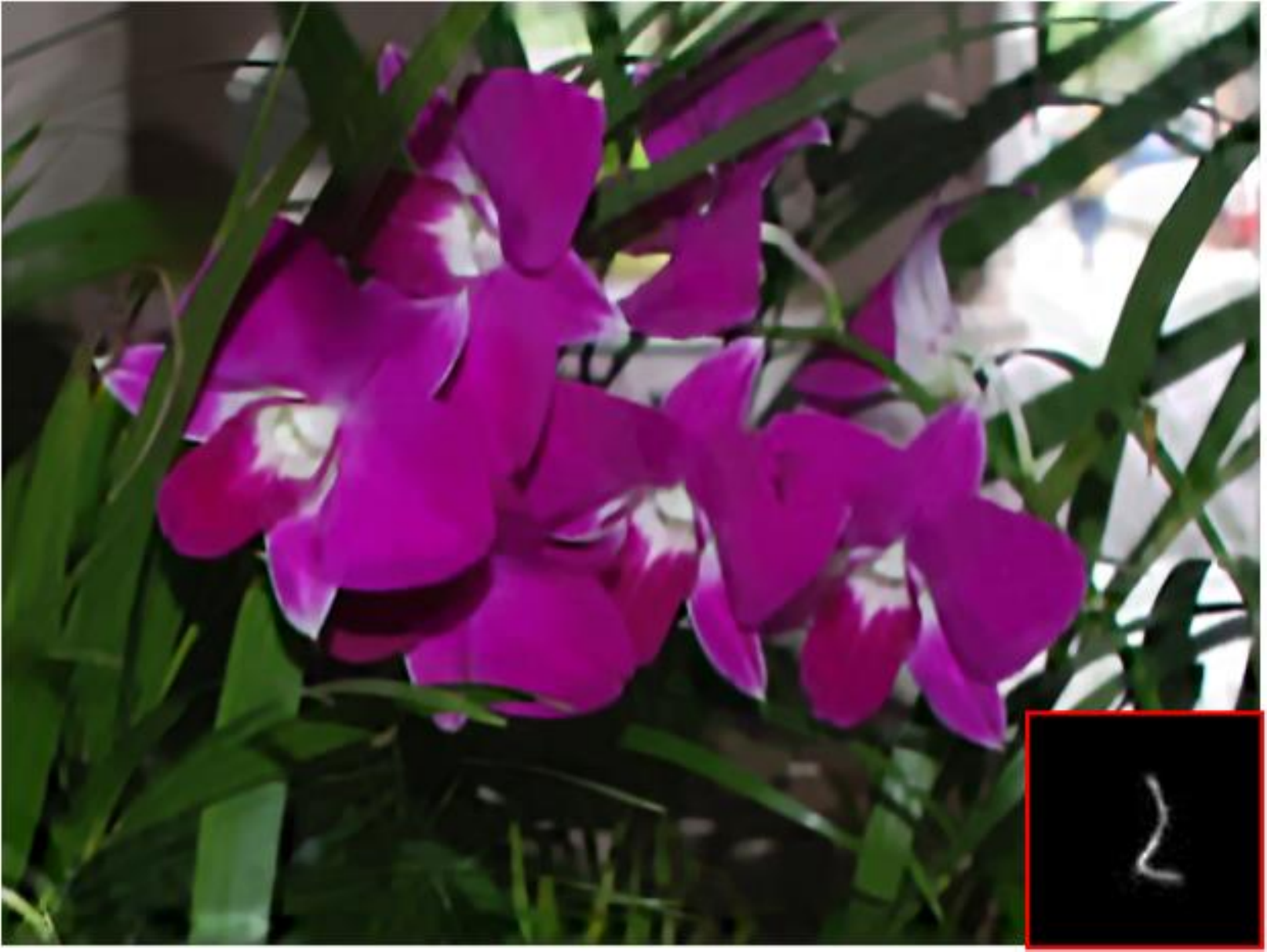}}
\caption{{\itshape Flower.} Image Size: $618\times464$, Kernel Size: $69\times69$. (a) Blurry Image. (b) Krishnan \emph{et al.} \cite{Krishnan2011}. (c) Levin \emph{et al.} \cite{Levin2011EML}. (d) Michaeli \& Irani \cite{Michaeli2014}. (e) Pan \emph{et al.} \cite{Pan_2016_CVPR}. (f) Ours.}
\label{fig:flower1}
\end{figure}

\begin{figure}[!t]
\centering
\subfloat[]{
\label{fig:picasso_blur}
\includegraphics[width=0.48\linewidth]{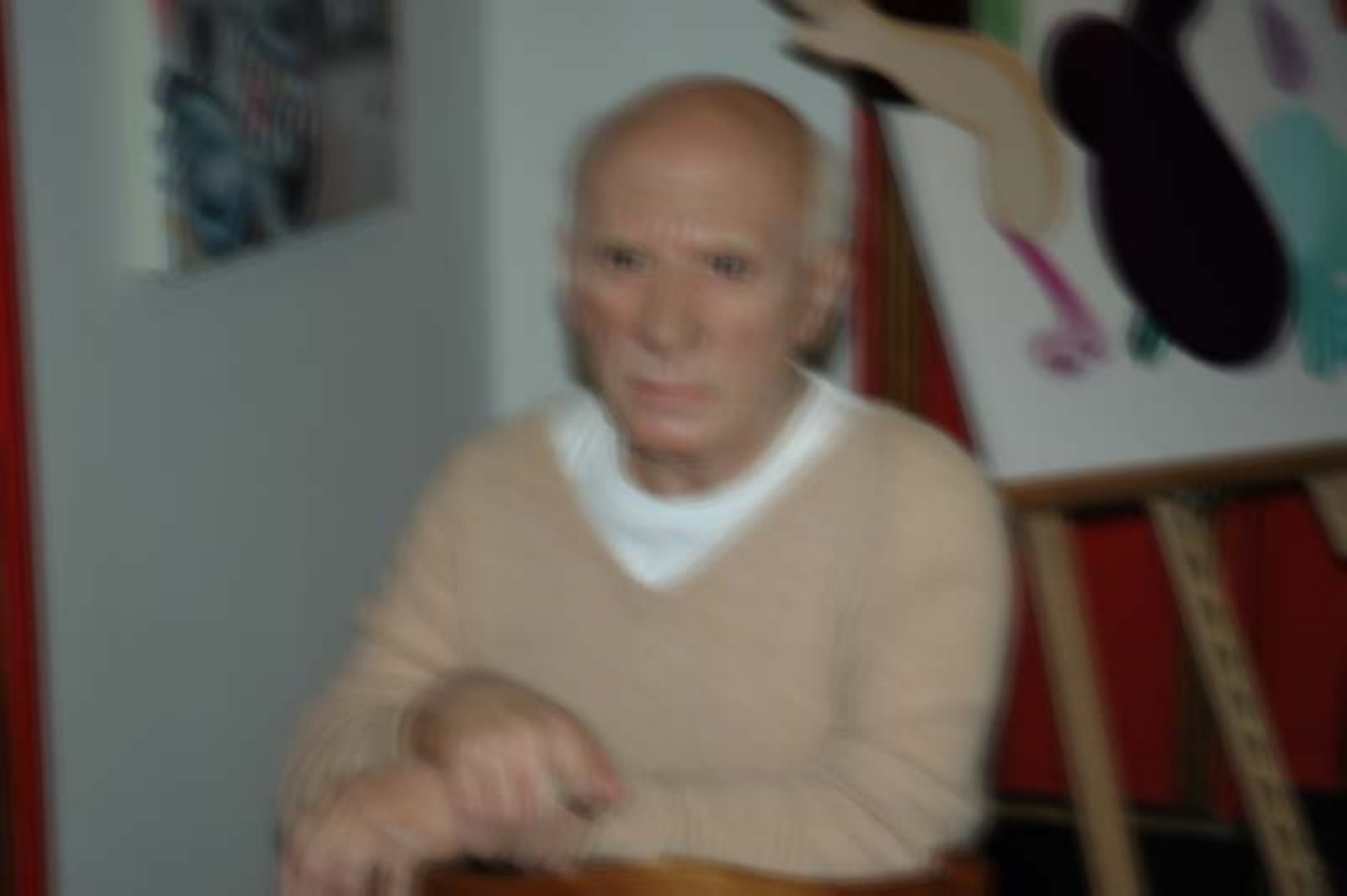}}
\subfloat[]{
\label{fig:picasso_krishnan}
\includegraphics[width=0.48\linewidth]{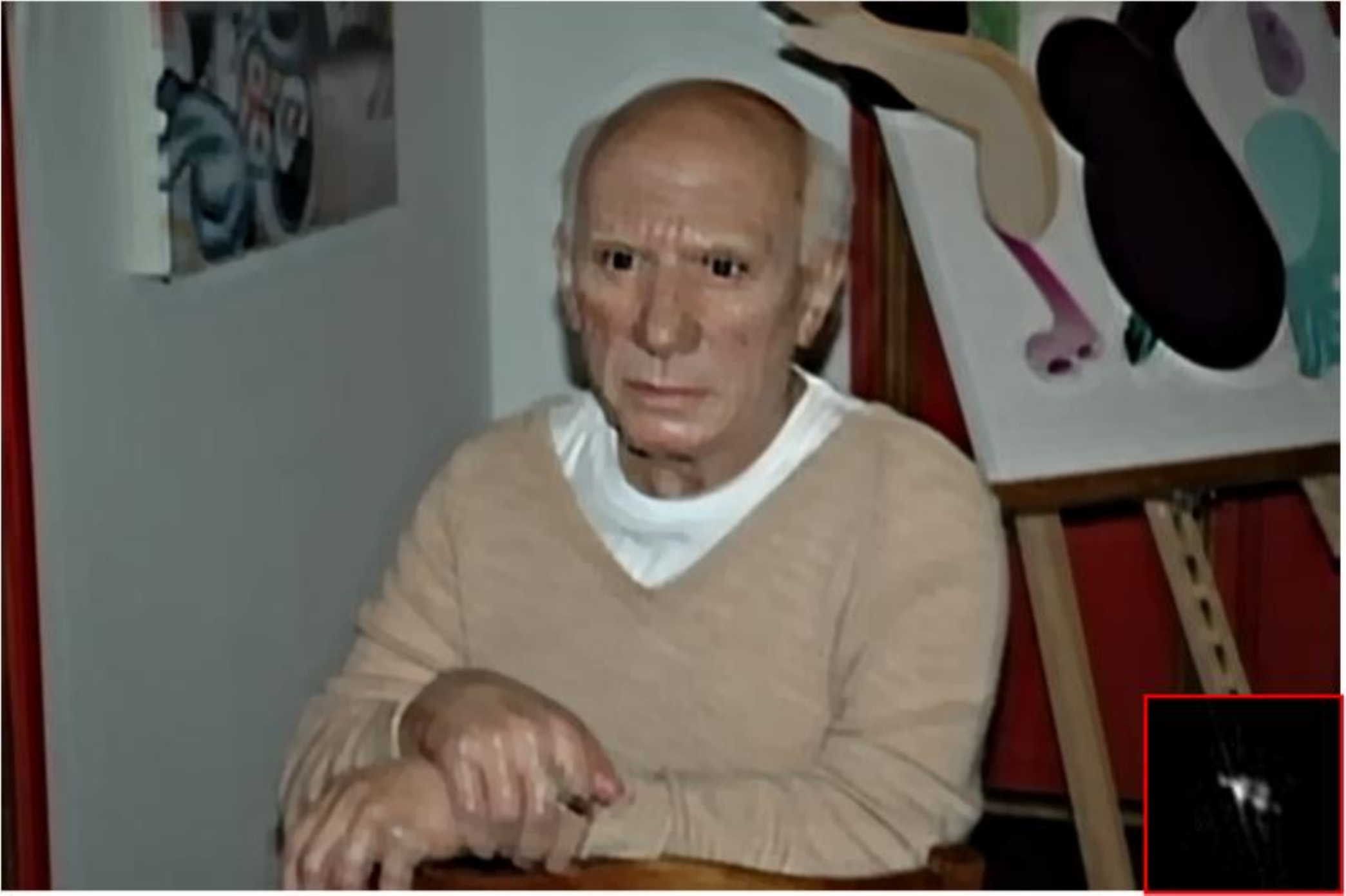}}\\
\subfloat[]{
\label{fig:picasso_levin}
\includegraphics[width=0.48\linewidth]{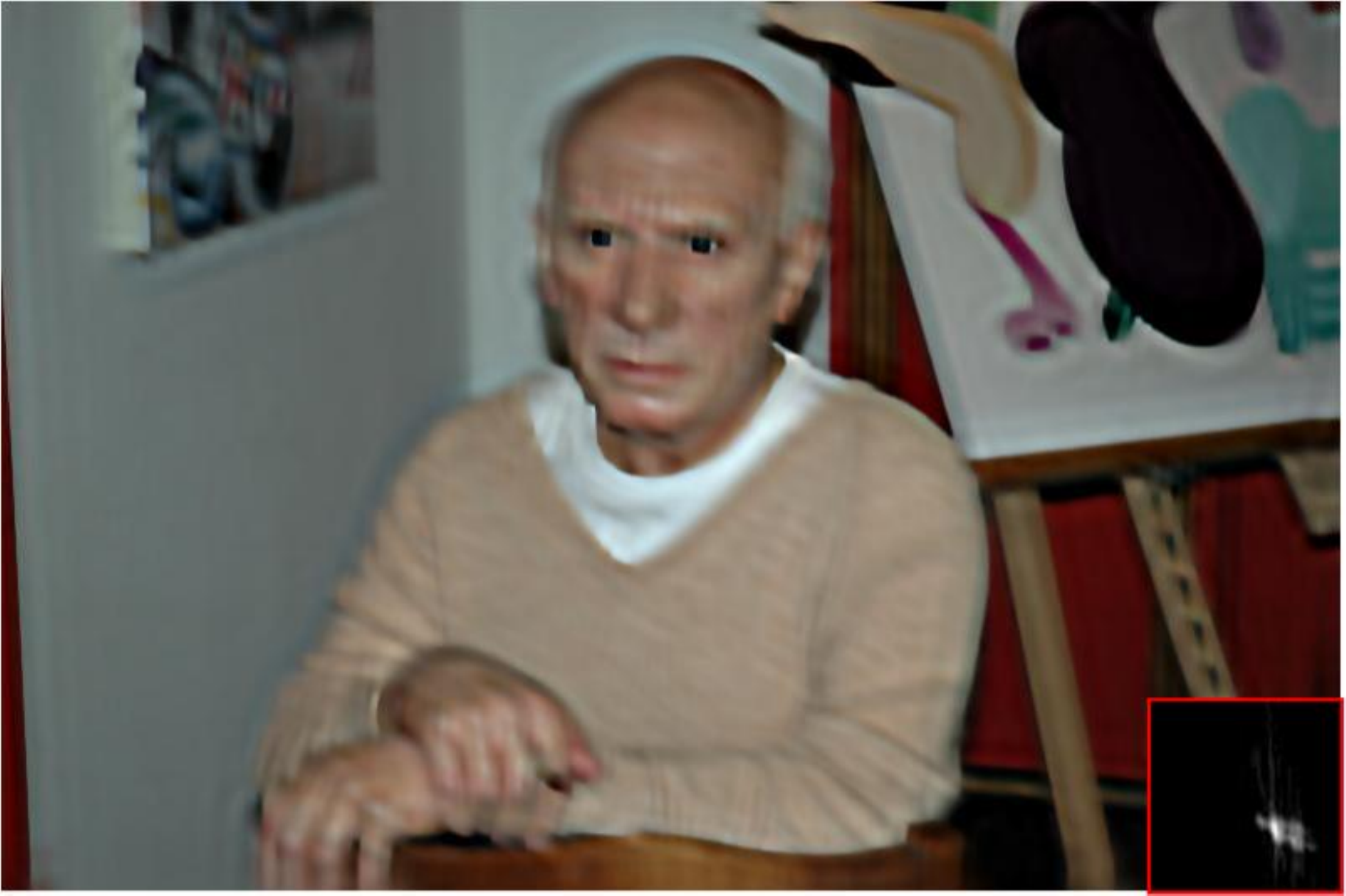}}
\subfloat[]{
\label{fig:picasso_michaeli}
\includegraphics[width=0.48\linewidth]{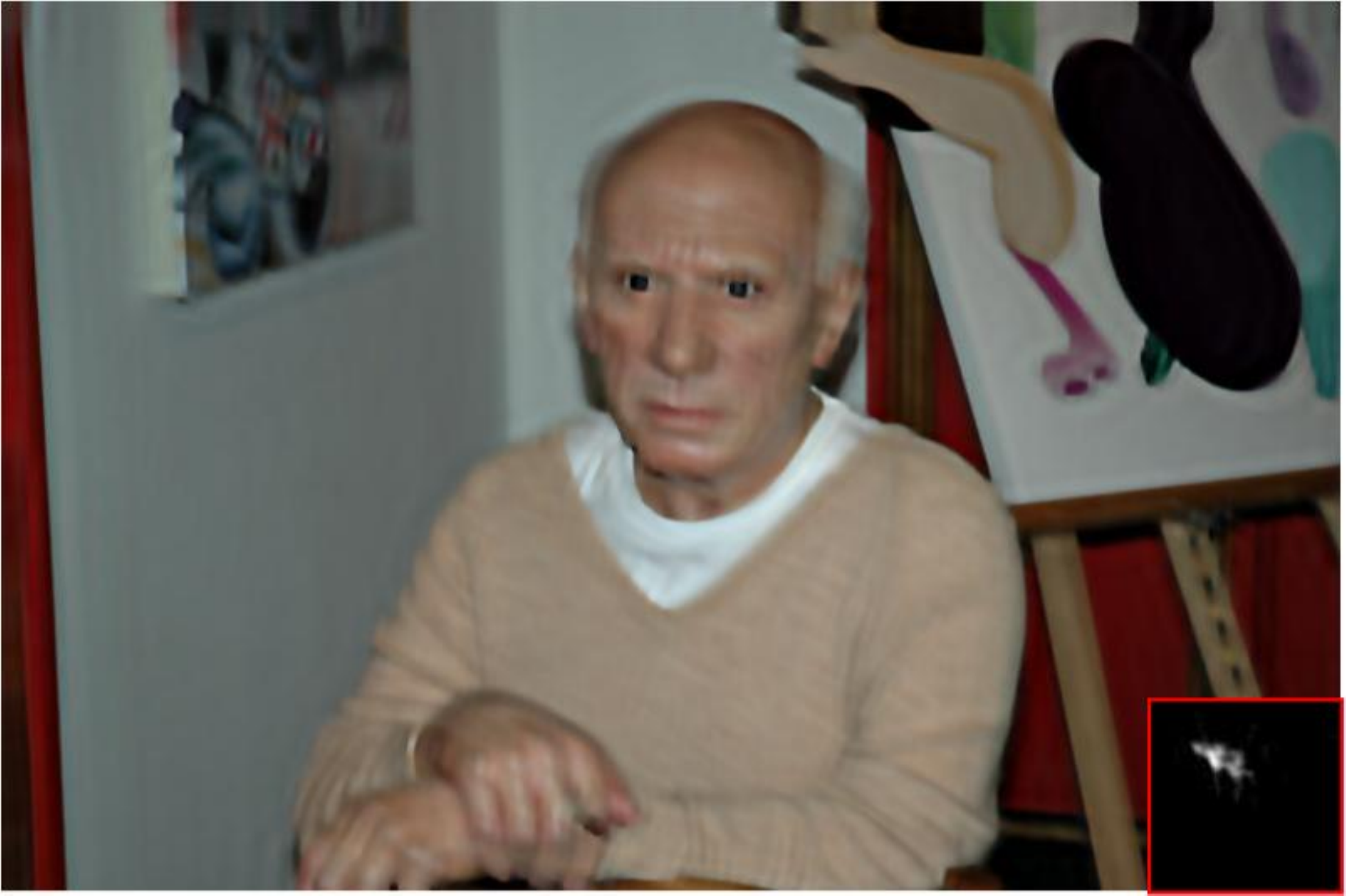}}\\
\subfloat[]{
\label{fig:picasso_pan}
\includegraphics[width=0.48\linewidth]{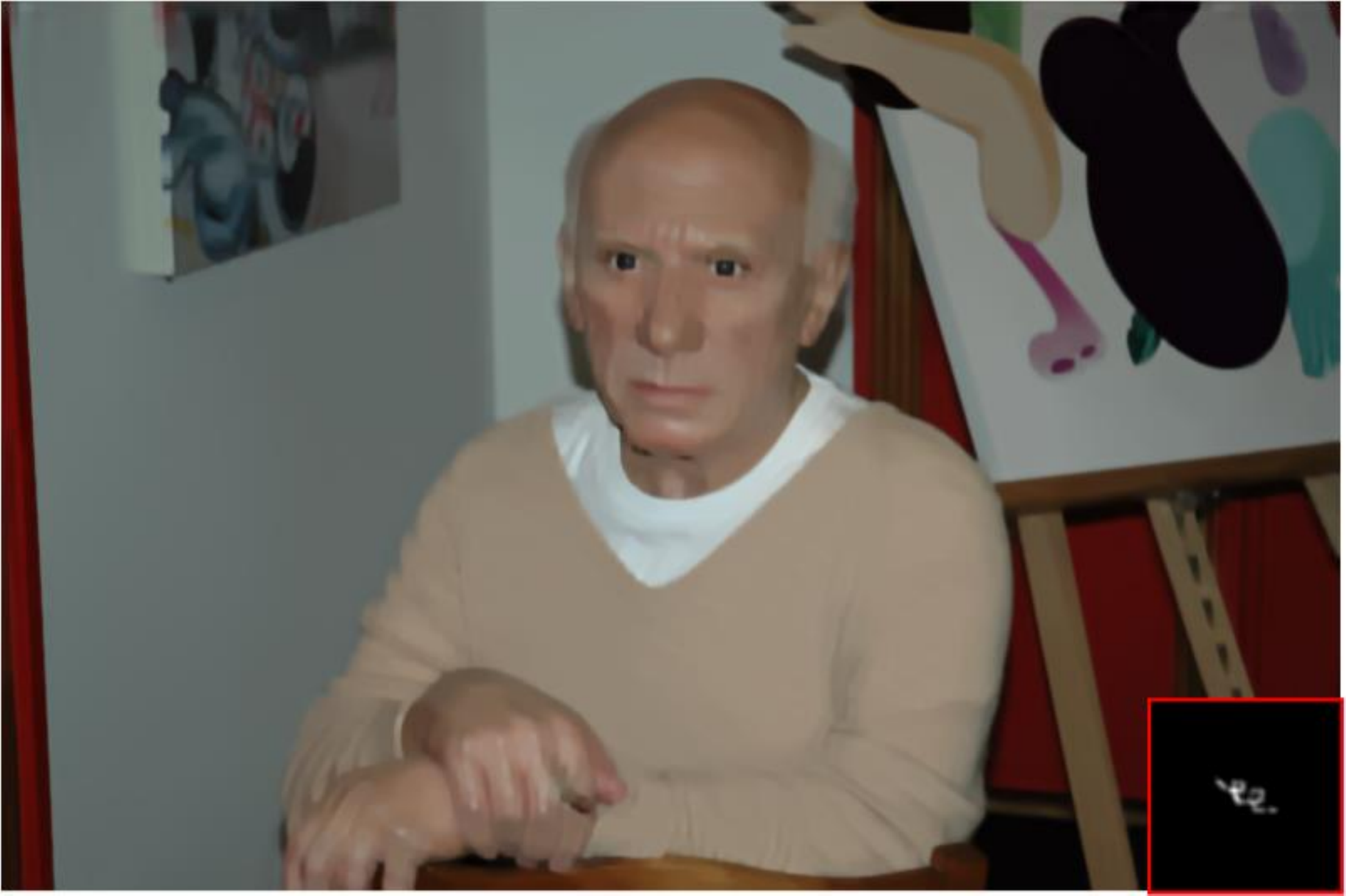}}
\subfloat[]{
\label{fig:picasso_ours}
\includegraphics[width=0.48\linewidth]{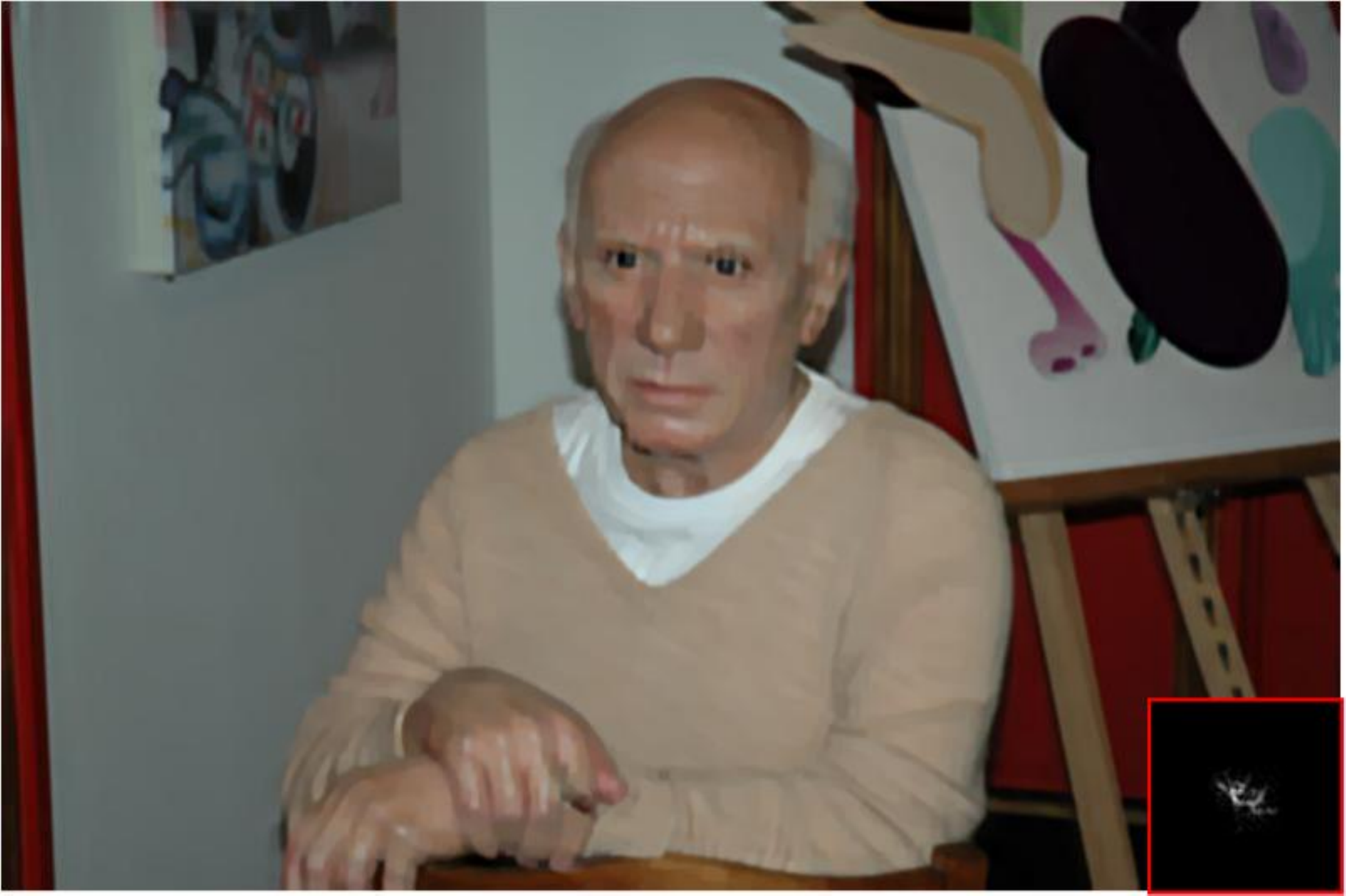}}
\caption{{\itshape Picasso.} Image Size: $800\times532$, Kernel Size: $69\times69$. (a) Blurry Image. (b) Krishnan \emph{et al.} \cite{Krishnan2011}. (c) Levin \emph{et al.} \cite{Levin2011EML}. (d) Michaeli \& Irani \cite{Michaeli2014}. (e) Pan \emph{et al.} \cite{Pan_2016_CVPR}. (f) Ours.}
\label{fig:picasso}
\end{figure}

\begin{figure}[htb]
\centering
\subfloat[]{
\label{fig:Pietro_blur}
\includegraphics[width=0.48\linewidth]{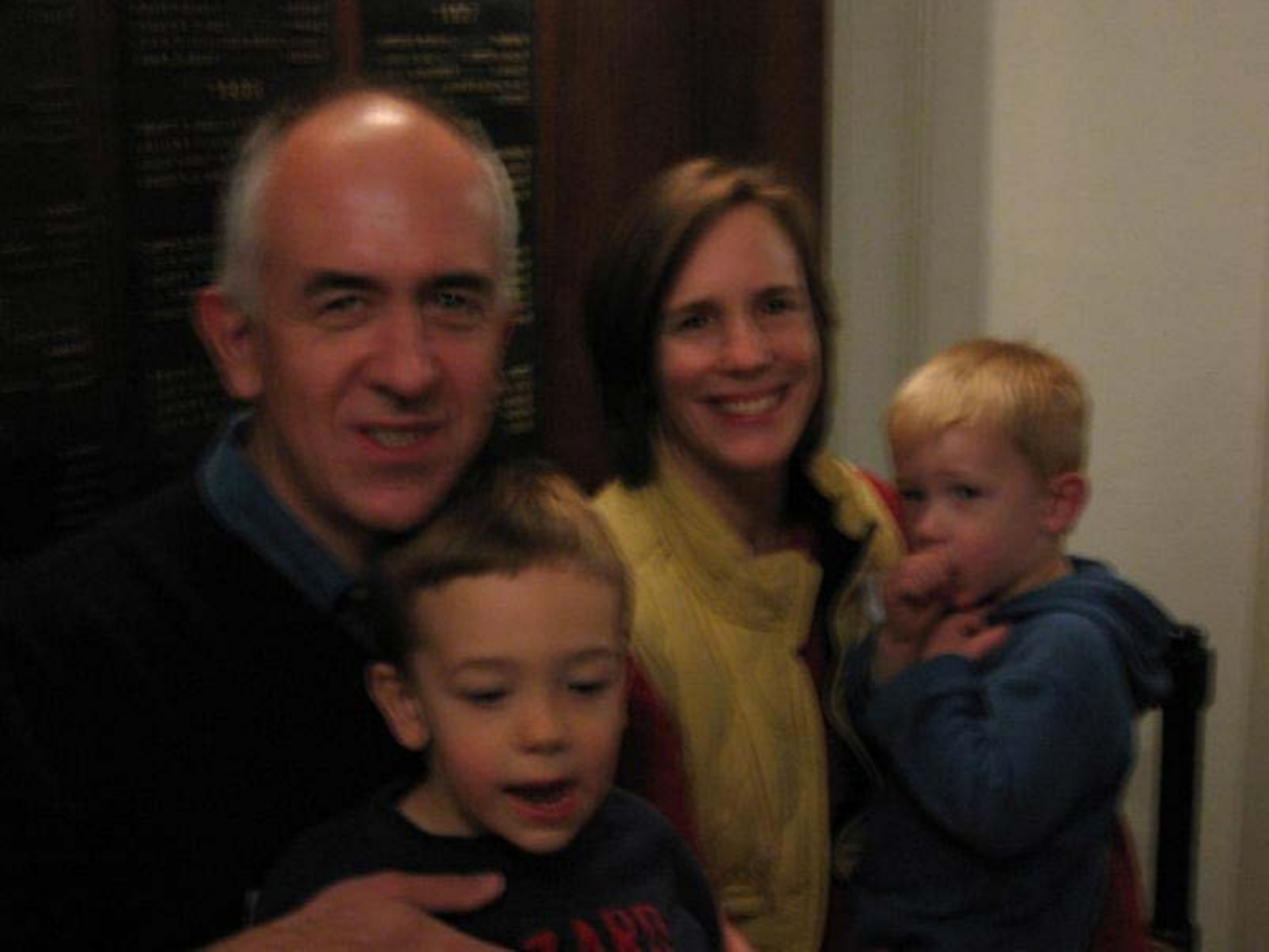}}
\subfloat[]{
\label{fig:Pietro_krishnan}
\includegraphics[width=0.48\linewidth]{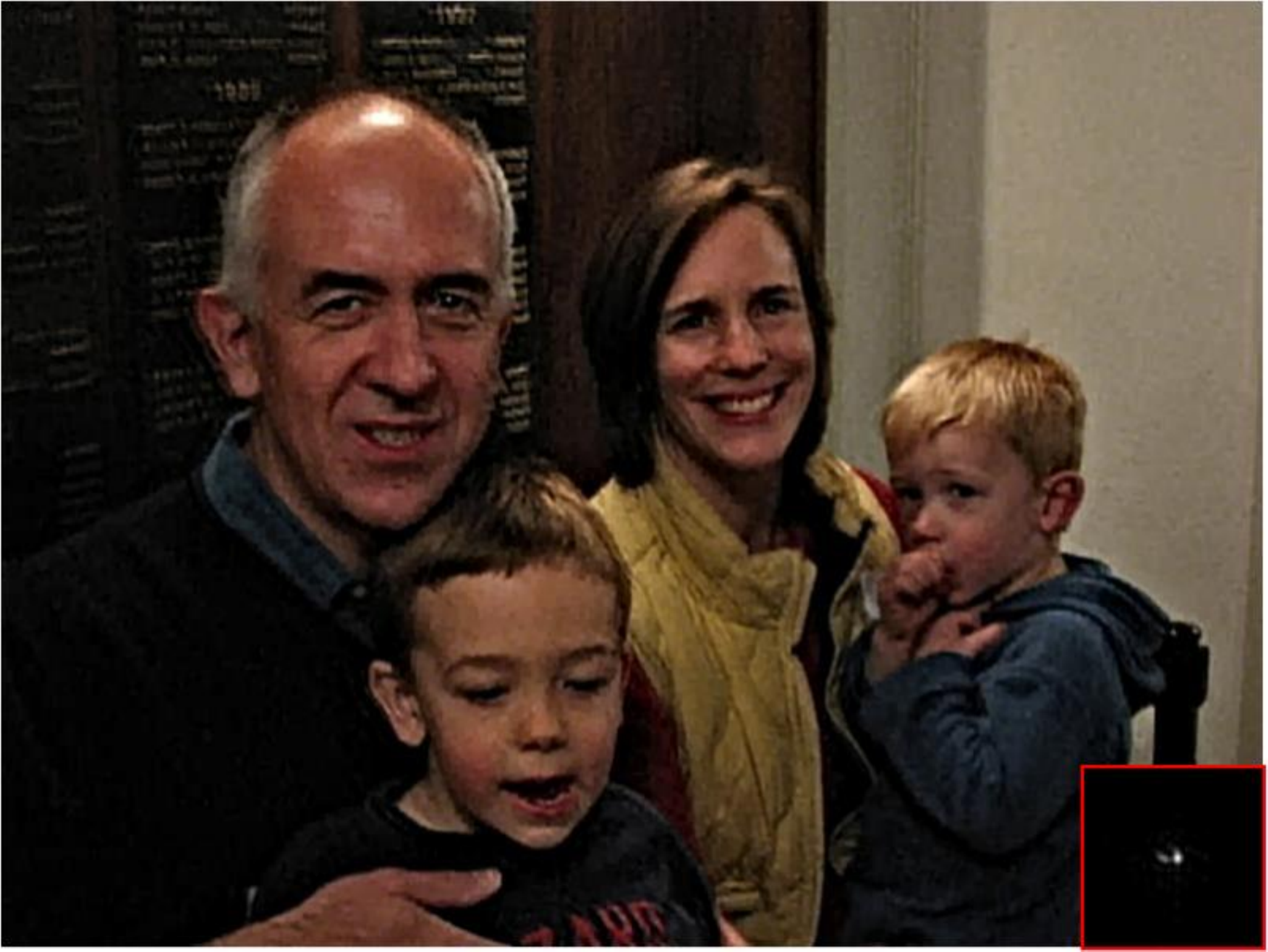}}\\
\subfloat[]{
\label{fig:Pietro_levin}
\includegraphics[width=0.48\linewidth]{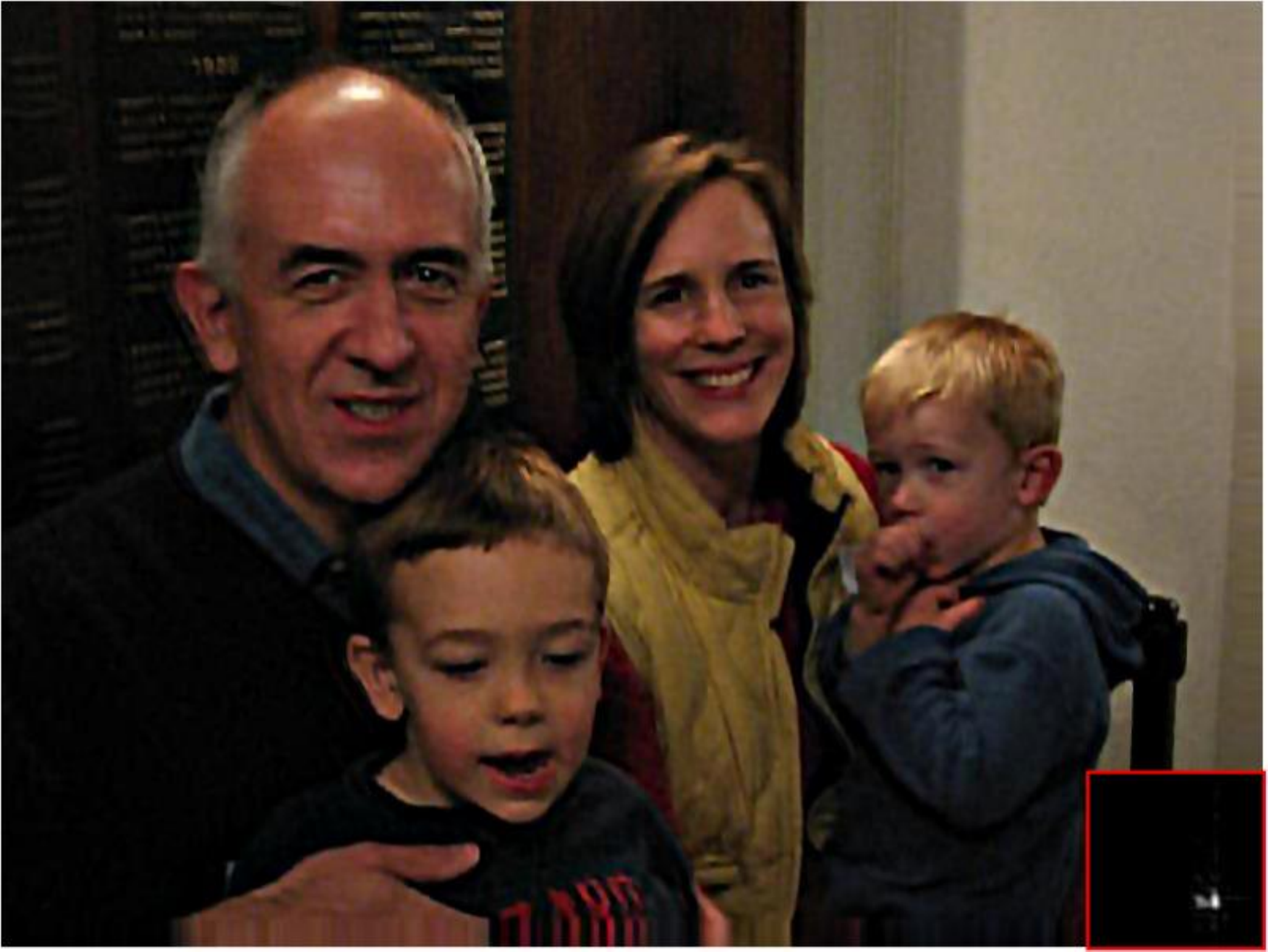}}
\subfloat[]{
\label{fig:Pietro_michaeli}
\includegraphics[width=0.48\linewidth]{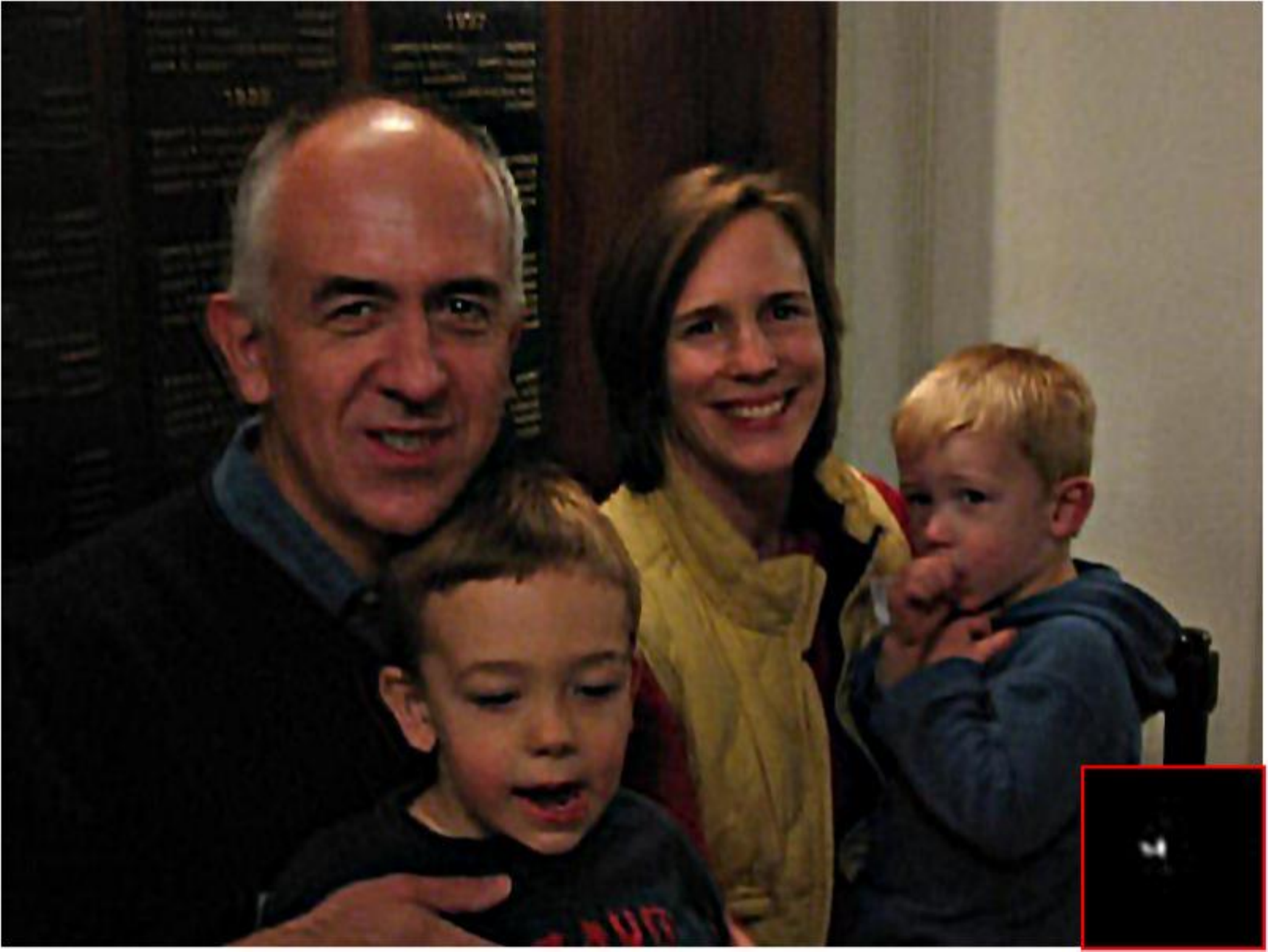}}\\
\subfloat[]{
\label{fig:Pietro_pan}
\includegraphics[width=0.48\linewidth]{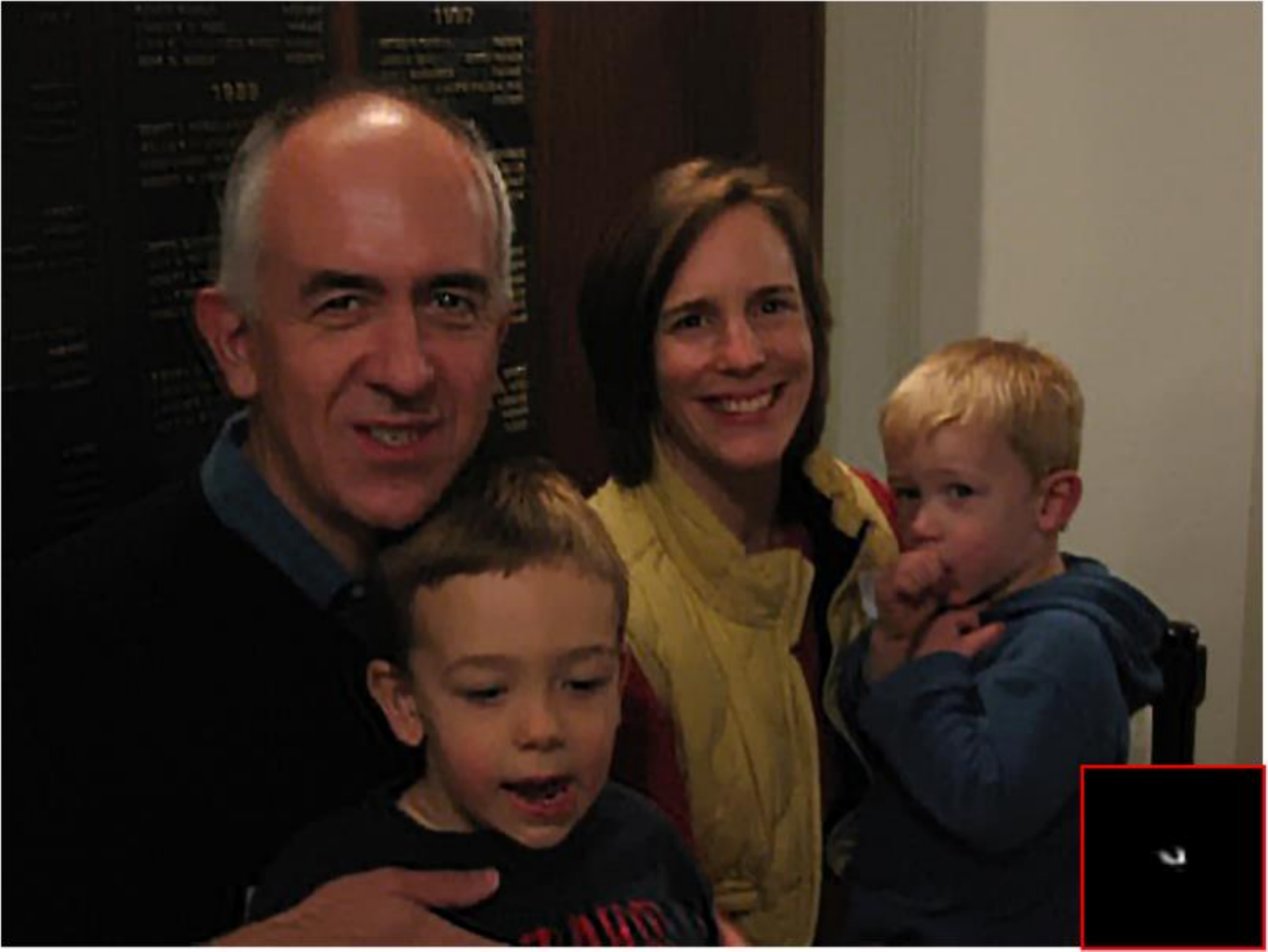}}
\subfloat[]{
\label{fig:Pietro_ours}
\includegraphics[width=0.48\linewidth]{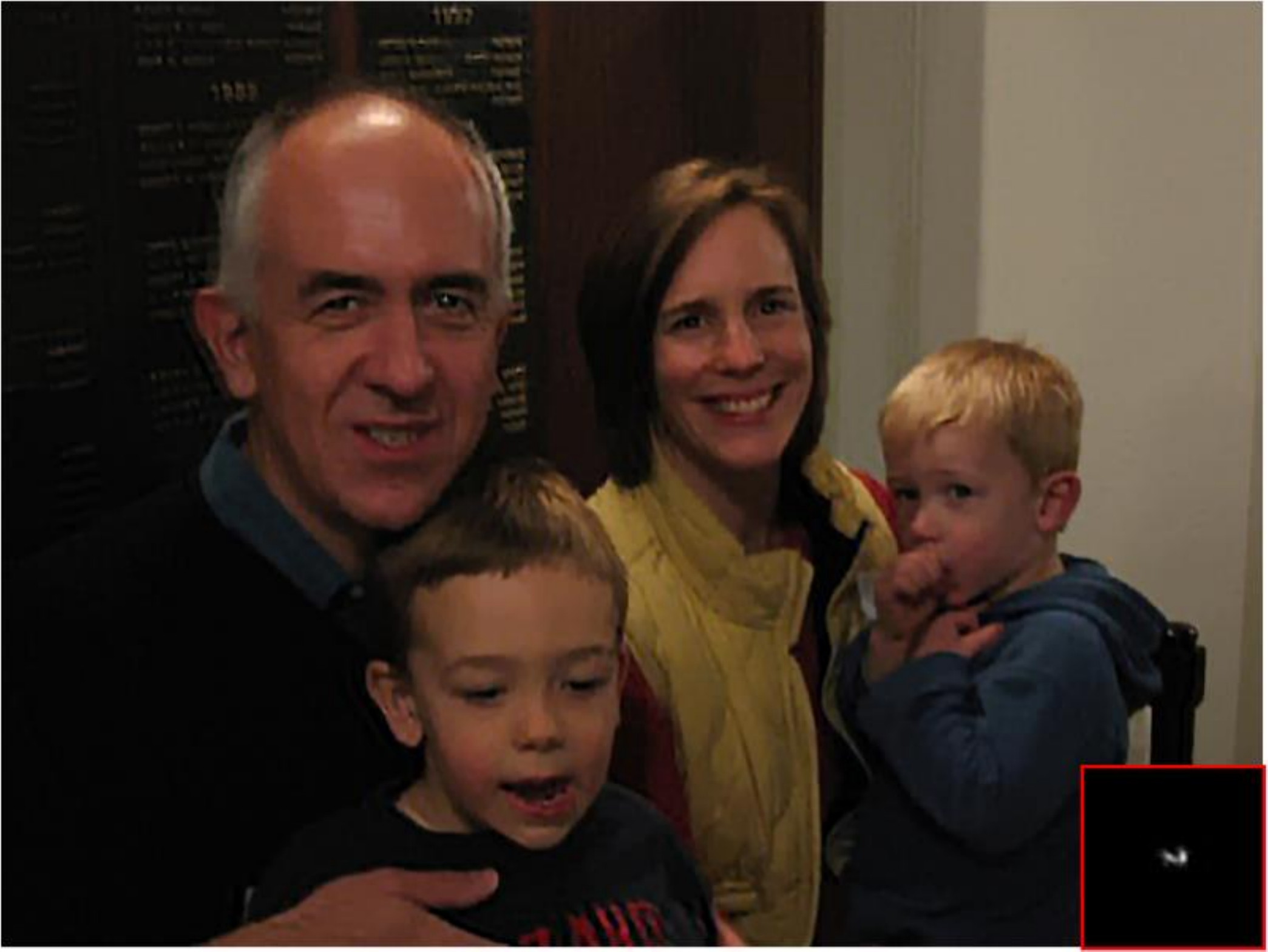}}
\caption{{\itshape Pietro.} Image Size: $800\times600$, Kernel Size: $69\times69$. (a) Blurry Image. (b) Krishnan \emph{et al.} \cite{Krishnan2011}. (c) Levin \emph{et al.} \cite{Levin2011EML}. (d) Michaeli \& Irani \cite{Michaeli2014}. (e) Pan \emph{et al.} \cite{Pan_2016_CVPR}. (f) Ours.}
\label{fig:Pietro}
\end{figure}

\begin{figure}[!t]
\centering
\subfloat[]{
\label{fig:roma_blur}
\includegraphics[width=0.48\linewidth]{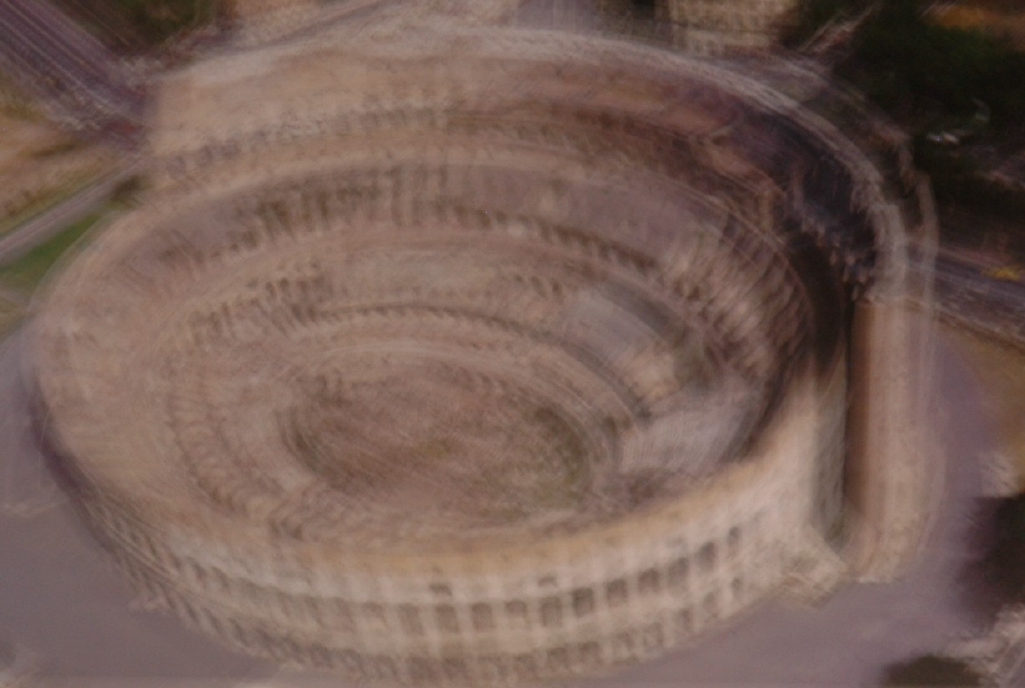}}
\subfloat[]{
\label{fig:roma_krishnan}
\includegraphics[width=0.48\linewidth]{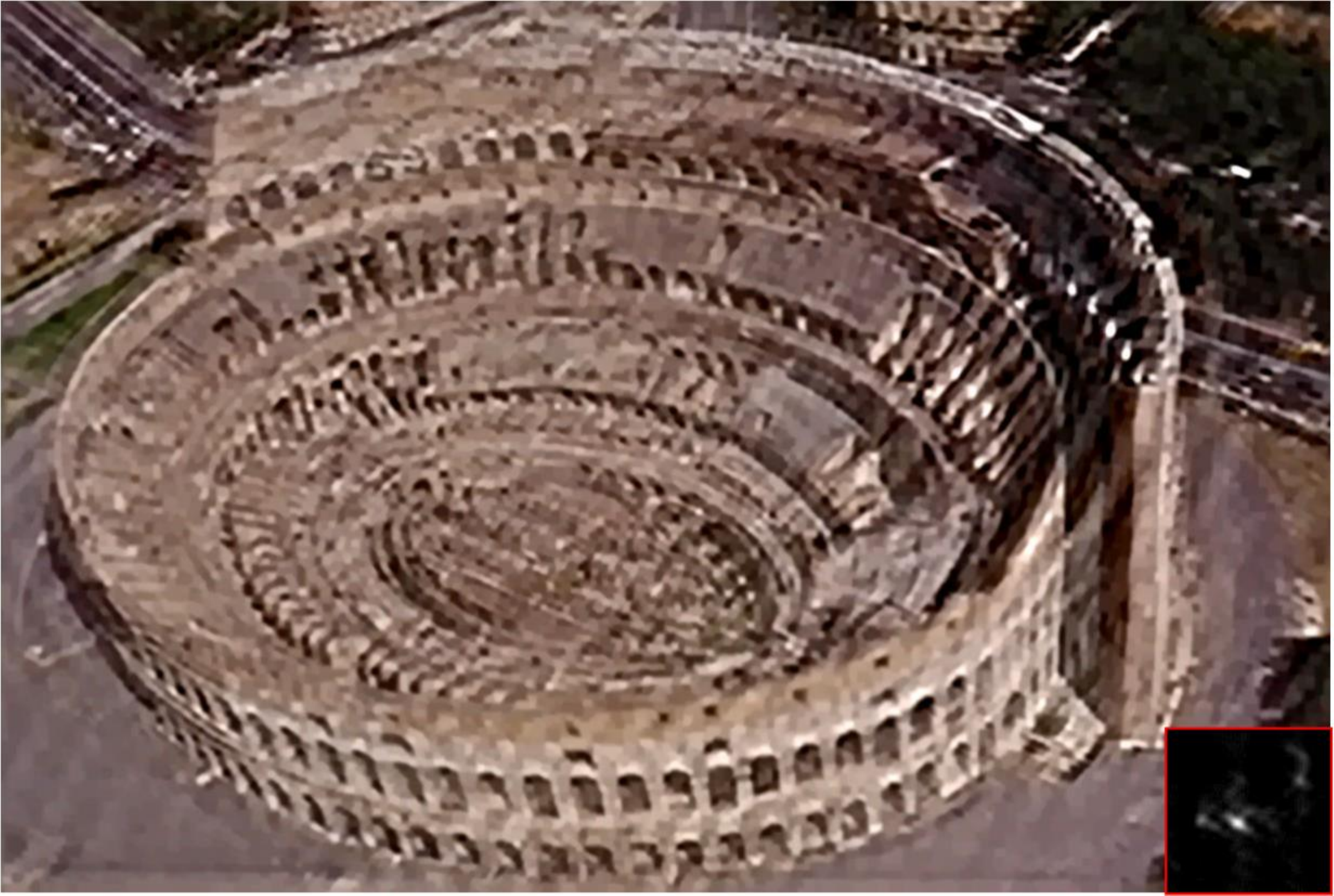}}\\
\subfloat[]{
\label{fig:roma_levin}
\includegraphics[width=0.48\linewidth]{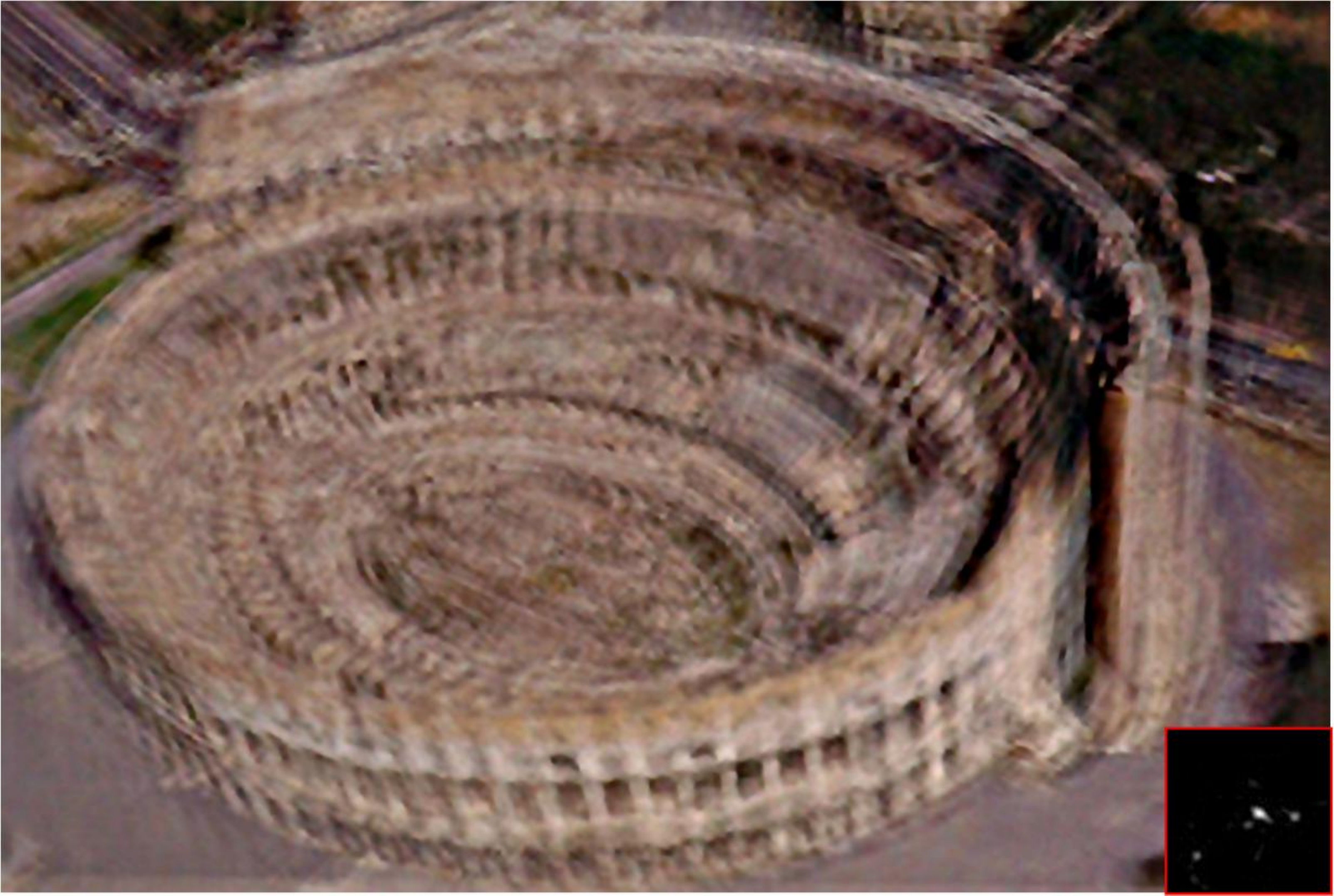}}
\subfloat[]{
\label{fig:roma_michaeli}
\includegraphics[width=0.48\linewidth]{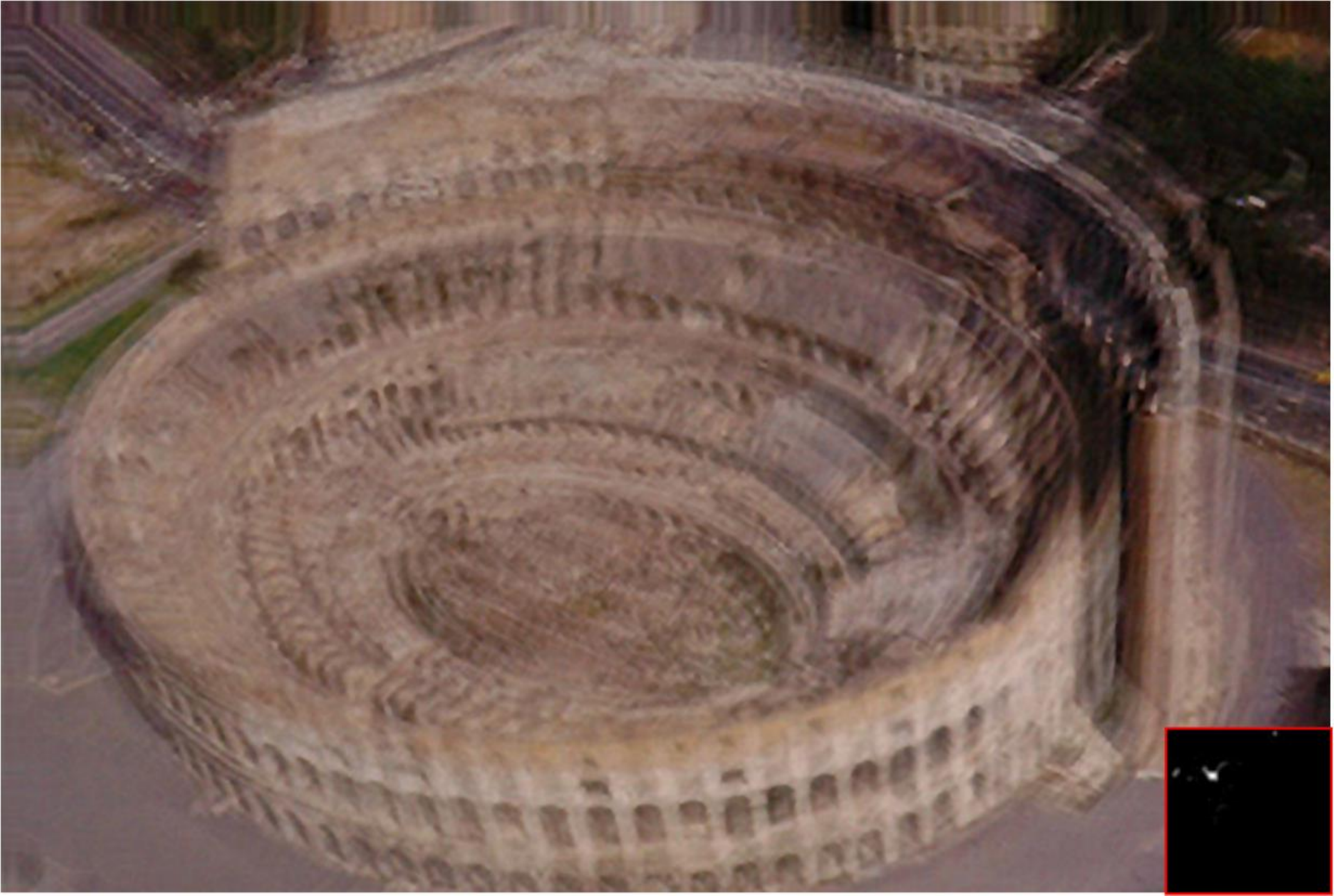}}\\
\subfloat[]{
\label{fig:roma_pan}
\includegraphics[width=0.48\linewidth]{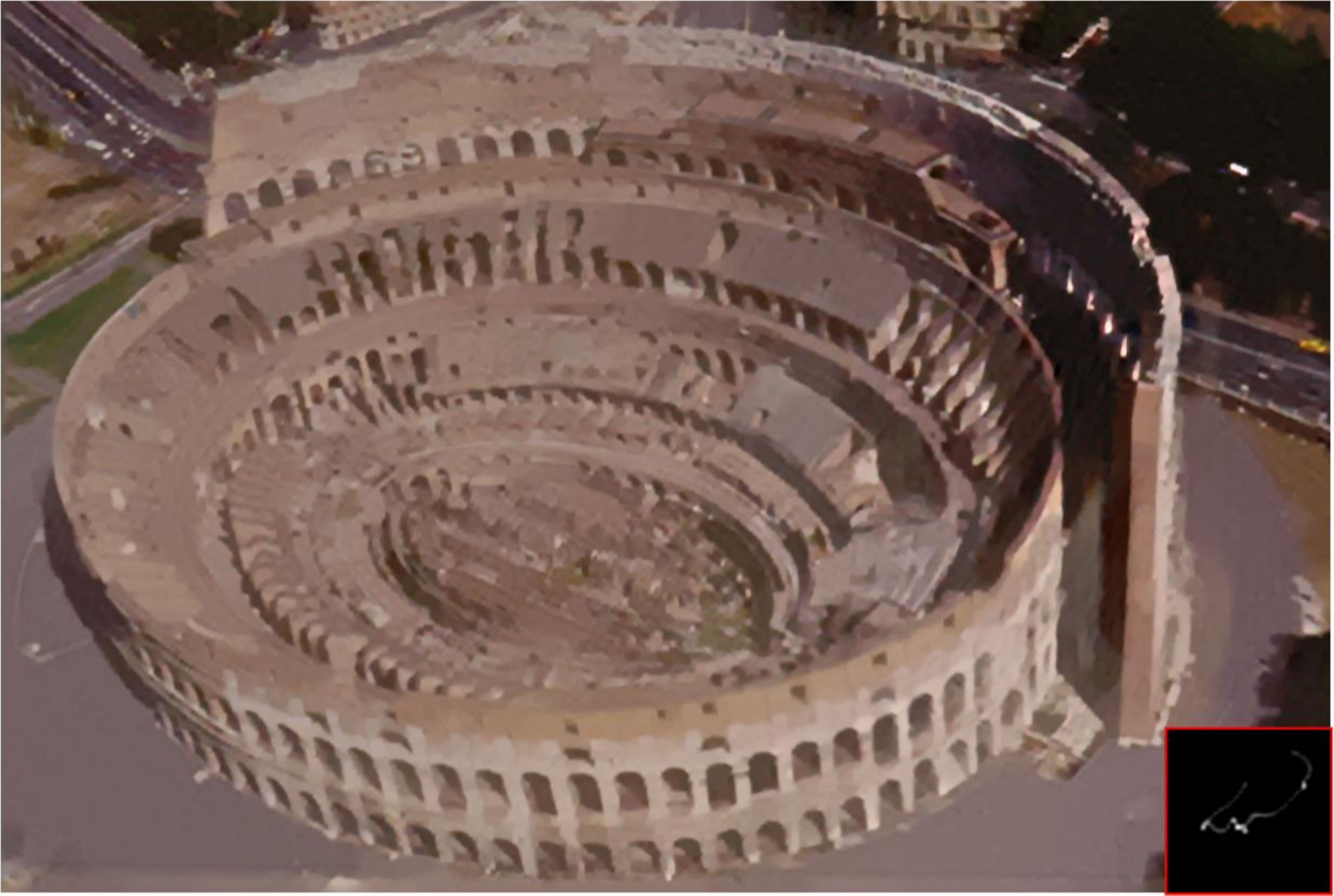}}
\subfloat[]{
\label{fig:roma_ours}
\includegraphics[width=0.48\linewidth]{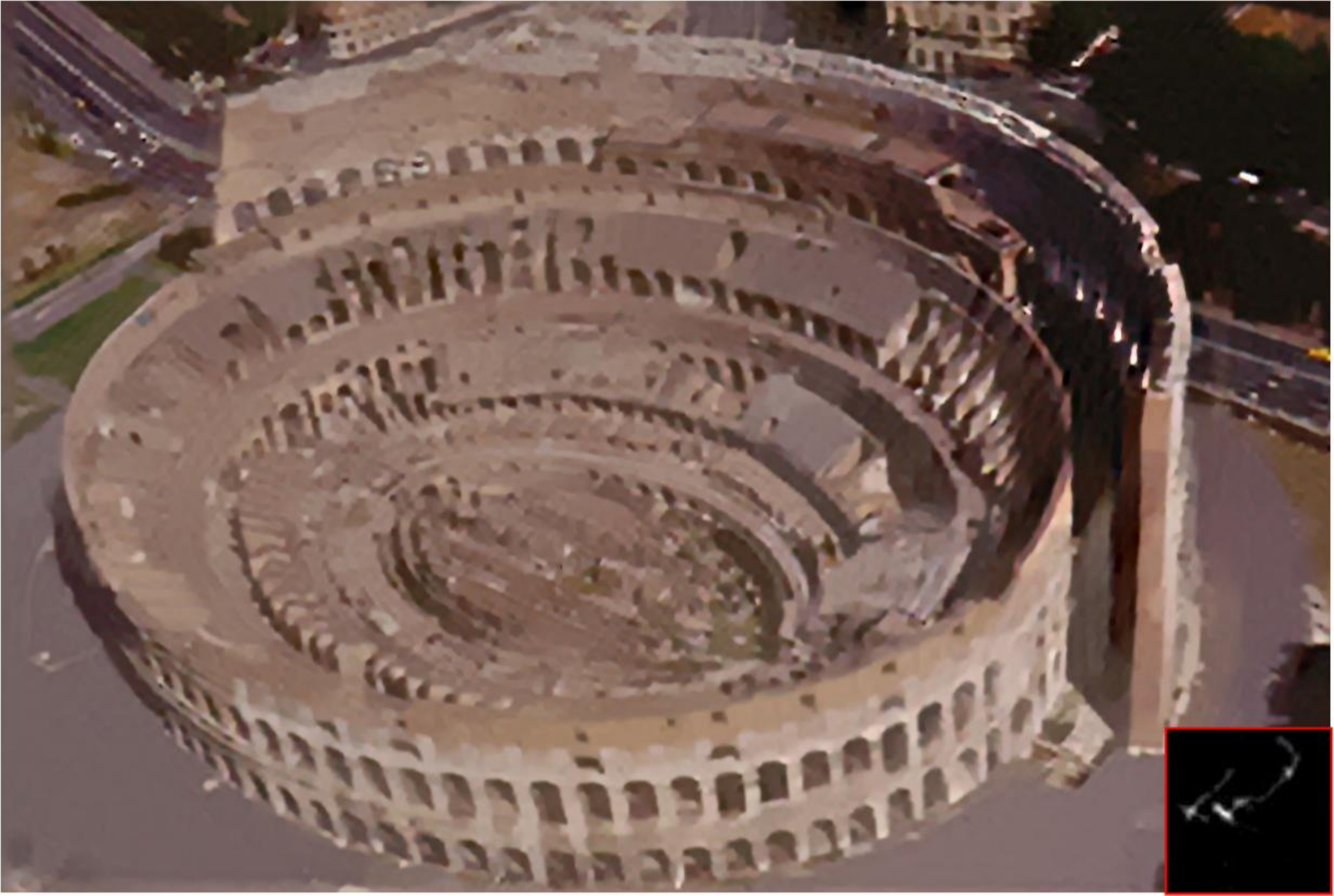}}
\caption{{\itshape Roma.} Image Size: $1229\times825$, Kernel Size: $95\times95$. (a) Blurry Image. (b) Krishnan \emph{et al.} \cite{Krishnan2011}. (c) Levin \emph{et al.} \cite{Levin2011EML}. (d) Michaeli \& Irani \cite{Michaeli2014}. (e) Pan \emph{et al.} \cite{Pan_2016_CVPR}. (f) Ours.}
\label{fig:roma}
\end{figure}

In the second part of this section, we apply our algorithms on real blurry images and compare with algorithms, of which the implementations are available, \emph{i.e.}, Krishnan \emph{et al.} \cite{Krishnan2011}, Levin \emph{et~al.} \cite{Levin2011EML}, Michaeli \& Irani \cite{Michaeli2014} and Pan \emph{et al.} \cite{Pan_2016_CVPR}.
In Fig.~\ref{fig:flower1}, \ref{fig:picasso}, \ref{fig:Pietro} and \ref{fig:roma}, we blindly deblur an image with different depths of field, a human portrait with a complex background, an image full of humans, and an image with very severe motion blur, respectively. Our algorithm estimates the blur kernels robustly and results in less artifacts in the recovered images.

Table \ref{tb:time} reports the running time of the competing algorithms and ours on real blurry images in Fig. \ref{fig:flower1}, \ref{fig:picasso}, \ref{fig:Pietro} and \ref{fig:roma}. All the algorithms run on the Matlab platform. According to Table \ref{tb:time}, our algorithm runs much faster than other competing algorithms.
The efficiency of our algorithm can be further improved with C++ and GPU implementation.

\begin{table}[!t]
\caption{Running time of different algorithms on real images.}
\label{tb:time}
\centering
\footnotesize
\begin{tabular}{|p{3em}<{\centering}|p{3em}<{\centering}p{4em}<{\centering}p{4em}<{\centering}p{3em}<{\centering}p{3em}<{\centering}|}
\hline
  Image & \cite{Krishnan2011} & \cite{Levin2011EML} & \cite{Michaeli2014} & \cite{Pan_2016_CVPR} & Ours \\
\hline
\hline
{\itshape Flower}&  170 s & 142 min & 276 min & 14 min & \textbf{74 s}      \\
{\itshape Picasso}&  189 s & 148 min & 292 min & 17 min & \textbf{102 s}      \\
{\itshape Pietro}&  248 s & 153 min & 306 min & 19 min & \textbf{103 s}     \\
{\itshape Roma}&  463 s & 667 min & 2167 min & 41 min & \textbf{282 s}     \\
\hline
\end{tabular}

\end{table}

\subsection{Non-Uniform Blurry Images}
\begin{figure}[!t]
\centering
\subfloat[]{
\label{fig:elephant_blur}
\includegraphics[width=0.48\linewidth]{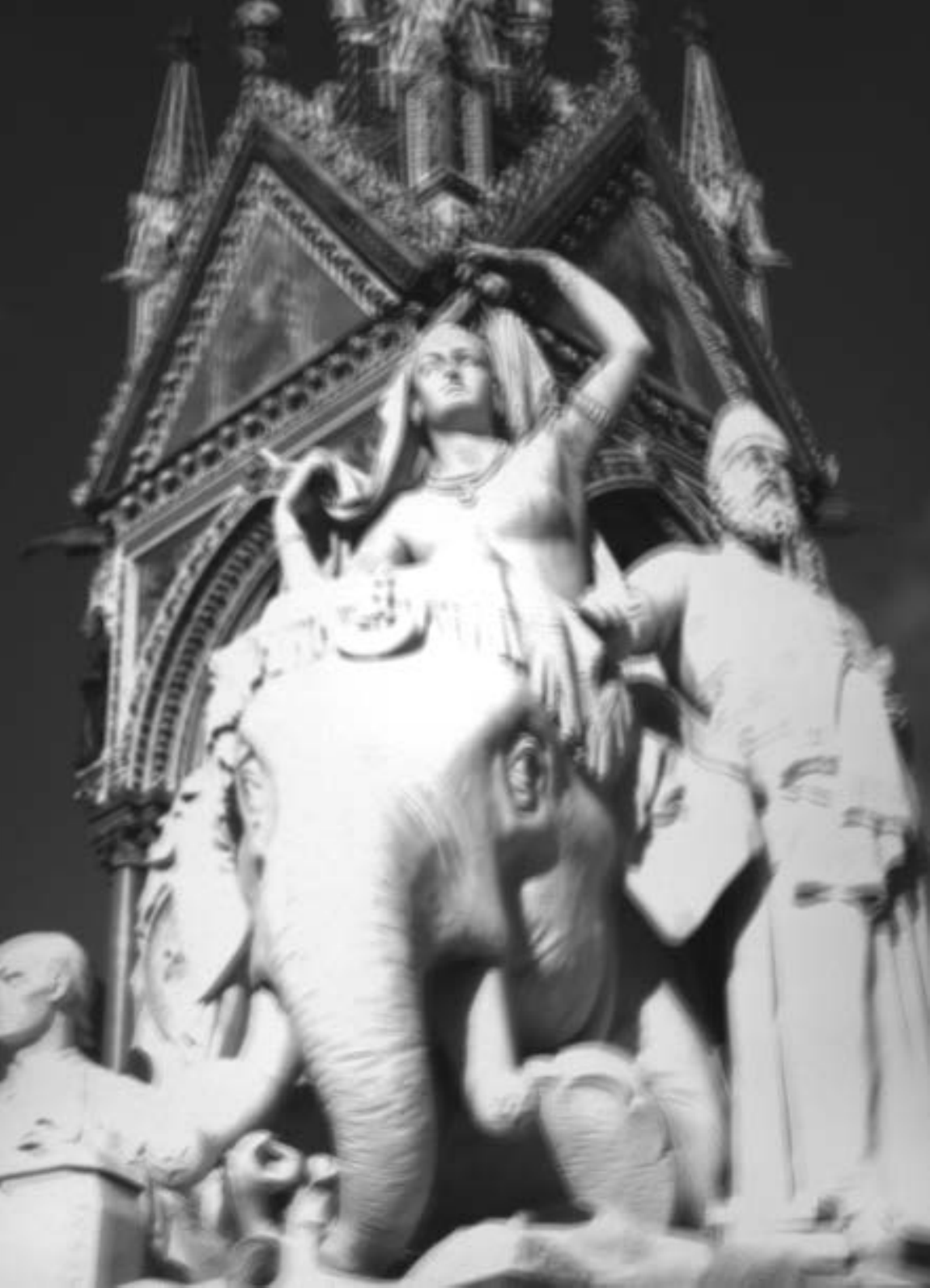}}
\subfloat[]{
\label{fig:elephant_2d}
\includegraphics[width=0.48\linewidth]{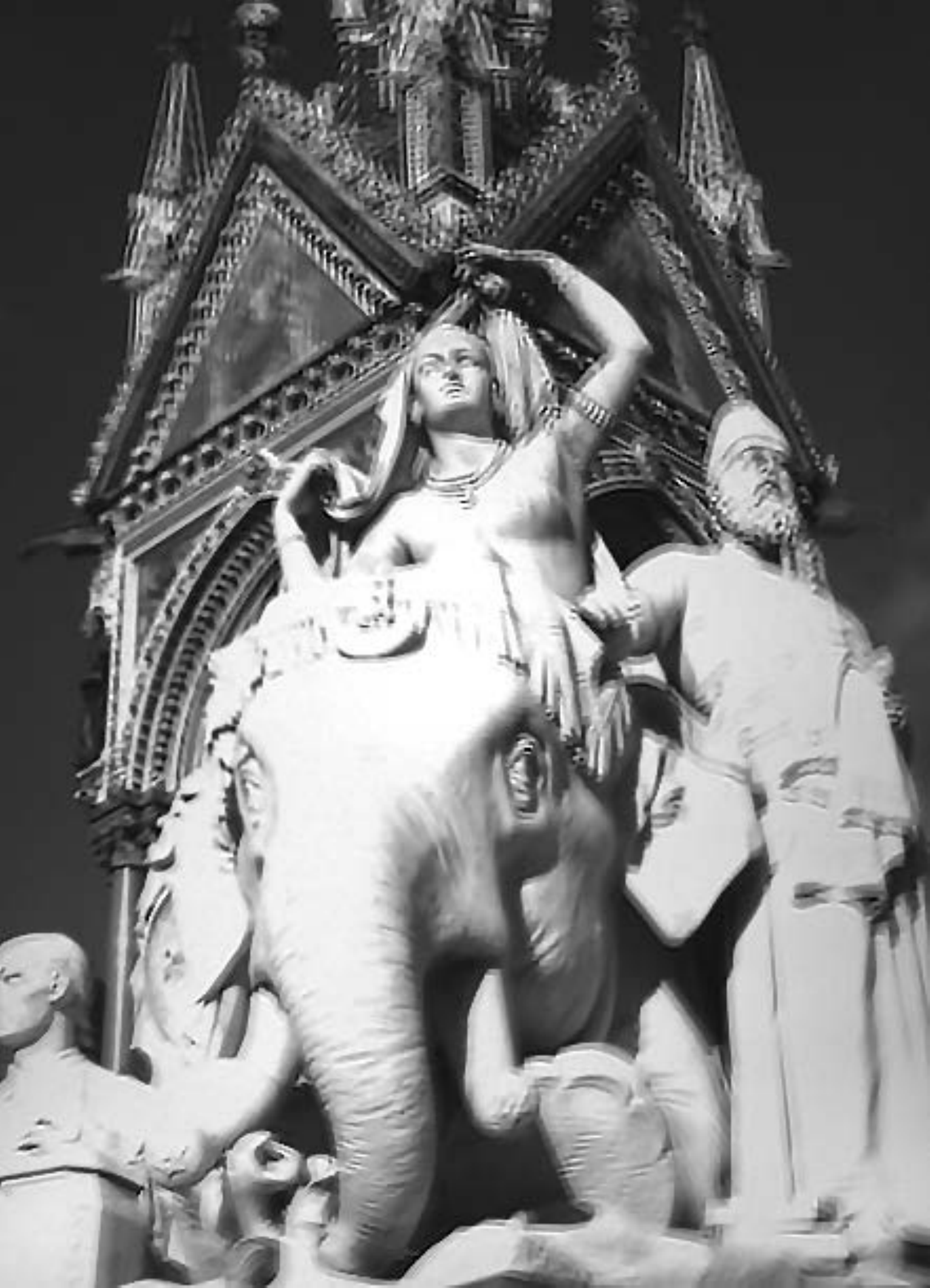}}\\
\subfloat[]{
\label{fig:elephant_harmeling}
\includegraphics[width=0.48\linewidth]{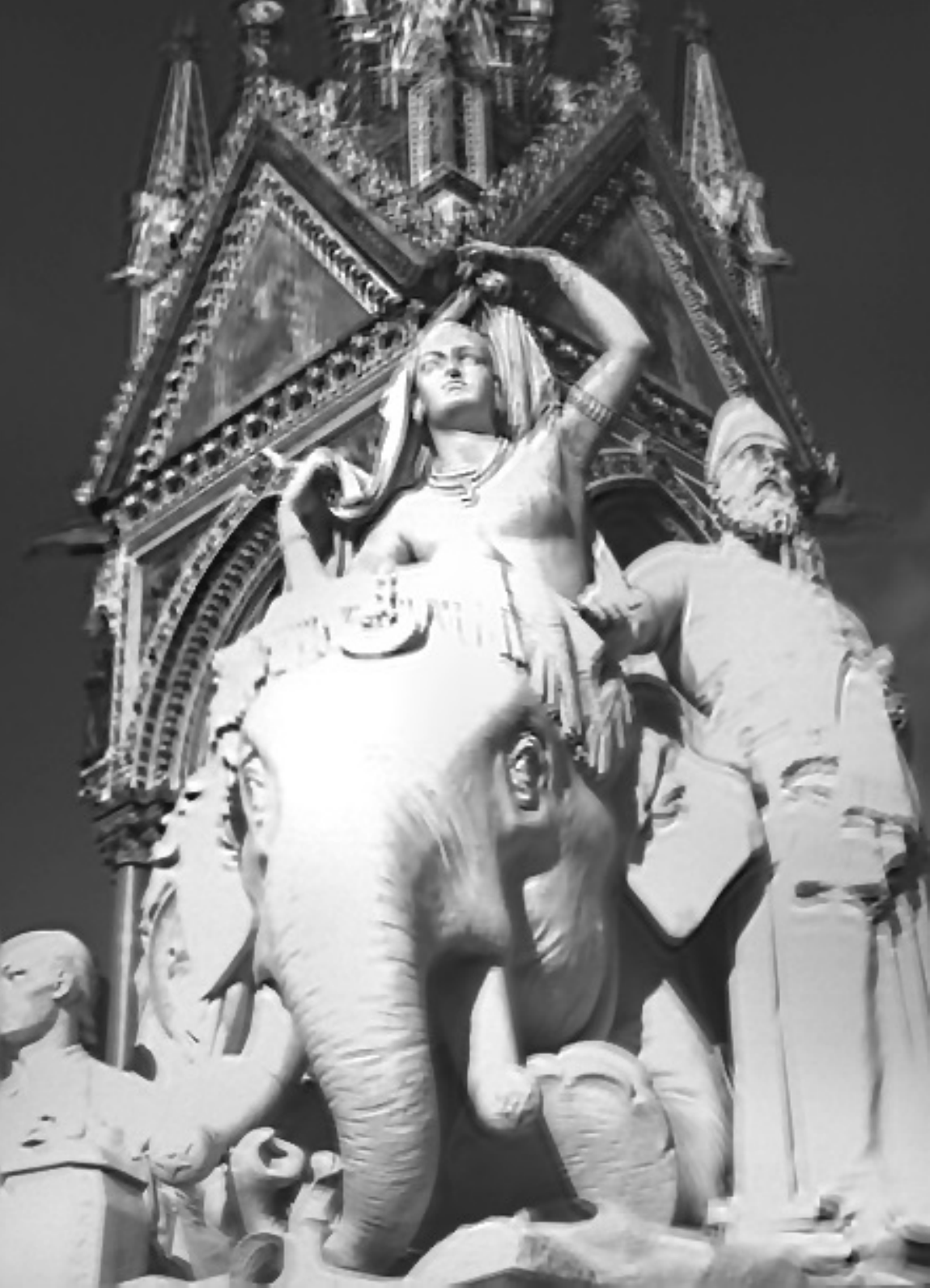}}
\subfloat[]{
\label{fig:elephant_hirsch}
\includegraphics[width=0.48\linewidth]{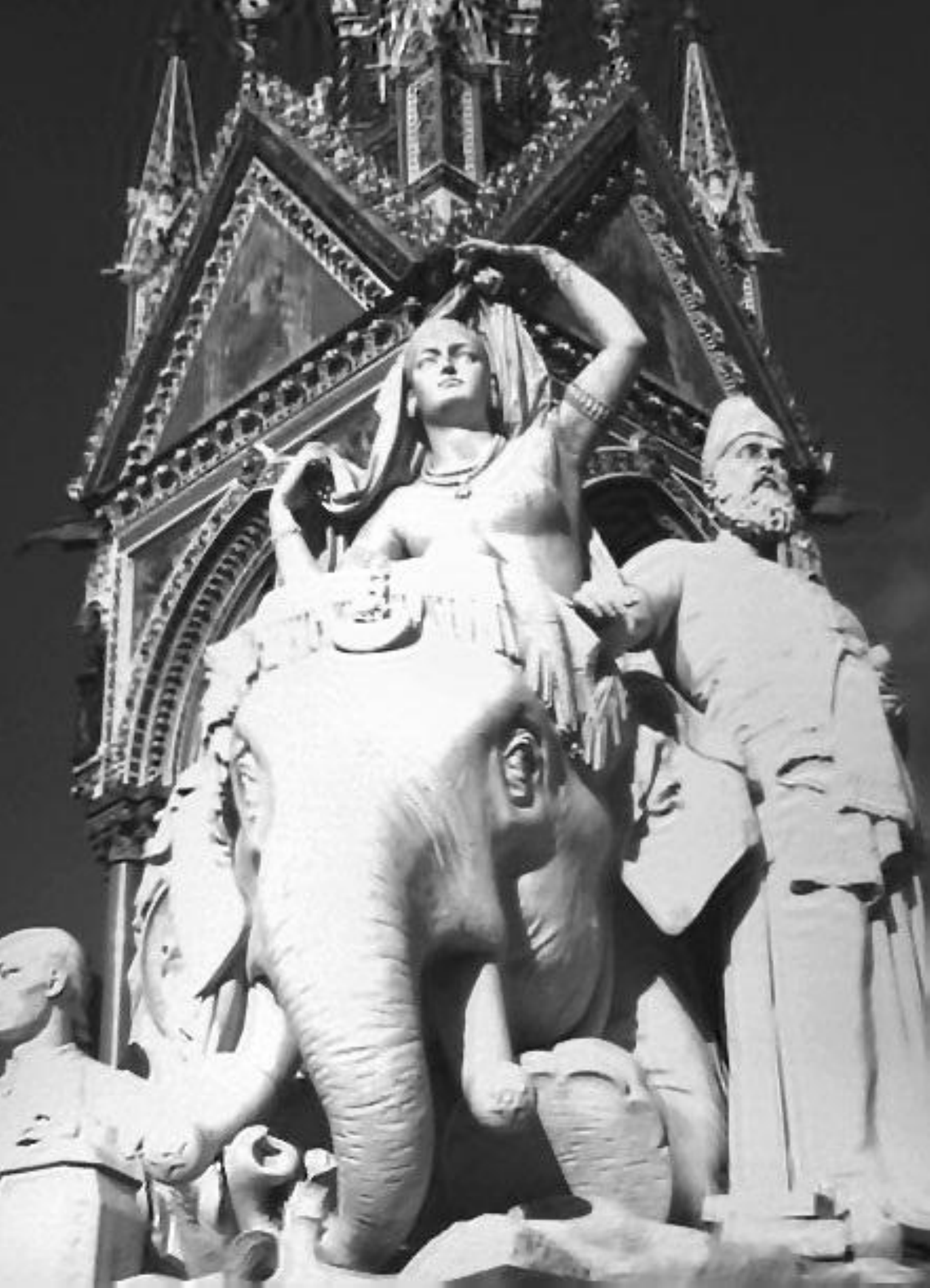}}\\
\subfloat[]{
\label{fig:elephant_ours}
\includegraphics[width=0.48\linewidth]{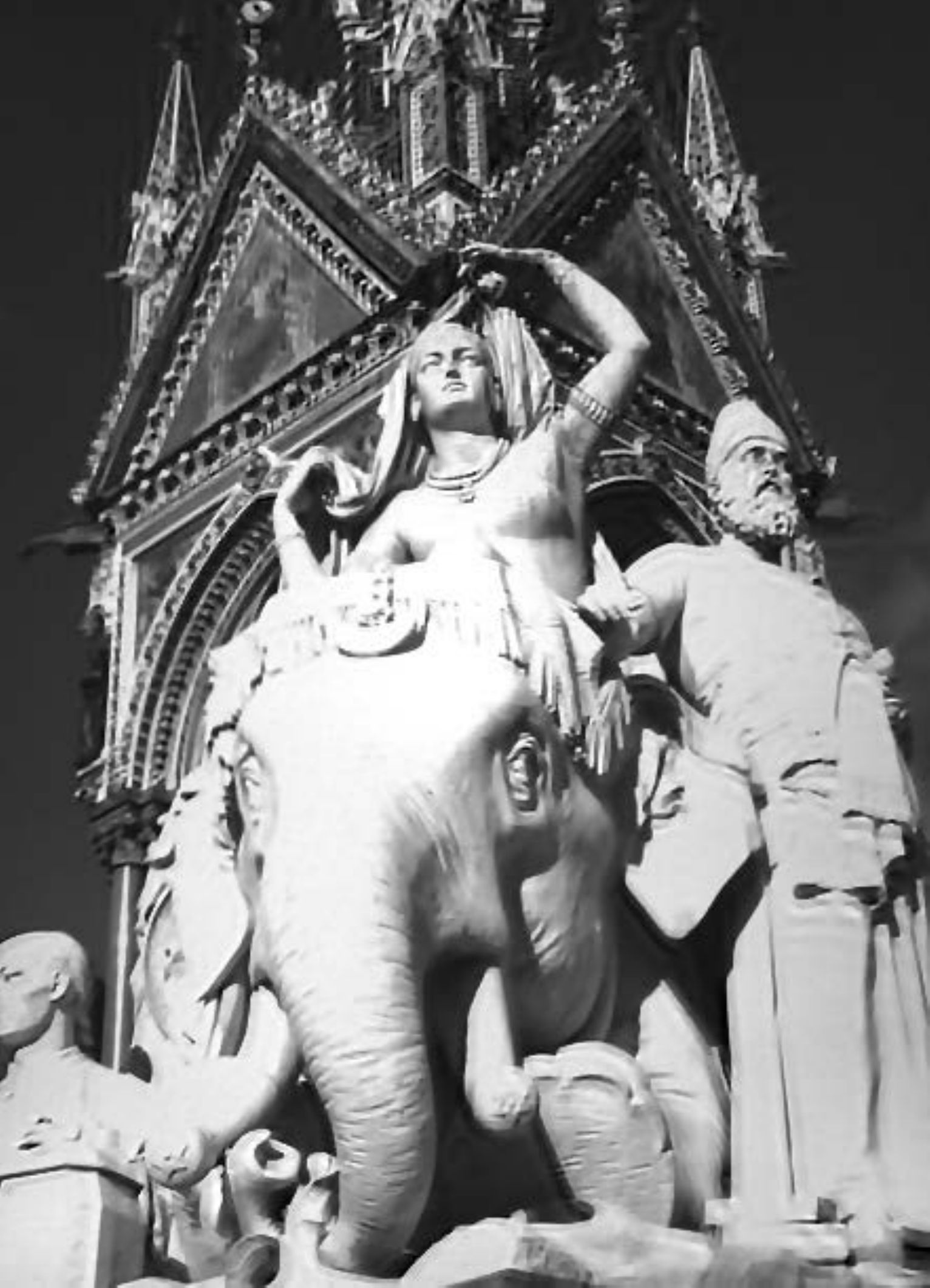}}
\subfloat[]{
\label{fig:elephant_kernel}
\includegraphics[width=0.48\linewidth]{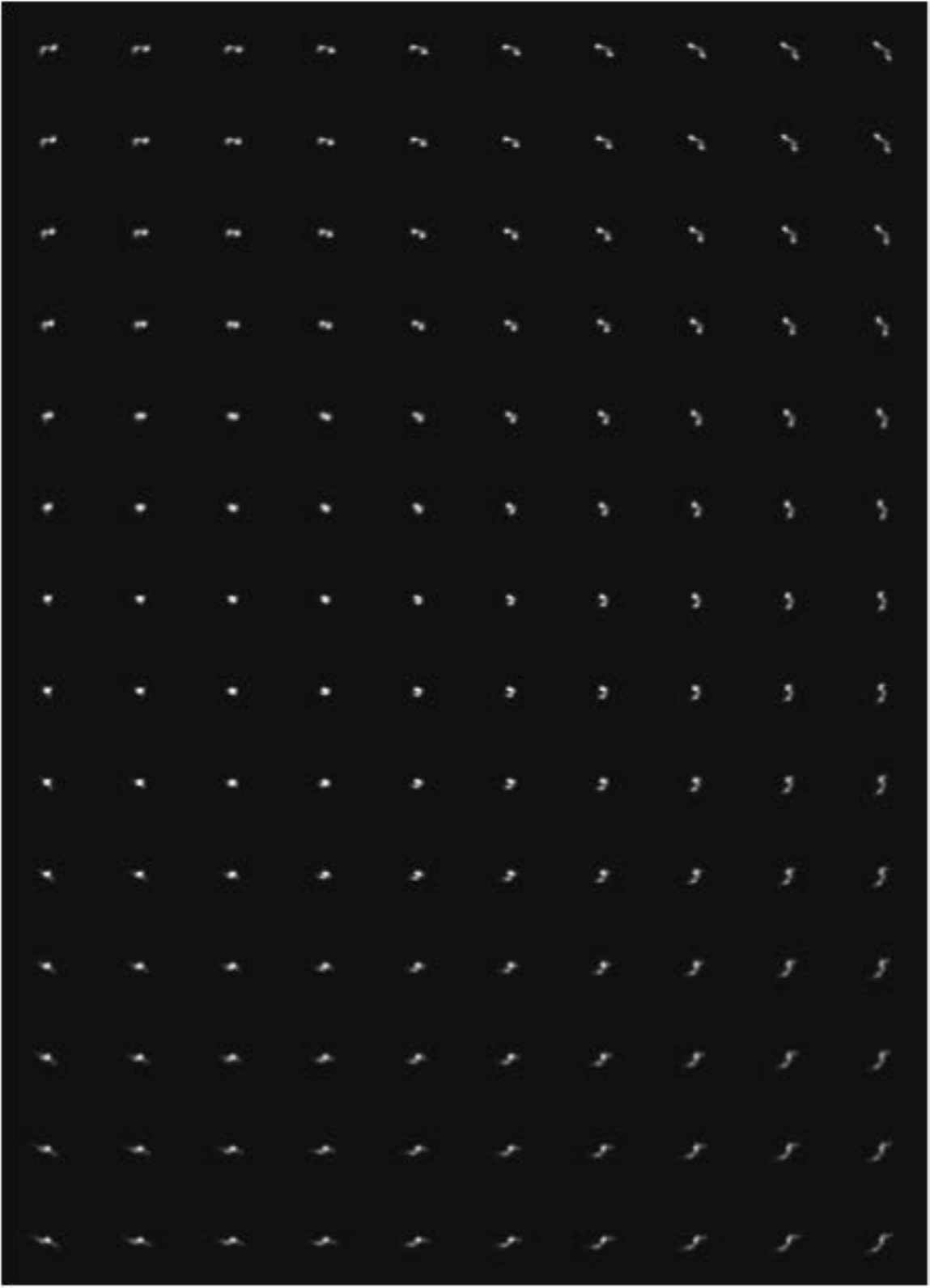}}
\caption{{\itshape Elephant.} Image Size: $601\times401$. (a) Blurry Image. (b) Our uniform deblurring. (c) Harmeling \emph{et al.} \cite{Harmeling2010NIPS}. (d) Hirsch \emph{et al.} \cite{non_uniform_Hirsh}. (e) Our non-uniform deblurring. (f) Estimated non-uniform kernel.}
\label{fig:elephant}
\end{figure}

\begin{figure}[!t]
\centering
\subfloat[]{
\label{fig:Butchershop_blur}
\includegraphics[width=0.48\linewidth]{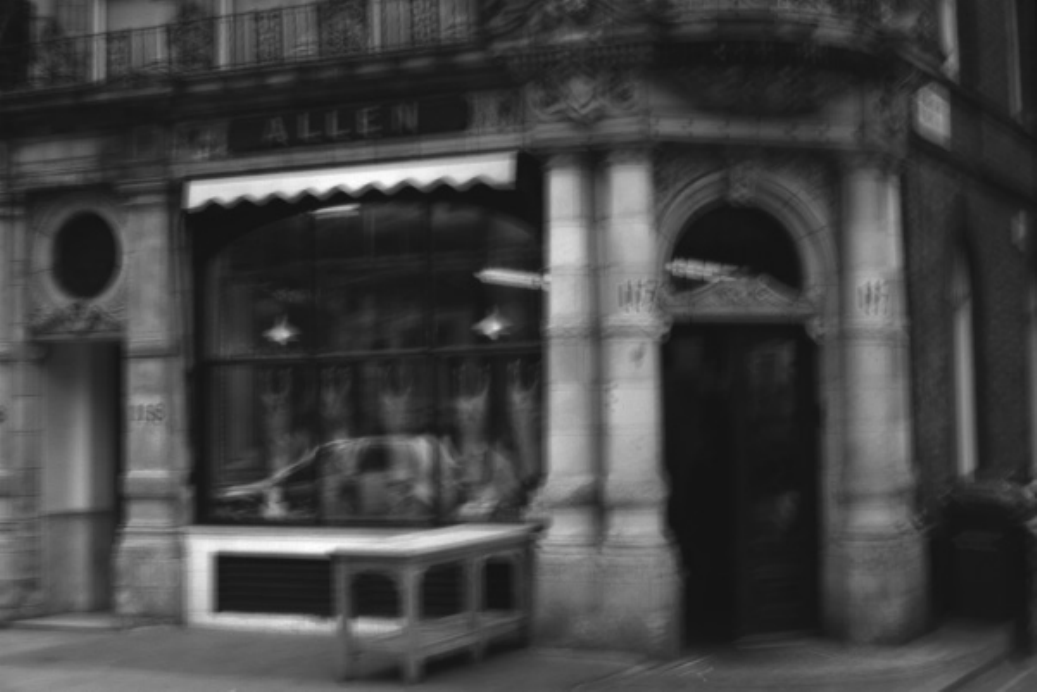}}
\subfloat[]{
\label{fig:Butchershop_harmeling}
\includegraphics[width=0.48\linewidth]{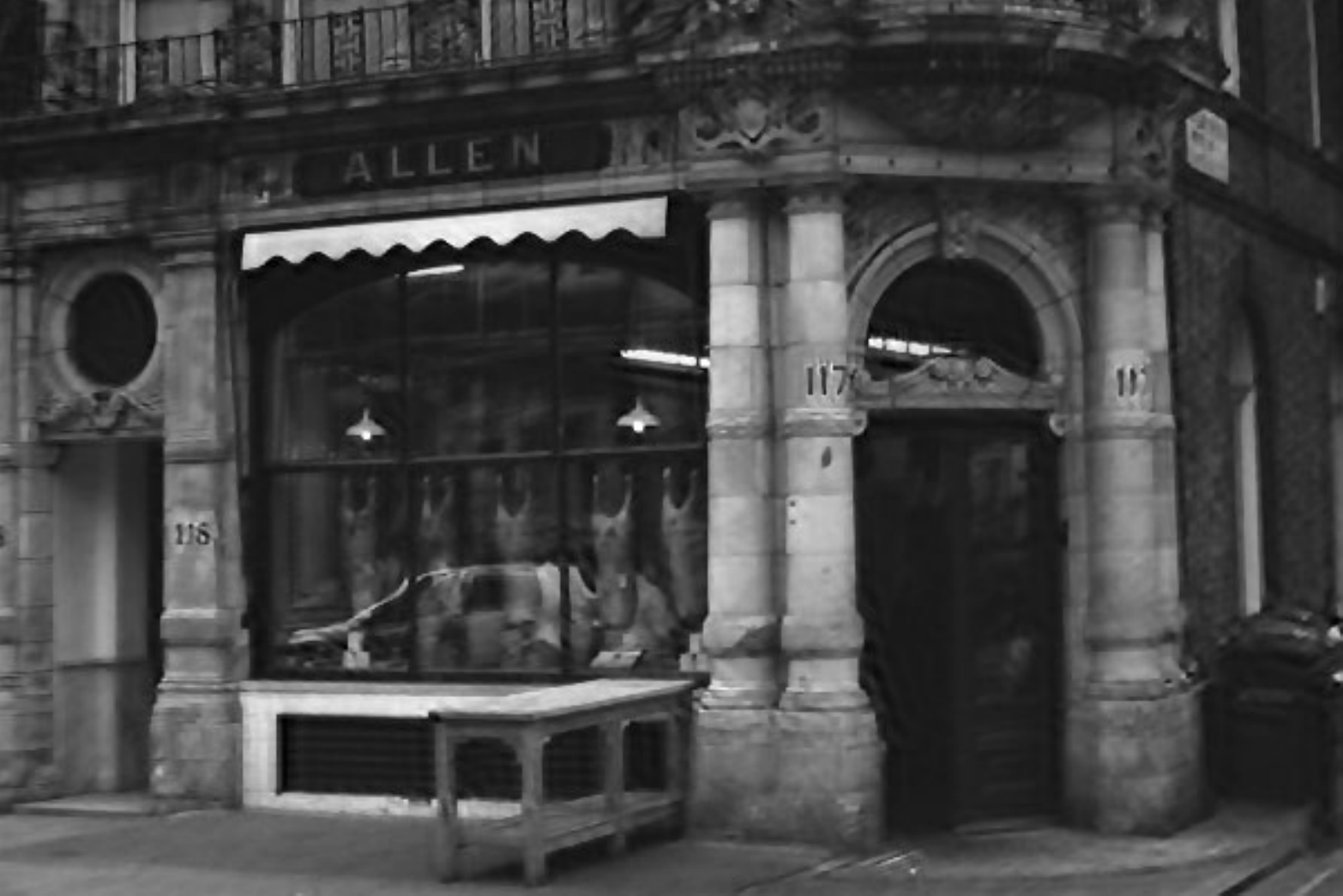}}\\
\subfloat[]{
\label{fig:Butchershop_xu}
\includegraphics[width=0.48\linewidth]{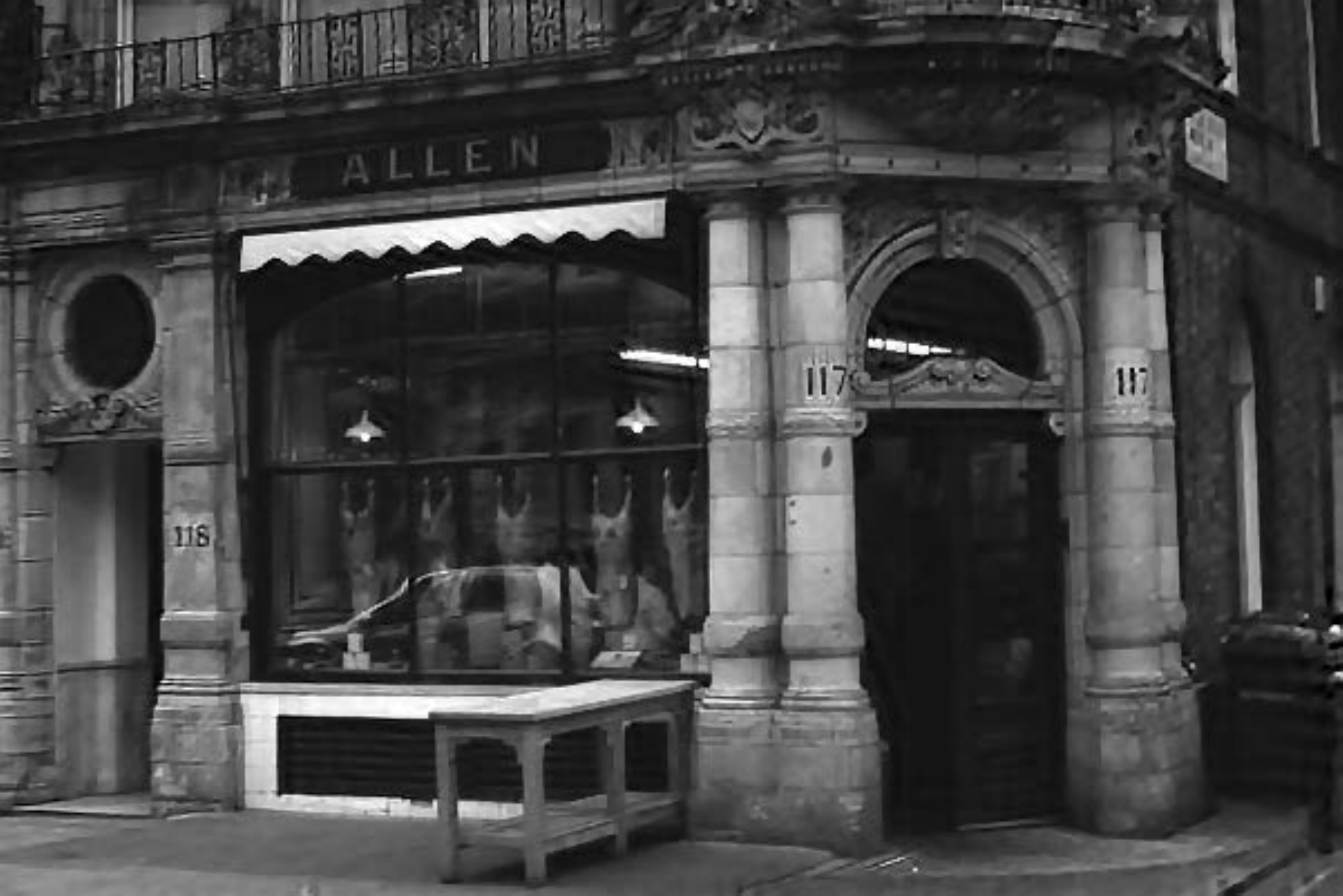}}
\subfloat[]{
\label{fig:Butchershop_Pan}
\includegraphics[width=0.48\linewidth]{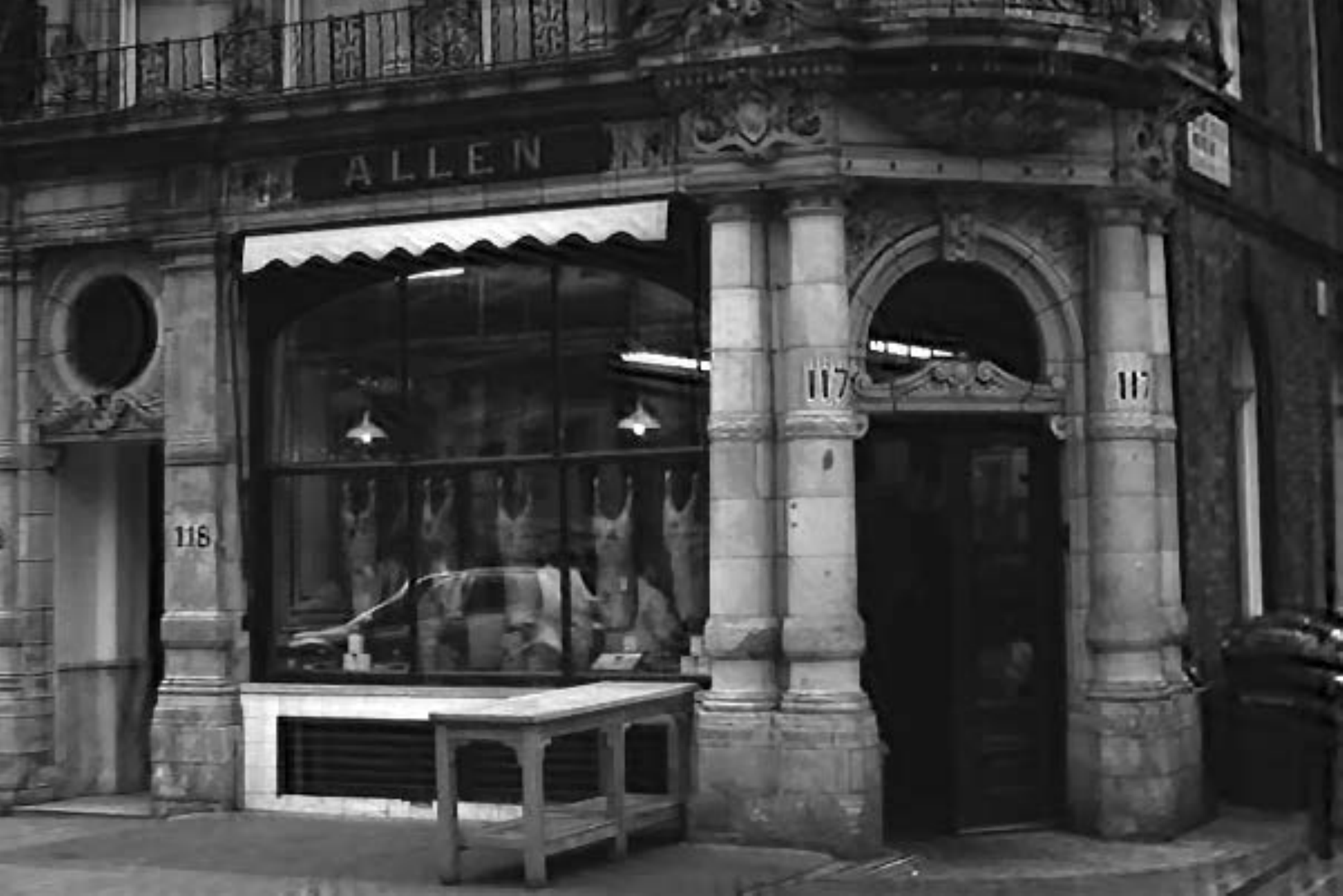}}\\
\subfloat[]{
\label{fig:Butchershop_ours}
\includegraphics[width=0.48\linewidth]{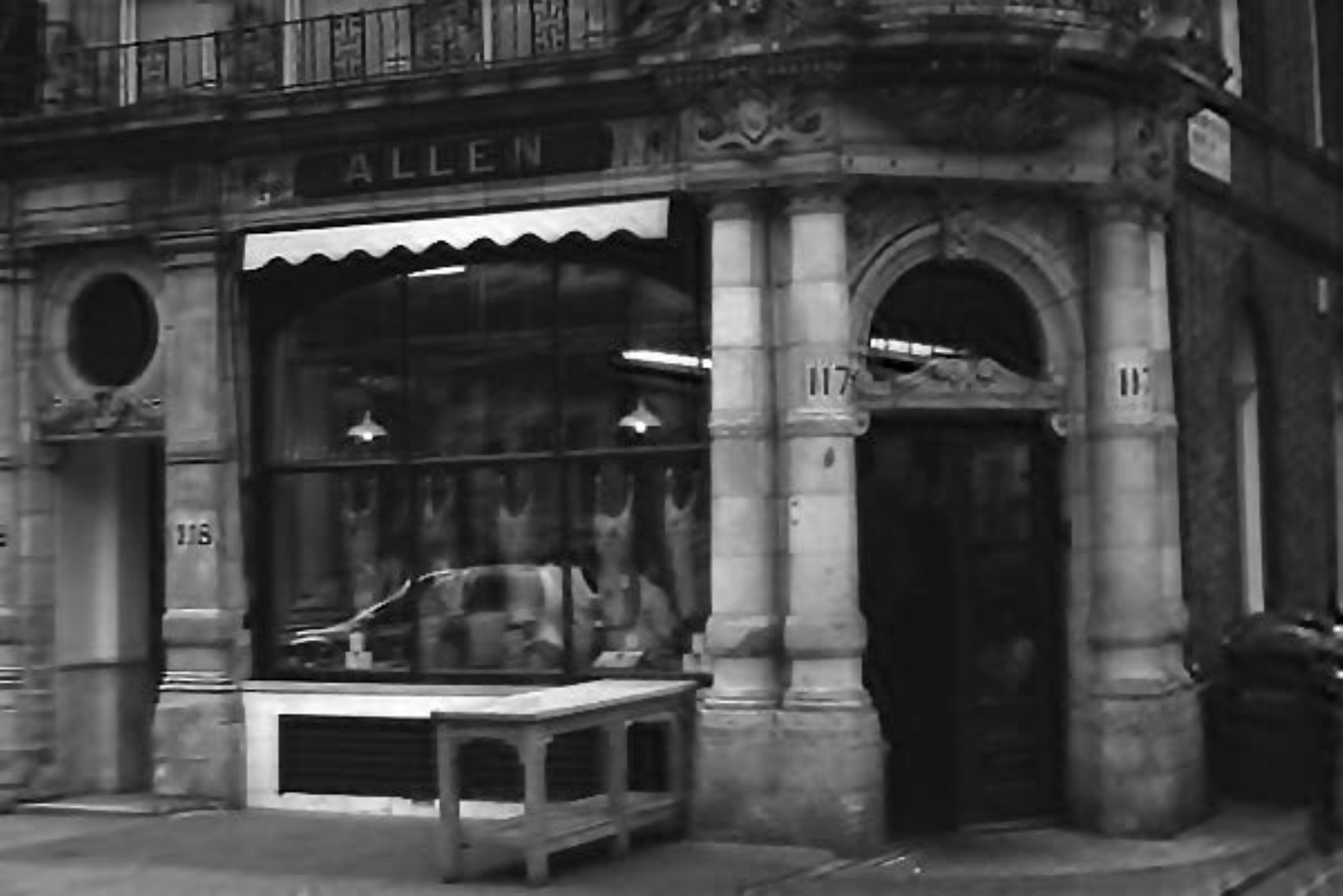}}
\subfloat[]{
\label{fig:Butchershop_kernel}
\includegraphics[width=0.48\linewidth]{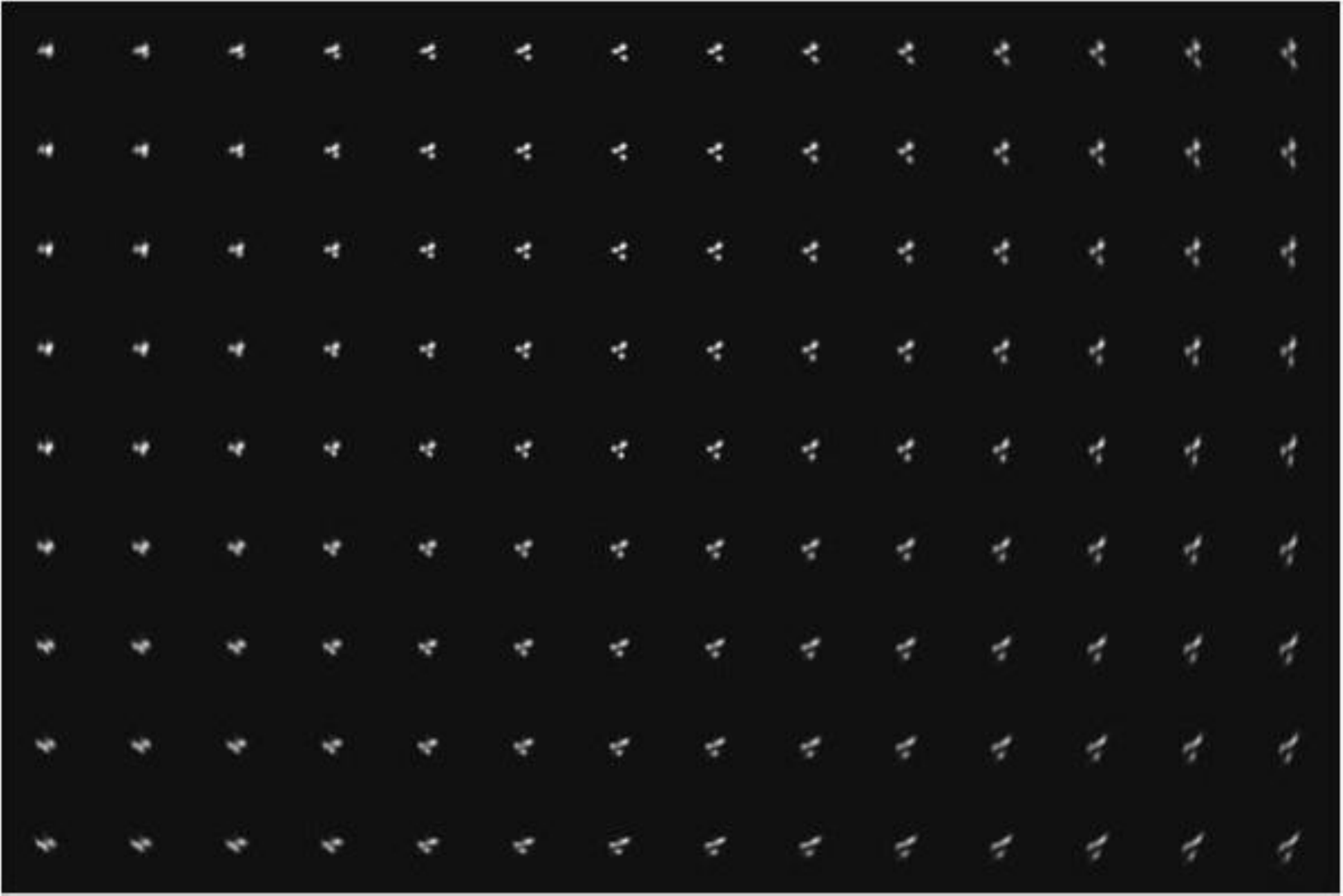}}
\caption{{\itshape Butchershop.} Image Size: $601\times401$. (a) Blurry Image. (b) Harmeling \emph{et al.} \cite{Harmeling2010NIPS}. (c) Xu \emph{et al.} \cite{xuli2013Unnatural} (d) Pan \emph{et al.} \cite{Pan2016Pami}. (e) Our non-uniform deblurring. (f) Estimated non-uniform kernel.}
\label{fig:Butchershop}
\end{figure}

In Fig.~\ref{fig:elephant} and \ref{fig:Butchershop}, we demonstrate the performance of our algorithm on the non-uniform blind deblurring problem. Directly using uniform deblurring algorithm fails to restore the correct sharp image, since the existing blur does not follow the convolution model. as shown in Fig. \ref{fig:elephant_2d}. Extended to non-uniform blur model (\ref{eq:nu_blur_model2}), our algorithm can tackle the spatial-variant blurry cases. We compare our algorithm with recent non-uniform blind deblurring algorithms \cite{Harmeling2010NIPS,non_uniform_Hirsh,xuli2013Unnatural,Pan2016Pami}, of which the results are available. Our results are visually comparable or better than those of the competing methods.

\section{Conclusion}
\label{sec:conclusion}
In this paper, we propose a \textit{Multi-Scale Latent Structure} prior, which is derived from down-sampling a blurry image. With this prior, we design an efficient algorithm to jointly restore both the latent sharp image and the blur kernel from a single blurry observation. The qualitative and quantitative experiments demonstrate that our algorithm is competitive against the state-of-the-art methods in blindly recovering uniform and non-uniform blurry images.

\textit{Limitation}: In our approach, we require that there are objects with profiles, of which the sizes are much larger than that of the blur kernel. If the precondition is disobeyed, objects will be removed after down-sampling. In this case, the data fidelity term of the optimization function fails to work and will converge to an uncertain kernel. Nevertheless, to the best of our knowledge, it is also difficult for other blind deblurring algorithms to deal with the blurry images, in which all objects are smaller than blur kernels.

Recently, there have been methods focusing on specific object deblurring problems, for example, text or face deblurring. These methods can work better by considering the properties of specific objects.
In the future, we would like to incorporate the advantage of these techniques with our algorithm to make a more robust unified framework. Besides, although the parameter \textit{kernel size} in our algorithm is not sensitive and can be set relatively large, it is still more preferred to have it automatically set according to the individuality of the blurry images. In the future, we would like to investigate how to infer the optimal \textit{kernel size} from a blurry image to make our algorithm more practical.


%





\ifCLASSOPTIONcaptionsoff
  \newpage
\fi



%

\bibliographystyle{IEEEbib}

\vfill




%








\end{document}